\documentclass[opre,nonblindrev]{informs4}

\usepackage{eqndefns-left}
\Equationvalidatefalse
\RequirePackage{tgtermes}
\RequirePackage{newtxtext}
\RequirePackage{newtxmath}
\RequirePackage{bm}
\RequirePackage{endnotes}

\definecolor{revisionblue}{RGB}{0,0,0}
\newcommand{\rev}[1]{{\color{revisionblue}#1}}
\newenvironment{revision}{\begingroup\color{revisionblue}\ignorespaces}{\endgroup}

\newif\ifonecolumnequations
\onecolumnequationstrue

\OneAndAHalfSpacedXII

\usepackage[utf8]{inputenc}
\usepackage[T1]{fontenc}
\usepackage{booktabs}
\usepackage{nicefrac}
\usepackage{microtype}
\usepackage{xcolor}
\usepackage{mathtools}
\usepackage{enumitem}
  \usepackage[hidelinks]{hyperref}
  \usepackage[capitalize,noabbrev]{cleveref}
\usepackage{pifont}
\usepackage{multirow}
\usepackage{algorithm}
\usepackage{algorithmic}
\usepackage{tikz}
\usepackage{subcaption}
\usepackage{placeins}

\usetikzlibrary{shapes,decorations,arrows,calc,arrows.meta,fit,positioning}
\tikzset{
    -Latex,auto,node distance =1 cm and 1 cm,semithick,
    state/.style ={ellipse, draw, minimum width = 0.7 cm},
    point/.style = {circle, draw, inner sep=0.04cm,fill,node contents={}},
    bidirected/.style={Latex-Latex,dashed},
    el/.style = {inner sep=2pt, align=left, sloped}
}

\let\arxivSavedH\h
\let\arxivSavedArgmax\argmax
\let\arxivSavedArgmin\argmin
\let\arxivSavedCenternot\centernot
\let\h\undefined
\let\argmax\undefined
\let\argmin\undefined
\let\centernot\undefined
\usepackage{shortcuts}
\let\h\arxivSavedH
\let\argmax\arxivSavedArgmax
\let\argmin\arxivSavedArgmin
\let\centernot\arxivSavedCenternot
\providecommand{\cA}{\mathcal A}
\providecommand{\cD}{\mathcal D}
\providecommand{\cG}{\mathcal G}
\providecommand{\cC}{\mathcal C}

\providecommand{\hatunderpi}[1]{\hat{\pi}_{#1:T}}

\usepackage{natbib}
\bibpunct[, ]{(}{)}{,}{a}{}{,}
\def\bibfont{\small}

\crefname{assumption}{assumption}{assumptions}

\newcommand{\norm}[1]{\lVert#1\rVert}

\providecommand{\mathscr}{\mathcal}
\newcommand{\blipadv}{\tau}
\newcommand{\boundadv}{B_\blipadv}
\newcommand{\FGen}{T}
\newcommand{\PJ}[2]{\mathbb{P}_{#1#2}}
\newcommand{\PI}[2]{\mathbb{P}_{#1}\otimes\mathbb{P}_{#2}}
\newcommand{\EE}{\mathbb{E}}

\renewcommand{\qed}{\unskip\nobreak\hfill\ensuremath{\square}}

\Crefname{equation}{eqn.}{Eqs.}

\EquationsNumberedThrough
\TheoremsNumberedThrough
\ECRepeatTheorems
\MANUSCRIPTNO{}

\makeatletter
\def\theARTICLETOP{}
\def\theARTICLEABSTRACT{%
  \par\vspace{12pt}
  {\ABSfont\noindent\theABSTRACT\par
   \vskip5pt\theFUNDING\theKEYWORDS\theSUBJECTCLASS}
  \par\vspace{12pt}}
\makeatother
\RRHSecondLine{}
\LRHSecondLine{}
\hypersetup{pdftitle={Decision-Centered Abstractions via Orthogonal Estimation of Difference-of-Q Functions},pdfauthor={Defu Cao and Angela Zhou}}

\begin{document}

\RUNAUTHOR{Cao and Zhou}
\ECAUpunct{: }
\RUNTITLE{Decision-Centered Abstractions}
\TITLE{Decision-Centered Abstractions via Orthogonal Estimation of Difference-of-Q Functions}

\ARTICLEAUTHORS{%
\AUTHOR{Defu Cao}
\AFF{University of Southern California}
\AUTHOR{Angela Zhou}
\AFF{University of Southern California}
}

\ABSTRACT{%
Offline reinforcement learning enables evaluation and optimization of sequential decisions from historical data, when it is not possible to deploy new policies online due to safety, cost, and other concerns. Big data advances enable rich state information, but may naively include reward- and action- irrelevant dynamics that are ultimately unnecessary for learning optimal actions. We introduce state abstractions that target preservation of the \textit{difference-of-$Q$} functions, and we propose to learn these abstractions via causal machine learning of the difference-of-$Q$ function and standard statistical sparse learning. Under a nonparametric additive-rewards model, we characterize when decision-centered abstractions are simpler than the full state space, motivating our estimation procedure. We develop a dynamic generalization of the R-learner \citep{nie2021learning,lewis2021double} for estimating difference of $Q$-functions, for discrete-valued actions $a,a_0$. We leverage orthogonal estimation to improve convergence rates, even if the required estimates of $Q$ and behavior policy converge at slower rates and prove consistency of policy optimization under a margin condition. The method can leverage black-box estimators of the $Q$-function and behavior policy to target estimation of a more structured $Q$-function contrast, and uses simple squared-loss minimization. We demonstrate variance improvements from our estimator and how our approach enables us to isolate the information needed for sequential decision-making, which can be less than that for state prediction, in simulated data and simulator-augmented real data. 
}

\SUBJECTCLASS{Dynamic programming/optimal control: Markov, infinite state;
Statistics: estimation, nonparametric;
Decision analysis: sequential}

\maketitle

\section{Introduction}
Offline reinforcement learning (RL) enables evaluating and optimizing sequential decisions when online experimentation is not possible due to cost or safety. Offline reinforcement learning therefore demands more accurate statistical learning from limited historical data. Recent advances in machine learning enable rich state information from complex sensors, for example text, images, and other high-dimensional information. However, such measurements may capture \textit{too much information} that is ultimately unneeded for making optimal decisions and slows down statistical learning. In operations, social media and other big data sources offer opportunities to improve demand forecasting, but \citet{fu2023value} finds heterogeneous impacts of topics on demand estimation. Rich state information may improve baseline demand estimation, but may not be needed to make decisions, such as pricing promotions which may instead depend on simpler information such as consumer price sensitivity. Standard offline reinforcement learning algorithms may spend scarce data learning predictable future value and dynamics that ultimately do not affect which action is preferred.

Instead, we target the information needed to make optimal decisions. We introduce \textit{decision-centered abstractions}, mappings of the state space to a simpler state space that preserves the value of the \textit{difference} of $Q$ (state-action-conditional value) functions. Decision-centered abstractions may further discard information needed to predict the value under each action separately, and therefore discard more information than typical transition or $Q$-preserving abstractions studied in the literature. Our decision-centered abstractions motivate changing the estimand to the \textit{difference of Q functions}. (Difference-of-$Q$ functions are closely related to advantage functions in reinforcement learning.) Our results characterize \textit{when} simpler decision-centered abstractions exist by refining prior graphical criteria, and answer \textit{how} to learn them with a new dynamic orthogonal estimation framework and standard sparse statistical learning. As a result, in a thresholded-LASSO linear setting, we give theoretical guarantees for off-policy evaluation that depend on the dimension of the decision-relevant part of the state, not the ambient dimension of the full state.  


{Indeed, canonical operations models recognize the importance of the \textit{difference of values} for optimal decisions. In network revenue management, bid-price controls approximate the \textit{marginal} opportunity cost of accepting or rejecting a request rather than either value level for its own sake \citep{talluri1998analysis,adelman2007dynamic}. In hospital discharge planning, \citet{shi2021timing} show that the priority ordering of patients depends on how much an \textit{additional} inpatient day changes readmission risk, not on the absolute risk level. \citet{maystre2023optimizing} studies long-term recommendation and identifies \textit{decision-centered abstractions} where comparing two recommendations requires only shared user information and consumption history for those two items, while the value of other items cancels from the comparison. In these context-rich examples and others, decision-relevant contrasts are substantially simpler than contextual main effects.}

Our approach builds on an analogous distinction in causal inference between predicting outcomes vs. predicting how outcomes change under different actions. Recent work emphasizes learning causal difference-of-effect functions such as the conditional average treatment effect (CATE) \citep{wager2018estimation,foster2019orthogonal,kunzel2019metalearners,kennedy2020optimal}. The difference that actions make on outcomes may be smoother or sparser than the main effects (what happens under either action by itself, $Q^\pi$). Estimating the causal difference or contrast enables more robust and efficient estimation leveraging orthogonal and double machine learning \citep{chernozhukov2018double,kennedy2022semiparametric}. We develop the sequential analogue: an orthogonal estimator for difference-of-$Q$ functions that generalizes residualized learning \citep{nie2021quasi} to MDPs.


Our contributions are threefold. First, we introduce decision-centered abstractions and we refine graphical criteria for recovering the coordinate-support of decision-centered abstractions under sparse additive rewards. Second, we develop an orthogonal estimator for the difference-of-Q. We introduce two variants depending on the depth of recursively-unrolled future $Q$: one-step vs. full unrolling that rewrites entirely based on recursive estimation of difference-of-$Q$ functions  akin to approaches inspired by SNMMs \citep{robins2004optimal}. 
Our main results achieve orthogonality with respect to current-timestep behavior policy and $Q$ functions, while our two variants trade off between dependence on future-$Q$ estimation error vs. variance reduction. 
Under well-specified difference-of-$Q$, with behavior policy and conditional-mean regressions $o_p(n^{-\frac 14})$-convergent and a weaker \textit{projected orthogonality requirement} that the \textit{actionwise-difference} of future $Q$-estimation errors is $o_p(n^{-\frac 12})$, we obtain worst-case $O_p(n^{-\frac 12})$ rates for estimating the difference of $Q$ functions and attaining the optimal policy value. (Another variant can remove the requirement on future action-projected $Q$-estimation errors, at the expense of additional variance). The difference of estimation errors across actions can be much simpler in the exact same exogenous-dynamics settings that motivate our work. Third, we establish structure-adaptive estimation via standard sparse statistical learning (thresholded LASSO) and policy-learning guarantees. 

Our paper proceeds as follows.\footnote{A preliminary version of partial results was accepted to the AISTATS conference under a different title, ``Structured Difference-of-$Q$ via Orthogonal Estimation'' \citep{cao2024structured}. Relative to that conference version, we additionally define decision-centered abstractions and give graphical criteria for finding simpler ones (\Cref{sec-abstractions}), generalize our orthogonal estimation via recursive substitution to $L=T-t$ in \Cref{sec-method}, and have expanded the experiments. } \Cref{sec-abstractions} characterizes decision-centered abstractions, \Cref{sec-method} introduces our orthogonal estimation and its statistical guarantees, and \Cref{sec-policy-opt} translates our evaluation guarantees into improvements in local policy improvement and optimization. In \Cref{sec-experiments}, we illustrate improvements on simulated and simulator-augmented real data. 


\section{Related work}
We discuss the most closely related methodological work here; see \Cref{sec-addl-discussion} for further discussion and \Cref{sec-abstractions} for discussion of related work on state abstraction. There is a large literature on offline reinforcement learning (evaluation and optimization) \citep{jin2021pessimism,xie2021bellman}, including approaches that leverage importance sampling or introduce marginalized versions \citep{jl16,thomas2015high,kallus2019double,liu2018breaking}. For Markov decision processes, {other papers study statistically semiparametrically efficient or doubly-robust estimation, but of the \textit{averaged policy value} $\E[V_1^{\pi^e}(S_1)]$, rather than MSE convergence of the difference-of-Q function as we do here \citep{kallus2019double,kallus2019efficiently,xie2023semiparametrically}}. The literature on dynamic treatment regimes (DTRs) studies a method called advantage learning \citep{schulte2014q}, although DTRs in general lack reward at every timestep, whereas we are particularly motivated by sparsity implications that arise jointly from reward and transition structure. Beyond policy value estimation, we seek the entire contrast function. 

We discuss crucial distinctions from other works studying advantage functions. Advantage functions appear in RL and dynamic treatment regimes \citep{neumann2008fitted,murphy2003optimal}. Unlike advantage functions in RL and DTRs, our difference-of-Q is $\pi_t$ policy-independent at time $t$, unlike advantage functions, simplifying optimization. 
\looseness=-1
\citet{pan2023learning,pan2024skill} propose direct advantage estimation that avoids $Q$-functions but requires a challenging nonconvex constraint. \citet{farias2022markovian} develop an estimator called Differences-In-Q for the difference in average value under all-treat or all-control. The estimator averages a difference of Q functions for variance reduction. Our method can be used in their setting, although their target is a different averaged policy-value estimand. Closest to our goal is \citet{shi2022statistically}, who derive a pseudo-outcome for estimating $Q$-contrasts in the infinite-horizon setting. Our method differs in focusing on the finite-horizon case, leveraging an orthogonal R-learner formulation that provides improved statistical guarantees and practical convex estimation.

Many works have recognized the crucial importance of simpler dynamic structure for better operational decision-making. We view our approach as generally complementary. \citet{hu2025optimal} study optimal targeting in dynamic systems, such as patient-driven queueing systems, and identify structure that enables threshold policy characterizations on simpler heterogeneous treatment effect estimands, rather than generic $Q$ functions. \citet{sinclair2023hindsight} notes that known system transitions on exogenous demand permits many operations MDPs to be optimized via backtesting, which can improve sample complexity of online learning as well \citep{wan2024exploiting,zhang2025reinforcement}. \citet{bennouna2025learning} analogously pursue minimal state representations via clustering, though they seek to preserve value functions. In machine learning, recent works have focused on filtering irrelevant dynamics. Some require additional latent representation assumptions with more complex inverse-kinematics estimation \citep{efroni2021provably,efroni2022sparsity}, or rely on causal graphical structure to identify model-preserving abstractions learned by representation learning (but without theoretical guarantees) \citep{wang2021task,wang2022denoised,liu2023}. We characterize the state information sufficient to preserve the difference-of-$Q$ functions, which can further discard action-irrelevant value information, and whose sparsity is adaptive to sources of underlying structure. However, when additional structural information is known, further specific improvements are possible. 
At a high level, our method is similar to the dynamic DR learner studied in \citep{lewis2021double} in that we extend the R-learner identification approach to a sequential setting, although the estimand is quite different. They generalize structural nested-mean models by estimating ``blip-to-zero'' functions but only consider heterogeneity based on a fixed \textit{initial} state and dynamic treatment regimes with terminal rewards. Contemporaneous with the first version of this paper, \citet{ma2025deepblip} develops a history-dependent neural version of \citet{lewis2021double}, although the Markovian analogue of their estimator resembles our fully-unrolled variant, rather than our preferred one-step specification.
Note that the (single-stage) R-learner loss function is an overlap-weighted \citep{li2019addressing} regression against the doubly-robust score (DR-learner \citep{kennedy2020optimal}); \citet{javurek2026orthogonal} focuses instead on estimating $Q$ functions.  We do make a margin assumption to relate convergence of Q-function contrasts to policy value, analogous to \citep{shi2022statistically,hu2024fast}.

\section{Decision-centered Abstractions}\label{sec-abstractions}
\subsection{Problem Setup}
We consider a finite-horizon Markov Decision Process, $\mathcal M = (\mathcal{S}, \mathcal{A}, r, P, \gamma, T)$ of state space $\mathcal{S}$, discrete action space $\mathcal{A}$, reward function $r: \mathcal{S}\times\mathcal{A} \to \mathcal{R} $, transition probability $P\colon \mathcal{S} \times \mathcal{A} \rightarrow \Delta(\mathcal{S})$, where  $\Delta(\mathcal{S})$ is the set of distributions over $\mathcal{S}$, discount factor $0\leq \gamma<1$, and time horizon of $T$ steps. We let $t=1, \dots, T$ index timesteps. We let the state spaces $\mathcal{S}\subseteq \mathbb{R}^d$ be continuous, and assume the action space $\mathcal{A}$ is finite.
Following causal conventions we denote $\pi(a\mid s)$ as the probability of taking action $a$ in state $s$; at times we omit dependence on function arguments referring to the policy function $\pi$. 
Capital letters denote random variables $(S_t, A_t, …)$, lower case letters $s,a$  denote evaluation at a generic value. 

The value function is $V^\pi_t(s) := \E_\pi[ \sum_{t'=t}^{T} \gamma^{t'-t} R_{t'} \mid S_t=s ]$ where $\E_\pi$ denotes expectation under the joint distribution induced by the MDP $\mathcal{M}$ running policy $\pi$. The $Q-$function is the $(s,a)$-conditional expectation of discounted future rewards.
The optimal value and $Q$-functions are denoted $V^*,Q^*$ under the optimal policy. 
\begin{assumption}[Markov and sequential unconfoundedness]
\label{asn-sequential-unconf}
The observed-data law factorizes as follows:
\begingroup
\setlength{\abovedisplayskip}{6pt}
\setlength{\abovedisplayshortskip}{6pt}
\setlength{\belowdisplayskip}{6pt}
\setlength{\belowdisplayshortskip}{6pt}
\begin{align*}
\textstyle p(S_1,A_1,R_1, \dots, S_T,A_T,R_T) = p_{S_1}(S_1)
\left\{ \prod_{t=1}^{T-1} \pi_t^b(A_t \mid S_t)
p(S_{t+1},R_t \mid S_t,A_t)\right\} 
\pi_T^b(A_T \mid S_T)p(R_T\mid S_T,A_T).
\end{align*}
\endgroup
Under the behavior policy $\pi_b$, actions $A_t$ were taken with probability depending on observed states $S_t$ alone (and not on any unobserved states).
\end{assumption}
This corresponds to assuming MDP rather than POMDP structure and is a common implicit assumption in RL.

At times we omit the function evaluation where it is clear from context, for example referring to the reward function $R_t(S_t,A_t)$ as $R_t$ in longer expressions. 
We focus on estimating the difference of $Q$-functions (each under the same policy), $\blipadv_t^\pi(s) =Q_t^\pi(s,1) - Q_t^\pi(s,0).$ 
We focus on the offline reinforcement learning setting with a historical dataset of $n$ offline trajectories, $\textstyle \mathcal{D}=\{(S_t^{i}, A_t^{i}, R_t^{i},S_{t+1}^{i})_{t=1}^{T}\}_{i=1}^n$, where actions were sampled under a behavior policy $\pi^b.$ 
Notationally: following convention in statistical papers on causal inference, we denote the $\mathcal{L}_2(P)$-norm $\norm{f(X)}_2:=\mathbb{E}[ |f(X)|^2 ]^{1 / 2}$; expectations and norms are under the observational behavior distribution unless otherwise indicated.
\begin{revision}
\subsection{Decision-centered Abstractions}

Operational data may contain state variables needed to describe the
data-generating process, predict next state or main-effects of actions ($Q$ or $V$ value functions), but may not affect the value difference of taking different actions.


\paragraph{State abstractions.}
A state abstraction is a measurable map from the full state space to an
abstract state space, $\phi_t:\mathcal S\to\mathcal Z_t$.
A state abstraction retains only \(\phi_t(s)\) and treats any two full
states mapped to the same abstract state as indistinguishable. In the
coordinate-projection case,
for \(K\subseteq\{1,\ldots,d\}\), write $
    \phi_t^K(s)=s^K$.
Later we specialize \(K\) to
\(K_{\Delta,t}\) when discussing coordinate support for decision-centered abstractions. \citet{li2006towards} classify exact MDP state abstractions by the object
preserved under aggregation. For example, model-based abstractions $\phi_{\mathrm{model}}$ preserve one-step rewards and abstract transitions. In order of increasing state compression, $\phi_{Q^\pi}$ preserves $Q^\pi(s,a)$ for all policies $\pi$ and actions $a$, $\phi_{Q^*}$ preserves $Q^*(s,a)$ for all actions $a$, $\phi_{a^*}$ preserves a common optimal action and the common optimal value level, and the strictest abstraction $\phi_{\pi^*}$ preserves a common optimal action only.
Thus \(\pi^*\)-irrelevance answers only the representation question:
does there exist an optimal policy that can be written as a function of the
abstract state? 

These abstractions preserve less and less information. Every abstraction preserving the optimal Q-vector preserves the optimal $\blipadv$, and every abstraction preserving those $\blipadv$ contrasts preserves a greedy optimal action. 
The $\pi^*$-irrelevant abstractions are minimal in the classical state abstraction hierarchy, but they do not preserve enough information to support \textit{learning} the optimal policy. That is, \citet{li2006towards} show negative planning and $Q$-learning results where \(\phi_{\pi^*}\) can represent an optimal policy when it is known, but solving the induced abstract MDP need not recover the optimal policy. We introduce decision-centered abstractions which address a gap with existing state abstractions: they can filter out more information than existing model or $Q$-based abstractions, but preserve signal that enables learning (the regret of taking wrong actions), and with full-state nuisance $Q$-function estimation as in our later orthogonal learning procedure, they can be used for policy optimization.


\paragraph{Decision-centered abstraction.}

We formally define \textit{decision-centered abstractions} which preserve $\blipadv^\pi$ (given a continuation policy class). Decision-centered abstractions require that the difference-of-$Q$ function $\blipadv$ is constant within each abstract state, given a continuation policy. 

\begin{definition}[Decision-centered abstraction]
\label{def:decision-centered}
Fix a stage \(t\), a continuation-policy class
\(\Pi_{t+1:T}\), and an action-comparison class
\(\mathcal A_t^{\mathrm{cmp}}\subseteq\mathcal A\times\mathcal A\).
A measurable map \(\phi_t^\Delta:\mathcal S\to\mathcal Z_t\) is a
decision-centered abstraction for
\((\Pi_{t+1:T},\mathcal A_t^{\mathrm{cmp}})\) if, for every
\(\pi\in\Pi_{t+1:T}\), every
\((a,a_0)\in\mathcal A_t^{\mathrm{cmp}}\), and all \(s,\tilde s\)
in the relevant support,
\[
\phi_t^\Delta(s)=\phi_t^\Delta(\tilde s)
\quad\Longrightarrow\quad
\blipadv_t^\pi(s;a,a_0)=\blipadv_t^\pi(\tilde s;a,a_0).
\]
\end{definition}

For the remainder of this section, the continuation-policy class
$\Pi_{t+1:T}$ and action-comparison class $\mathcal A_t^{\mathrm{cmp}}$ are fixed and
suppressed from the support notation. 

\paragraph{Sparser decision-centered abstractions via sparser coordinate support.}
We are particularly interested when valid decision-centered abstractions identify simpler operational states, and accordingly we track the \textit{coordinate support} of valid decision-centered abstractions. Define
\begin{equation}
K_{\Delta,t}
:=
\left\{
    j\in[d]:
    \blipadv_t^\pi(\cdot;a,a_0)
    \text{ is nonconstant in }s_j
    \text{ for some }\pi\in\Pi_{t+1:T},\ (a,a_0)\in\mathcal A_t^{\mathrm{cmp}}
\right\},
\label{eq:decision-support}
\end{equation}
where ``nonconstant in $s_j$'' means that there exist two states
$s,\widetilde s$ in the relevant support such that
$s_{-j}=\widetilde s_{-j}$ but $\blipadv_t^\pi(s;a,a_0)
\neq
\blipadv_t^\pi(\widetilde s;a,a_0)$. If the relevant stage-$t$ support is a
Cartesian product of its coordinate projections, then
$\phi_t^\Delta(s)=s_{K_{\Delta,t}}$ is a decision-centered abstraction. Indeed,
any two supported states with the same $K_{\Delta,t}$-coordinates can be joined
by changing their coordinates in $K_{\Delta,t}^c$ one at a time. Product support
keeps every intermediate state in the support, and the definition of
$K_{\Delta,t}$ ensures that each change leaves every required difference-of-$Q$
unchanged.
For fixed-policy evaluation, $\Pi_{t+1:T}=\{\pi^e\}$; for policy
optimization, the continuation classes are selected backward. 

Studying the coordinate support is motivated by classical linear sparsity, where only a subset of coordinates is required for prediction. Later on, we discuss how we can combine our modified estimand (the difference-of-$Q$ function) with standard sparse statistical learning parametrizations therein in order to learn decision-centered abstractions.

  Notably, decision-centered abstractions crucially restore learning signal to $\pi^*$-irrelevant abstractions that preserve the optimal action, but decision-centered abstractions are not good candidates for generic \textit{state-aggregation}, in that plug-in MDP optimization on decision-centered abstractions may not recover the optimal policy. Decision-centered abstractions crucially preserve the \textit{decision loss} of incorrect actions in addition to their identity. Recall that $\pi_t^*(s)=\mathbb I[\tau_t^*(s)>0], \hat\pi_t(s)=\mathbb I[\hat\tau_t(s)>0]$.
The decision loss satisfies that 
\[
Q_t^*(s,\pi_t^*(s))
-
Q_t^*(s,\hat\pi_t(s))
=
|\tau_t^*(s)|
\indic{\hat\pi_t(s)\neq\pi_t^*(s)}.
\]
The
difference-of-\(Q\) target additionally retains the cost of choosing the
wrong action, whereas \(\pi^*\)-irrelevant abstraction only retains the identity of correct actions. 



 \begin{proposition}[Position in the abstraction
  hierarchy]
  \label{prop:hierarchy}
  For any states \(s,\tilde s\) in the
  relevant stage-\(t\) support,
  under a fixed reference action \(a_0\) and tie-
  breaking rule,
  \[
  Q_t^*(s,\cdot)=Q_t^*(\tilde s,\cdot)
  \quad\Longrightarrow\quad
  \tau_t^*(s;\cdot,a_0)
  =
  \tau_t^*(\tilde s;\cdot,a_0)
  \quad\Longrightarrow\quad
  \pi_t^*(s)=\pi_t^*(\tilde s).
  \]
  Neither converse holds in general.
  \end{proposition}
The claim follows from \Cref{prop:contrast-abstraction-hierarchy}.


\begin{figure*}[t!]
\centering


\begin{minipage}[t]{0.485\textwidth}
\centering
\begin{tikzpicture}[
    scale=0.95,
    node distance=2.5mm and 6mm, 
    state/.style={circle, draw, minimum width=0.55cm, inner sep=1.2pt},
    -Latex, semithick
]
    \node[state] (s) at (0,0) {$S^\rho_0$};
    \node[state] (x) [above=of s] {$S^{\rho_c}_0$};
    \node[state] (a) [right=of s] {$A_0$};
    \node[state] (r) [below=of a] {$R_0$};
    \node[state] (s1) [right=of a] {$S^{\rho}_1$};
    \node[state] (a1) [right=of s1] {$A_1$};
    \node[state] (r1) [below=of a1] {$R_1$};
    \node[state] (x1) [above=of s1] {$S^{\rho_c}_1$};

    \path (s) edge (r)
          (s) edge (a)
          (a) edge (s1)
          (a) edge[dotted] (x1)
          (a) edge (r)
          (x) edge (a)
          (x) edge (x1)
          (s) edge (x1)
          (s) edge[bend right=18] (s1)
          (s1) edge (a1)
          (s1) edge (r1)
          (a1) edge (r1)
          (x1) edge (a1);
\end{tikzpicture}
\end{minipage}
\hfill
\begin{minipage}[t]{0.485\textwidth}
\centering
\begin{tikzpicture}[
    scale=0.90,
    state/.style={circle, draw, minimum width=0.55cm, inner sep=1.2pt},
    policy/.style={-Latex, semithick, densely dashed, gray!75},
    -Latex,
    semithick
]
    \node[state] (s) at (0,0) {$S^\rho_0$};
    \node[state] (x) at (0,1.05) {$S^{\rho_c}_0$};
    \node[state] (a) at (1.35,0) {$A_0$};
    \node[state] (s1) at (2.70,0) {$S^\rho_1$};
    \node[state] (x1) at (2.70,1.05) {$S^{\rho_c}_1$};
    \node[state] (a1) at (4.05,0) {$A_1$};
    \node[state] (R0) at (1.35,-0.85) {$R_0$};
    \node[state] (R1) at (4.05,-0.85) {$R_1$};

    \path
        (s) edge (a)
        (s) edge[bend right=18] (s1)
        (a) edge (s1)
        (x) edge (x1)
        (s1) edge (a1)
        (s) edge (R0)
        (a) edge (R0)
        (x) edge (R0)
        (s1) edge (R1)
        (a1) edge (R1)
        (x1) edge (R1);
    \draw[policy] (x) --
        node[pos=0.48,above,font=\scriptsize,text=gray!75] {$\pi^b$}
        (a);
    \draw[policy] (x1) --
        node[pos=0.48,above,font=\scriptsize,text=gray!75] {$\pi^b$}
        (a1);
\end{tikzpicture}
\end{minipage}
\caption{(a) Reward-relevant/irrelevant factorization \citep{zhou2024rewardrelevancefiltered}. Dotted $A_t\!\to\!S^{\rho_c}_{t+1}$ is optional. The optimal $Q$ is sparse in $s^\rho$ (i.e. doesn't change if $s_0^{\rho_c}$ changes). (b) Exogenous–Endogenous MDP \citep{dietterich2018discovering}. Rewards are additively separable: $R_t=r_t^\rho(S_t^\rho,A_t)+g_t(S_t^{\rho_c})$. $Q$ not sparse in $s^\rho$, but $Q(s,1)-Q(s,0)$ is.}
\label{fig:rewardrelevant}\label{fig:exoendo}
\end{figure*} 

Prior approaches leverage structural causal models (graphical models for Markov decision processes) to give conditional (in)dependence structural conditions for identifying which components of the state variable are sufficient for model-based abstractions. \Cref{fig:rewardrelevant,fig:exoendo} illustrates two similar causal graphs that nonetheless admit different abstractions. Such analyses partition the state variable into ``cluster nodes'' (concatenation over dimensions); in \Cref{fig:rewardrelevant,fig:exoendo} $S^\rho$ is called the ``endogenous'' part of the state and $S^\rho_c$ is the ``exogenous'' part of the state. In the graph on the left, \citet{zhou2024rewardrelevancefiltered},  $S^\rho$ is both a valid $Q-$ and decision-centered abstraction. The graph on the right illustrates \citep{pan2023learning}, a variant of the ``endogeneous-exogenous MDP''  \citet{dietterich2018discovering} with additive rewards, where $S^\rho$ is \textit{not} a valid $Q$-abstraction but is a valid decision-centered abstraction.
\textit{A priori}, either the graph on the left or right could be plausible, and therefore methods designed assuming one model is correct may not perform well if it is not. However, $S^\rho$ remains a valid \textit{decision-centered abstraction} across these different graphical substructures. 


\paragraph{Decision-centered abstractions need not support state aggregation} The classical literature also studies \textit{state aggregations}, with the goal of preserving dynamic programming policy optimization guarantees on the abstracted state space alone \citep{van2006performance,tsitsiklis1996feature,duan2019state}. The finest abstraction, $\pi^*$-irrelevant abstraction, is unsuitable for learning because optimal dynamic programming on a $\pi^*$-irrelevant abstraction need not preserve the optimal policy. Decision-centered abstractions similarly do not make good \textit{state aggregations}.

\begin{proposition}[No induced abstract-MDP guarantee]
\label{prop:no-abstract-planning}
Decision-centered abstractions do not guarantee that Bellman planning after
state aggregation recovers a ground-optimal policy.
\end{proposition}

\Cref{prop:no-abstract-planning} rules out learning $\phi^\Delta(S_t)$ and conducting dynamic programming on the decision-abstracted state space $\phi^\Delta(S_t)$ directly. Nonetheless, decision-centered abstractions can be used for policy optimization when $\blipadv_t^\pi$ is learned as a function of $\phi^\Delta(S_t)$, which we do in \Cref{sec-policy-opt}.


\subsection{Characterizing When Decision-Centered Abstractions Reduce State Complexity}

When are minimal decision-centered abstractions smaller than the original state space? We seek a primitive characterization of which state coordinates belong to
$K_{\Delta,t}$. 
Importantly, characterizing sufficient decision-centered abstraction coordinate support is not a property of the causal structural graph alone: it also depends on \textit{effect modification}, the functional dependence of rewards on state coordinates. Although this subsection focuses on developing graphical criteria, our later sections on learning and estimation \textbf{do not} require knowledge of the causal graph. 

Prior work leverages graphical causal models to isolate the subcomponents of state that are needed for policy optimization. IFactor \citep{liu2023} subindexes each current-state coordinate along two dimensions: whether the previous action has a direct causal edge into the coordinate (\(S_t^{a\cdot}\) if so; \(S_t^{\bar a\cdot}\) otherwise), and whether the coordinate is reward-relevant (\(S_t^{\cdot r}\) if so; \(S_t^{\cdot\bar r}\) otherwise). The first dimension describes controllability, not relevance to the current decision. Combining these dimensions classifies coordinates by controllability and reward relevance. IFactor develops causal-graphical criteria to obtain a \textit{model-irrelevance abstraction}: coordinate $j$, \(S_t^j\), is reward-relevant when it has a directed path to a
future reward, and such reward-relevant states are sufficient for
policy optimization under standard causal graphical assumptions. 

These prior works focus on preserving $Q$ functions. In our work, we focus on sufficient decision-centered abstractions based on $\blipadv$, the difference of $Q$ functions. In order to refine these abstractions, we do require additional functional form information as to whether differencing the reward across actions eliminates the effect modification from any particular state coordinate $S^j$. In particular, we assume that the reward is additively separable and will track which coordinate enters which additive terms, augmenting the graphical structure.
\paragraph{Additive reward scopes.}
Throughout this subsection, \(A_t\) is a single finite-valued action
variable rather than a vector of action components. We borrow the additive
term-scope representation of \citet{enouen2025higher} only for the reward
function. At every time \(t\), suppose
\begin{equation}
 r(S_t,A_t)
 =
 \sum_{H\in\mathcal H_t}
 r_{t,H}(S_{t,H},A_t).
 \label{eq:additive-reward-scopes}
\end{equation}
Here each \(H\subseteq[d]\) records the state coordinates entering one
additive term. A term may depend on the single action variable \(A_t\), or it
may be constant in its action argument. Accordingly, the actual parent set
of the term node \(r_{t,H}\) consists of \(\{S_t^j:j\in H\}\), together with
\(A_t\) only when \(r_{t,H}\) is nonconstant in \(A_t\). The decomposition
is irreducible with respect to these actual parent sets: no term can be
written as a sum of functions on proper subsets of its parents. Thus
\(\mathcal H_t\) records which state coordinates and, when applicable, the
single action variable, enter the same irreducible additive reward term. By keeping track of the additive reward terms, we can distinguish
state-dependent contributions that cancel from action-value contrasts from
those that do not.

\paragraph{Graphical refinement of IFactor with additive reward scopes.}
We augment the causal graph induced by an MDP with IFactor's state classification to additionally indicate \textit{which incoming direct causal dependences into rewards enter in which additive terms}.
Aggregate reward nodes alone do not encode this additive
separation. We therefore represent each summand
\(r_{t,H}(S_{t,H},A_t)\) as a separate deterministic node and connect the
term nodes to the period-\(t\) reward through a deterministic sum node.
The incoming state edges to \(r_{t,H}\) are exactly
\(\{S_t^j\to r_{t,H}:j\in H\}\); the edge
\(A_t\to r_{t,H}\) is present only when the term is nonconstant in its
action argument.

\begin{figure}[ht!]
\centering
\resizebox{0.98\textwidth}{!}{%
  \begin{tikzpicture}[
  font=\small,
  latent/.style={
    draw,
    circle,
    minimum size=0.90cm,
    inner sep=1pt,
    font=\scriptsize
  },
  rewardirrelevant/.style={latent,fill=violet!13},
  rewardrelevant/.style={latent,fill=blue!13},
  observed/.style={
    draw,
    circle,
    fill=gray!22,
    minimum size=0.86cm,
    inner sep=1pt,
    font=\scriptsize
  },
  term/.style={
    draw,
    rounded corners=2pt,
    fill=yellow!16,
    minimum width=2.22cm,
    minimum height=0.62cm,
    align=center,
    inner sep=2pt,
    font=\scriptsize
  },
  operator/.style={
    draw,
    densely dashed,
    circle,
    fill=white,
    minimum size=0.58cm,
    inner sep=0pt,
    font=\bfseries
  },
  block/.style={draw=black!48,rounded corners=3pt},
  trans/.style={-{Latex[length=1.8mm]},semithick,black!78},
  actionedge/.style={-{Latex[length=1.8mm]},semithick,red!72!black},
  rewardedge/.style={-{Latex[length=1.8mm]},semithick,blue!68!black},
  policy/.style={-{Latex[length=1.8mm]},semithick,densely dashed,gray!72},
  guide/.style={font=\scriptsize,text=black!62,align=center}
]
  \node[rewardirrelevant] (x00) at (-1.50,0.60)
    {\(S_t^{\bar a\bar r}\)};
  \node[rewardirrelevant] (x01) at (-1.50,-0.55)
    {\(S_t^{a\bar r}\)};
  \node[rewardrelevant,draw=blue!72!black,very thick] (x02) at (0.0,-0.10)
    {\(S_t^{\bar a r}\)};
  \node[rewardrelevant] (x03) at (0.0,-1.25)
    {\(S_t^{a r}\)};

  \node[rewardirrelevant] (x10) at (2.50,0.60)
    {\(S_{t+1}^{\bar a\bar r}\)};
  \node[rewardirrelevant] (x11) at (2.50,-0.55)
    {\(S_{t+1}^{a\bar r}\)};
  \node[rewardrelevant] (x12) at (4.00,-0.10)
    {\(S_{t+1}^{\bar a r}\)};
  \node[rewardrelevant] (x13) at (4.00,-1.25)
    {\(S_{t+1}^{a r}\)};

  \draw[block] (-2.22,-1.13) rectangle (-0.78,1.18);
  \draw[block] (-0.72,-1.83) rectangle (0.72,0.48);
  \draw[block] (1.78,-1.13) rectangle (3.22,1.18);
  \draw[block] (3.28,-1.83) rectangle (4.72,0.48);
  \node[guide,anchor=south] at (-1.50,1.24) {\(S_t^{\bar r}\)};
  \node[guide,anchor=north] at (0.00,-1.89) {\(S_t^{r}\)};
  \node[guide,anchor=south] at (2.50,1.24) {\(S_{t+1}^{\bar r}\)};
  \node[guide,anchor=north] at (4.00,-1.89) {\(S_{t+1}^{r}\)};

  \draw[trans] (-0.78,0.60) -- (1.78,0.60);
  \draw[trans] (0.72,0.10) to[out=0,in=180] (1.78,-0.35);
  \draw[trans] (0.72,-1.30) -- (3.28,-1.30);

  \node[observed,draw=blue!72!black,very thick] (action) at (2.10,-2.35)
    {\(A_t\)};
  \draw[actionedge] (action.north) to[out=90,in=-105] (x11.south);
  \draw[actionedge] (action.east) to[out=25,in=-145] (x13.south west);
  \draw[policy] (x02.south east) to[out=-35,in=145] (action.north west);
  \draw[policy] (x03.south east) to[out=-25,in=175] (action.west);

  \node[term] (nexttermother) at (6.25,0.55)
    {\(r_{t+1,H'}\)};
  \node[term,draw=blue!72!black,very thick] (nextterm) at (6.25,-0.65)
    {\(r_{t+1,H}\)};
  \node[guide] (otherterms) at (7.70,-1.25) {\(\cdots\)};
  \node[operator] (nextsum) at (8.00,-0.05) {\(+\)};
  \node[observed] (rewardnext) at (9.20,-0.05) {\(R_{t+1}\)};
  \draw[rewardedge] (x12.east) to[bend left=8] (nexttermother.west);
  \draw[rewardedge] (x12.east) to[bend right=7] (nextterm.north west);
  \draw[rewardedge] (x13.east) to[bend right=7] (nextterm.south west);
  \draw[rewardedge] (nexttermother.east) to[bend left=7] (nextsum.north west);
  \draw[rewardedge] (nextterm.east) to[bend right=7] (nextsum.south west);
  \draw[rewardedge] (otherterms.north) -- (nextsum.south);
  \draw[rewardedge] (nextsum) -- (rewardnext);

\end{tikzpicture}%
}
\caption{IFactor's state factorization augmented with additive reward
scopes. In the state superscripts, \(a\) means directly controllable by the
previous action, \(\bar a\) means not directly controllable, \(r\) means
reward-relevant, and \(\bar r\) means reward-irrelevant. The black, red, and gray edge meanings
follow the IFactor construction of \citet{liu2023}. At each time, the aggregate reward is expanded into deterministic additive term
nodes for each $H$ scope and a deterministic \(+\) node:
\(R_k=\sum_{H\in\mathcal H_k}r_{k,H}(S_{k,H},A_k)\).
For each additive term node, \(A_k\) is a parent only when that term is nonconstant in
the action argument. 
The highlighted example shows how an uncontrollable but reward-relevant
coordinate can nevertheless be decision-relevant: a descendant of
\(S_t^{\bar a r}\) and a descendant of \(A_t\) enter the same future reward
scope.}
\label{fig:ifactor-reward-scopes}
\end{figure}
\FloatBarrier

\paragraph{Graphical characterization.}
For a node or node set \(U\), let \(\operatorname{De}^{\pi}(U)\) denote its
descendants in the policy-induced augmented graph, including \(U\) itself.
We introduce a graphical criterion to identify the decision-centered coordinate support. A current-state coordinate satisfies the criterion when some current or future additive reward term is a common descendant of that coordinate and the current action. For example, if
\(S_t^j\) affects \(S_{t+1}^p\), \(A_t\) affects \(S_{t+1}^q\), and a later
reward contains \(r_H(S_{t+1}^p,S_{t+1}^q, A_{t+1})\), then the state and action
consequences enter the same reward term. If the reward instead separates as
\(r_{H_1}(S_{t+1}^p,A_{t+1})+r_{H_2}(S_{t+1}^q,A_{t+1})\), they enter different terms.

This graphical criterion generally recovers an outer support; in order to assert that it equals the decision-centered coordinate support, we need to assume a no-cancellation condition on reward contrasts, so that no reward components cancel each other due to their \textit{functional form} without this surfacing in the augmented graph. For example, we need to exclude edge cases where after averaging over future trajectories, state-dependent contrasts from different current or future reward terms cancel in the total Difference-of-(Q). To make this formal, we define the contribution of reward
term \((k,H)\) to the action contrast by
\begin{equation}
 d_{t,k,H}^{\pi,a,a_0}(s)
 :=
 \gamma^{k-t}
 \left\{
\E_{\pi}[r_{k,H}(S_{k,H},A_k)\mid S_t=s,\ A_t=a]
-
 \E_{\pi}[r_{k,H}(S_{k,H},A_k)\mid S_t=s,\ A_t=a_0]
 \right\}.
\label{eq:termwise-contrast}
\end{equation}
The decision-centered coordinate support is obtained by combining this graphical criterion with a graphical \textit{faithfulness}, no-cancellation condition on the contrast. 

\begin{proposition}[Decision-centered coordinate support characterization]
\label{prop:reward-scope-characterization}
Assume a Markov structural model with an action node at
each stage and additive rewards, as in
\Cref{eq:additive-reward-scopes}. Then
\begin{equation}
K_{\Delta,t}
\subseteq
\left\{
 j\in[d]:
 \begin{array}{l}
 \text{there exist }
 \pi\in\Pi_{t+1:T},\
 k\in\{t,\ldots,T\},\
 H\in\mathcal H_k
 \\[1mm]
 \text{such that }
 r_{k,H}
 \in
 \operatorname{De}^{\pi}(S_t^j)
 \cap
 \operatorname{De}^{\pi}(A_t)
\end{array}
\right\}.
\label{eq:graphical-decision-support}
\end{equation}
If the following faithfulness condition holds, the inclusion is an equality:
for every \(j\) satisfying the above graphical criterion, there exist
\(\pi\in\Pi_{t+1:T}\), \((a,a_0)\in\mathcal A_t^{\mathrm{cmp}}\), and an $r_{k,H}$ term such that both
\(d_{t,k,H}^{\pi,a,a_0}(\cdot)\) and
\(\blipadv_t^\pi(\cdot;a,a_0)\) are nonconstant in \(s_j\).
\end{proposition}
Consequently, every coordinate that fails the graphical condition can be
omitted from each required difference-of-$Q$ and \(K_{\Delta,t}^c\) is the maximal set of raw state
coordinates that can be omitted. Note that faithfulness assumptions are classical in causal discovery to draw conclusions from graphical structure; it is a mild technical assumption \citep{spirtes2001causation}. For policy optimization, we recommend applying
the characterization backward. \footnote{Once $K_{\Delta,k}$ has been characterized for
$k=t+1,\ldots,T$, restrict the
continuation policies used at stage $t$ to depend at stage $k$ only on
$S_k^{K_{\Delta,k}}$, and characterize $K_{\Delta,t}$ under that
continuation class; the behavior policy and
all nuisance regressions may continue to use the full state.}
\end{revision}

\section{Estimation: Orthogonal Difference-Of-Q Learning}\label{sec-method}

We now introduce orthogonal estimation to reduce variance in estimating the difference-of-$Q$ function, which we can flexibly combine with support recovery to learn abstractions. 

\looseness=-1
\subsection{Policy Evaluation (Identification)}
We first give an overview of the derivation. The arguments are broadly a generalization of the so-called residualized R-learner \citep{nie2021quasi}; \citep{lewis2021double} considers a similar generalization for structural nested mean models without state-dependent heterogeneity. For this section, we discuss the true population $Q,\blipadv,$ etc., functions without decoration. For brevity, we denote $Q$-functions under some policy $\pi$ from time $t+1$ onwards as $Q_t^\pi$.


Our goal is to estimate the difference-of-Q function: 
\begin{equation}
    \blipadv_t^\pi(S_t) :=  Q_t^\pi(S_t, 1) - Q_t^\pi(S_t, 0).
\end{equation} 
\begin{revision}
For a fixed evaluation policy $\pi^e$, define
$V_{t+1}^{\pi^e}(s)
:=
\sum_{a'\in\cA}
\pi_{t+1}^e(a'\mid s)Q_{t+1}^{\pi^e}(s,a')$.
Then $Q_t^{\pi^e}(S_t,A_t)
=
\E[
R_t+\gamma V_{t+1}^{\pi^e}(S_{t+1})
\mid S_t,A_t
]$. By Markovianity, marginalizing this equation over actions gives
\begin{equation}
 m_t^{\pi^e}(S_t)
 :={}
 \E_{\pi^b}\!\left[
 R_t+\gamma V_{t+1}^{\pi^e}(S_{t+1})\mid S_t
 \right]
=Q_t^{\pi^e}(S_t,0)
 +\pi_t^b(1\mid S_t)\blipadv_t^{\pi^e}(S_t),
 \label{eq:diff-m-main}
\end{equation}
and, differencing these, we see that the true Difference-of-$Q$ satisfies the following moment condition:
\begin{equation}
\E[
\{R_t+\gamma V_{t+1}^{\pi^e}(S_{t+1})-m_t^{\pi^e}(S_t)\}
-\{A_t-\pi_t^b(1\mid S_t)\}\blipadv_t^{\pi^e}(S_t)
\mid S_t,A_t
]=0.
\label{eq:diff-moment-main}
\end{equation}
\end{revision}

\begin{revision}
\paragraph{Generalized pseudo-outcomes via $L$-step unrolling}

We can define a family of equivalent pseudo-outcomes based on recursively unrolling continuation values $V^{\pi_e}_{t+1}(S_{t+1})$ for $L\in\{0,\ldots,T-t\}$ steps into rewards and contrasts $\blipadv$.
Define the Bellman residual (where $V_{T+1}(S_{T+1})=0$):
\[
\xi_j^{\pi^e}
:=
R_j+\gamma V_{j+1}^{\pi^e}(S_{j+1})
-Q_j^{\pi^e}(S_j,A_j),
\text{ and } \E[\xi_j^{\pi^e}\mid S_j,A_j]=0
\]

Denote the original $L=0$ pseudo-outcome as $Y_{t,0}^{\pi^e}
:=
R_t+\gamma V_{t+1}^{\pi^e}(S_{t+1})$, satisfying that $\E[Y_{t,0}^{\pi^e}\mid S_{t},A_t] = Q_t^{\pi^e}(S_t,A_t).$
For $L=0,\ldots,T-t-1$, recursively define $Y_{t,L+1}^{\pi^e}
:=
Y_{t,L}^{\pi^e}
+
\gamma^{L+1}\xi_{t+L+1}^{\pi^e}.$ Unrolling this recursion, we obtain
the explicit $L$-unrolled pseudo-outcome:
\begin{align}
Y_{t,L}^{\pi^e}
={}&
Q_t^{\pi^e}(S_t,A_t)
+
\sum_{\ell=0}^{L}
\gamma^\ell\xi_{t+\ell}^{\pi^e}.
\label{eq:innovation-representation}\\
=&
\sum_{\ell=0}^{L}\gamma^\ell R_{t+\ell}
+
\sum_{\ell=1}^{L}\gamma^\ell
\{\pi_{t+\ell}^e(1\mid S_{t+\ell})-A_{t+\ell}\}
\tau_{t+\ell}^{\pi^e}(S_{t+\ell})
+
\gamma^{L+1}
V_{t+L+1}^{\pi^e}(S_{t+L+1}),
\label{eq:constructive-unrolling-representation}
\end{align}
where the second equality follows from the fact that $V_j^{\pi^e}(S_j)
=
Q_j^{\pi^e}(S_j,A_j)
+
\{\pi_j^e(1\mid S_j)-A_j\}
\tau_j^{\pi^e}(S_j)$.
But the clearest choices correspond to $L=0$ or $L=T-t$. The original moment equation of \Cref{eq:diff-moment-main} corresponds to $L=0$.
Full recursive unrolling corresponds to $L=T-t$ which  gives:
\begin{equation*} \textstyle 
Y_{t,T-t}^{\pi^e}
=
\sum_{j=t}^{T}\gamma^{j-t}R_j
+
\sum_{j=t+1}^{T}\gamma^{j-t}
\{\pi_j^e(1\mid S_j)-A_j\}
\tau_j^{\pi^e}(S_j).
\end{equation*}
The estimator with $L=T-t$ is closely related to the recursive substitution in SNMMs \citep{robins2004optimal}, and is therefore a direct state-conditional generalization of \citet{lewis2021double}, similar to \citet{ma2025deepblip}. The estimator with $L=0$ is new to the literature and trades off some orthogonality from future $Q$ estimation for \textit{action-projected orthogonality}, which is expected to be well-controlled in the structured settings motivating our analysis. Later in experiments, we find sizable improvements from the variance reduction at $L=0$. 
This $L$-indexed family of unbiased estimates which depends on the depth of recursive unrolling of rewards vs. value functions is analogous to an off-policy counterpart of tradeoffs between TD(0) and Monte Carlo samples. 

Next we construct the feasible loss function. Let $m_{t,L}^{\pi^e}(s)
=
\E[ Y_{t,L}^{\pi^e}\mid S_t=s]$;
in the population, $ m_{t,L}^{\pi^e}=m_t^{\pi^e}$ but in the sample, $\hat m_{t,L}$ regresses against different pseudo-outcomes.
For each sample size $n$, let $\cG_{t,n}$ denote the stage-$t$
candidate contrast class. In the main text we write $\cG_t$ and
suppress its dependence on $n$; the e-companion retains $n$ explicitly.
We use the same residualized squared loss,
\begin{equation}
\hat\tau_t^{\pi^e}
\in
\argmin_{\tau\in\mathcal G_t}
\sum_{k=1}^{K}\sum_{i\in\mathcal D_k}
\left\{
\hat{Y}_{it,L}^{\pi^e,-k}
-
\hat{m}_{t,L}^{\pi^e,-k}(S_{it})
-
\big(A_{it}-\hat\pi_t^{b,-k}(1\mid S_{it})\big)\tau(S_{it})
\right\}^2,
\label{eq:recursive-unrolled-loss-main}
\end{equation}
where, for each held-out fold $\mathcal D_k$, the entire future contrast sequence used
to construct $\hat{Y}_{t,L}^{\pi^e,-k}$ is estimated on the
complementary trajectories $\mathcal D_{-k}$. After fixing $L$, the main text may
suppress the $L$-index on $\hat\tau_t^{\pi^e}$ where it is clear from context; the constructed outcome and
its regression retain $L$.

\paragraph{Nuisance estimation.}
Given an evaluation policy $\pi^e$, estimate $V_j^{\pi^e}(s)$ by averaging over estimates of $Q_j^{\pi^e}$: $\hat V_j^{\pi^e}(s)
:=
\sum_{a'\in\cA}
\pi_j^e(a'\mid s)\hat Q_j^{\pi^e}(s,a')$. 
For $L=0$, construct
$\hat Y_{t,0}^{\pi^e}
=R_t+\gamma\hat V_{t+1}^{\pi^e}(S_{t+1})$.
For general $L$, construct $\hat Y_{t,L}^{\pi^e}$ as above, using fold-specific estimates of the
required future contrasts and, when $L<T-t$, the terminal continuation
value. Regress the resulting constructed outcome on $S_t$ to obtain
$\hat m_{t,L}^{\pi^e}$. 
The continuation $Q$-function can be estimated using standard methods such
as fitted-$Q$ evaluation
\citep{le2019batch,chakraborty2014dynamic,duan2021risk} or conditional
moment restrictions/GMM \citep{kallus2019double}. The behavior policy can
be estimated by probabilistic classification, although it is often known
by design.



\textbf{Choosing $L$.}
Every valid choice of $L$ is valid, but the variance differs.
\begin{proposition}[Oracle conditional-variance ordering across unrolling depths]
\label{prop:oracle-unrolling-variance}
Fix a stage $t$ and an evaluation policy $\pi^e$, and construct
$Y_{t,L}^{\pi^e}$ using the population future values and contrasts.
\begin{align}
&\operatorname{Var}(
Y_{t,L+1}^{\pi^e}\mid S_t,A_t
)
-
\operatorname{Var}(
Y_{t,L}^{\pi^e}\mid S_t,A_t
)
=
\gamma^{2(L+1)}
\operatorname{Var}(
\xi_{t+L+1}^{\pi^e}\mid S_t,A_t
)
\ge 0.
\label{eq:conditional-unrolling-variance-order}
\end{align}
\end{proposition}
Thus $L=0$ has weakly smallest oracle pseudo-outcome variance, but it relies on estimating the continuation $Q_{t+1}$
model, although only through action-dependent rather than action-common
errors. Full unrolling, $L=T-t$, instead uses a higher-variance return and
recursively estimated future contrasts, but eliminates the terminal
continuation-value model. Thus the endpoints trade off oracle variance against sensitivity to continuation-model error.

\paragraph{On-policy specialization.}
We can specialize our approach to estimate \textit{on-policy} $Q$-functions from offline data where $\pi_e = \pi_b$, in which case, estimation simplifies considerably. Since the observed return-to-go $\sum_{j=t}^T\gamma^{j-t}R_j$ also satisfies that $\E[ \sum_{j=t}^T\gamma^{j-t}R_j \mid S_t, A_t] = Q_t^{\pi^b}(S_t,A_t),$ it is a valid direct regression target for on-policy $Q$-function estimation, hence $\blipadv$ via differencing. When the behavior probabilities are known, we can further use them to construct control variates.
For a held-out fold $\cD_k$, condition on $\cD_{-k}$.  Let
$h^{-k}=(h_{t+1}^{-k},\ldots,h_T^{-k})$ be control variates, any square-integrable functions
estimated on $\cD_{-k}$, and define
\begin{equation}
Y_t^{b, \mathrm{cv}}(h):=\sum_{j=t}^T \gamma^{j-t} R_j+\sum_{j=t+1}^T \gamma^{j-t}\{\pi_j^b\left(1 \mid S_j\right)-A_j\} h_j\left(S_j\right)
\label{eq:cv-family}
\end{equation}

\begin{proposition}[Behavior-policy control variates]
\label{prop:behavior-policy-control-variates}
Fix \(t\), and let
\(h=(h_{t+1},\ldots,h_T)\) be a sequence of control-variate functions.  Suppose the behavior probabilities are known
and satisfy overlap.  Conditional on the sample used to construct \(h\),
the following statements hold.

\begin{enumerate}
\item[(i)]
For every \(h\), $\E[
Y_t^{b,\mathrm{cv}}(h)
\mid 
S_t,A_t
]
=
Q_t^{\pi^b}(S_t,A_t)$.

\item[(ii)]
For every fixed square-integrable function \(m\), the population loss
\[
\E[
(
Y_t^{b,\mathrm{cv}}(h)-m(S_t)
-\{A_t-\pi_t^b(1\mid S_t)\}f(S_t)
)^{2}
]
\]
is uniquely minimized at $f=\blipadv_t^{\pi^b}$. Thus \(m\) and \(h\) may affect precision but do not change the
population target.

\item[(iii)]
For every fixed \(f\) and \(h\), the loss in part~\textup{(ii)} is
minimized over square-integrable \(m\) by $m_t^*=V_t^{\pi^b}$.
Moreover, the optimal control variate
choice  $h_j^*=\blipadv_j^{\pi^b},
\;\; j=t+1,\ldots,T$,
weakly minimizes the conditional pseudo-outcome variance:
\[
\operatorname{Var}(
Y_t^{b,\mathrm{cv}}(h^*)
\mid 
S_t,A_t)
\le
\operatorname{Var}
(Y_t^{b,\mathrm{cv}}(h)
\mid 
S_t,A_t)
\qquad\text{a.s.}
\]
\end{enumerate}
\end{proposition}

Therefore, our method, when specialized to the on-policy setting where $\pi_e = \pi_b$, can be instantiated more directly as a static $R$-learner with pseudo-outcome rewards $\sum_{t=1}^T \gamma^{t-1} R_t$ \citep{nie2021quasi}. The on-policy setting has received increasing interest in studies of Markovian interference \citep{farias2022markovian,jia2026experimentation}. Our estimator family includes this special case and improves estimation of their averaged difference-of-$Q$ estimand $\E[Q_1^{\pi_b}(S_1,1)-Q_1^{\pi_b}(S_1,0)].$ Note that Appendix D.1 of \citet{jia2026experimentation} considers a similar approach; their approach conducts pseudo-outcome regression on the AIPW-style score while we develop an R-learner style estimation strategy.

\end{revision}
\textbf{Extension to multiple actions.}
So far we presented the method with $\mathcal{A}=\{0,1\}$ for simplicity, but all methods in this paper will extend to the multi-action case. We give the approach discussed in \citep{nie2021quasi}. For multiple actions, fix an arbitrary baseline choice $a_0 \in \mathcal{A}$, and for $a \in \mathcal{A}\setminus\{a_0\},$ define $\blipadv_{t}^\pi(s,a) := \blipadv_{a-a_0,t}^\pi(s) = {Q^\pi_t(s,a) -Q^\pi_t(s,a_0)}.$ For $a\in \mathcal{A},$ let $\pi^b(a\mid S_t)=P(A_t=a\mid S_t)$.  
    Then the analogous loss function is, for any $L\in\{0,\ldots,T-t\}$, where $\langle \cdot , \cdot \rangle$ is the dot product so that $\langle \Vec{A} - \Vec{\pi}^b(S_t), \blipadv_{t}^\pi(S_t)
    \rangle =
    {\sum_{a\in \mathcal{A}\setminus\{a_0\}} \{(\mathbb I[A_t=a]- \pi^b(a\mid S_t) ) \blipadv_{a,t}^\pi(S_t) \}}$:

 $\Vec{\blipadv}_t (\cdot) \in\argmin_\blipadv \{ 
    \E[ 
    ( \{     Y_{t,L}^{\pi^e}
    - m_{t,L}^{\pi^e}(S_t)\}
   - \langle \Vec{A} - \Vec{\pi}^b(S_t), \Vec{\blipadv}_{t}^{\pi^e}(S_t)
    \rangle
    )^2 
    ]
    \}$.

Again, so far we have discussed identification assuming the true $Q,m, \pi^b$ functions, etc. Next we discuss feasible estimation, and outside of this section we refer to the population-level true nuisance functions as $Q^{\pi},m^{\pi}, \pi^b, \blipadv^{\pi}$.



\textbf{Cross-fitting and Policy Optimization:} 
We also introduce cross-fitting for policy evaluation and optimization. We split the dataset $\mathcal{D}$ into a fixed number $K$ of approximately balanced folds, with $|\mathcal D_k|\asymp n/K$ uniformly over $k$.
We preserve trajectories by randomizing over trajectory index $i$ and define
$\mathcal{D}_{-k} := \bigcup_{k'\in[K]\setminus\{k\}}\mathcal{D}_{k'}$.
The two endpoints cross-fit differently. At $L=0$, for each held-out fold $\mathcal{D}_k$ we
estimate the behavior policy $\hat\pi_t^{b,-k}$ and the future
$Q$ model $\hat Q_{t+1}^{\pi^e,-k}$ on $\mathcal{D}_{-k}$; the constructed
outcome $\hat Y_{t,0}^{\pi^e,-k}$ contains no other estimated future
object. At $L=T-t$, no future $Q$ model is needed, but the constructed
outcome $\hat Y_{t,T-t}^{\pi^e,-k}$ involves future contrasts at every
stage $j>t$: we therefore estimate one entire training-only backward sequence
$\{\widetilde\tau_j^{\pi^e,-k}\}_{j=1}^{T}$ on $\mathcal{D}_{-k}$, so that
the full future sequence entering fold $k$ is itself fold-specific. Under
either scheme the outcome regression $\hat m_{t,L}^{\pi^e,-k}$ is fit on
$\mathcal{D}_{-k}$, and each $\hat\blipadv_t^{\pi^e}$ minimizes the loss in
\cref{eq:recursive-unrolled-loss-main} pooled over held-out folds, proceeding
backward in $t$. Every fold-specific nuisance fit is trained on $\mathcal D_{-k}$
before evaluation on $\mathcal D_k$. Later on, for greedy global policy optimization, we introduce uniform oracle inequalities that establish concentration uniformly over data-dependent optimized $\blipadv$-greedy policies. Throughout, superscript $-k$
means estimated on $\mathcal{D}_{-k}$ and evaluated on $\mathcal{D}_k$.

\subsection{Estimation analysis }\label{sec-analysis}
{We study the improved statistical rates of convergence from orthogonal estimation for policy evaluation. \Cref{thm-policy-evaluation} applies orthogonal statistical learning to our new estimand, which is Neyman orthogonal with respect to current-timestep nuisances and satisfies weaker projected orthogonality with respect to nuisance estimates at future timesteps. Analyzing the sample complexity of policy optimization is more challenging, which we discuss in the next section. } 
Our analysis of the method generally proceeds under the following assumptions.
\begin{assumption}[Independent and identically distributed trajectories]\label{asn-iid-trajs}
The $n$ trajectories are independent and identically distributed under a
fixed behavior-policy sequence
$\pi^b=(\pi_t^b)_{t=1}^T$. The behavior policy may vary with the stage
$t$, but the sequence is fixed across trajectories and is not updated
adaptively using previously collected trajectories.
\end{assumption}

\begin{assumption}[Boundedness]\label{asn-bounded values}
There exist finite constants $B_V$ and $\boundadv$ such that for every $t,s,\pi$, $|V_t^\pi(s)|\le B_V,
|\blipadv_t^\pi(s)|\le \boundadv$, including estimators thereof which may be clippsed. Rewards, $\blipadv$ and its estimators are uniformly bounded or clipped. 
\end{assumption}
\begin{assumption}[Sup-norm concentrability]\label{asn-concentrability}
At each time $t$, denote the marginal state-action distribution under a
policy $\pi$ as $d^{\pi}(s,a)$, with the time index suppressed. There exists
a constant $C_\infty$ such that, for any policy $\pi$, including a
non-stationary policy,
$\textstyle
\forall t,\pi,s,a:\ d_\pi(s,a)/d_{\pi^b}(s,a)\leq C_\infty.
$
\end{assumption}
\begin{assumption}[Overlap]\label{asn-overlap}
There exists $\epsilon>0$ such that $\pi_t^b(a\mid s)\ge\epsilon$
for all $t$, $s$, and $a$.
\end{assumption}
\Cref{asn-concentrability} implies \Cref{asn-overlap} with
$\epsilon=C_\infty^{-1}$: the policy that follows $\pi^b$ before time $t$
and selects $a$ at time $t$ attains marginal ratio $1/\pi_t^b(a\mid s)$.
Overlap also bounds the conditional action variance from below:
$\pi_t^b(1\mid s)\{1-\pi_t^b(1\mid s)\}\ge\epsilon(1-\epsilon)=:\lambda>0$,
the curvature constant used in the strong-convexity analysis.
{\Cref{asn-iid-trajs} can be generalized with standard tools}. 
\Cref{asn-concentrability}, sup-norm concentrability, translates error bounds under one state distribution to another and includes a wide class of MDPs \citep{munos2008finite}. Recent works introduce weaker concentrability, for example with pessimism, we leave this for future work and focus on the key novelty of our work. 

A key assumption for orthogonal estimation is that the product of estimation errors for the nuisance-functions converges quickly enough. 
\begin{assumption}[Current-stage nuisance product rates]
\label{asn-producterrorrates}
Fix $\pi^e$ and an unrolling depth $L$. For each fold $k$, conditional on
$\mathcal D_{-k}$, define
$m_{t,L}^{\pi^e,-k}(s):=\E[\hat Y_{t,L}^{\pi^e,-k}\mid
S_t=s,\mathcal D_{-k}]$ and the current-stage product remainder
\begin{equation}
\rho_{t,L}^{\pi^e,-k}
:=\boundadv^2
\norm{\{\hat\pi_t^{b,-k}-\pi_t^b\}^2}_2
+\boundadv
\norm{
\{\hat\pi_t^{b,-k}-\pi_t^b\}
\{\hat m_{t,L}^{\pi^e,-k}-m_{t,L}^{\pi^e,-k}\}
}_2.
\label{eq:diff-rho-L}
\end{equation}
Suppose uniformly over folds
\begin{equation} \textstyle
\max_{k\leq K}\rho_{t,L}^{\pi^e,-k}
=o_p(n^{-1/2}).
\label{eq:diff-current-products}
\end{equation}
A sufficient condition is that, uniformly over folds, the current
behavior-policy and conditional-mean regressions satisfy
\begin{equation*}
\norm{\hat\pi_t^{b,-k}-\pi_t^b}_4=o_p(n^{-1/4}),
\qquad
\norm{\hat m_{t,L}^{\pi^e,-k}-m_{t,L}^{\pi^e,-k}}_4=o_p(n^{-1/4}).
\end{equation*}
\end{assumption}
Thus we require consistent pseudo-outcome regression. We assume
well-specification to simplify theorem statements, with general
versions in the appendix.
\begin{assumption}[Well-specification of $\blipadv$]
\label{asn-wellspecified}
For every stage $t$, $\blipadv_t^{\pi^e}\in\cG_t$.
\end{assumption}
\Cref{thm-mse-rates-policy-evaluation-apx} allows misspecification through
the explicit approximation term
$\inf_{g\in\cG_t}\norm{g-\blipadv_t^{\pi^e}}_2$.

\looseness=-1
 We establish convergence rates of $\hat\blipadv^\pi$, depending on convergence rates of the nuisance functions. Broadly we follow the analysis of \citep{foster2019orthogonal,lewis2021double} for orthogonal statistical learning. We show the loss function is Neyman-orthogonal in \Cref{apx-orthogonality}. 

\begin{revision}
Some additional error in policy evaluation arises from the estimated future functions that appear in the feasible pseudo-outcome. But crucially, estimating action contrasts results in dependence on the \textit{difference across actions} in the estimation errors of future value functions $V_{t+1}^{\pi^e}$ and/or $\blipadv_t^{\pi^e}.$ Therefore, favorable settings for simpler decision-centered abstractions also lead to favorable dependence on error from future continuation-value estimations.  Define a general contrast operator: 
\begin{equation}
(\cC_{t,j}f)(s)
:=
\E[f(S_j)\mid S_t=s,A_t=1]-\E[f(S_j)\mid S_t=s,A_t=0].
\label{eq:diff-action-contrast-operator}
\end{equation}
Conditional on
$\mathcal{D}_{-k}$, the error due to estimating future continuation values is $b_{t,L}^{\pi^e,-k}:$
{\small
\begin{align}
&\E[ \hat Y_{t,L}^{\pi^e,-k}\mid S_t, A_t =1 ] - \E[ \hat Y_{t,L}^{\pi^e,-k}\mid S_t, A_t =0]=\blipadv_t^{\pi^e}+b_{t,L}^{\pi^e,-k}\notag\\
&
b_{t,L}^{\pi^e,-k}
=
\sum_{j=t+1}^{t+L}\gamma^{j-t}
\cC_{t,j}\!\left[
\{\pi_j^e-\pi_j^b\}
\{\widetilde\tau_j^{\pi^e,-k}-\blipadv_j^{\pi^e}\}
\right]
+
\mathbb I[L<T-t]\gamma^{L+1}
\cC_{t,t+L+1}[\hat V_{t+L+1}^{\pi^e,-k}-V_{t+L+1}^{\pi^e}].
\label{eq:diff-target-shift}
\end{align}
}
For example, $Q_{t+1}$-model error that shifts the transition
 equally under both actions cancels out. 

Next we define the learning-theoretic function classes and local Rademacher complexity that govern the analysis. For fold $k$, let $\ell_{t,L}^{-k}(O;g)$ denote the squared summand in
\eqref{eq:recursive-unrolled-loss-main} for trajectory $O$, and define the conditional pairwise
loss-difference class
\[
\mathcal H_{t,L,k}
:=
\left\{
\ell_{t,L}^{-k}(\mathord\cdot;g)
-\ell_{t,L}^{-k}(\mathord\cdot;g'):
g,g'\in\cG_t
\right\}.
\]
For a one-bounded scalar class $\mathcal F$, define its local Rademacher
complexity by
\[ \textstyle 
\mathscr R_m(\mathcal F;r)
:=
\E_{\epsilon_{1:m},O_{1:m}}
[
\sup_{\substack{f\in\op{star}(\mathcal F)\\ \norm{f}_2\le r}}
\frac1m\sum_{i=1}^m\epsilon_i f(O_i)
],
\qquad
\op{star}(\mathcal F):=\{cf:f\in\mathcal F,\ c\in[0,1]\},
\]
where the $\epsilon_i$ are independent Rademacher signs. A critical radius is
any $r>0$ satisfying $\mathscr R_m(\mathcal F;r)\le r^2$. With fixed-$K$ factors
suppressed, let $r_{t,L,n}$ be a deterministic upper bound on the largest
foldwise conditional critical radius. For any fixed $\delta\in(0,1)$, include
the additional requirement $r_{t,L,n}^2\gtrsim\log(K/\delta)/n$.
A union bound over the $K$ folds then gives failure probability at most
$\delta$. Thus $r_{t,L,n}$ measures the local complexity of the
candidate difference-of-$Q$ class. Under standard bounded-design conditions,
fixed-dimensional linear and polynomial classes have critical radius
$O(n^{-1/2})$. For
growing-dimensional or nonparametric classes, the rate is instead determined
by their localized metric entropy \citep[Chaps.~13--14]{wainwright2019high}. Given our high-level assumptions on product error rates, we state simplified results with ``$\lesssim$'' denoting $=O_p(\cdot)$ with high probability, omitting absolute multiplicative constants but for concentrability (\cref{asn-concentrability}). 

\begin{theorem}[$L_2$-error rates for policy evaluation]
\label{thm-policy-evaluation}
Fix a stage $t$, an evaluation policy $\pi^e$, and an unrolling depth
$L\in\{0,\ldots,T-t\}$. Suppose
\Cref{asn-iid-trajs,asn-bounded values,asn-overlap,asn-producterrorrates,asn-wellspecified}
hold for the corresponding stage-$t$ construction. Use $K$-fold
trajectory cross-fitting, with the pooled exact empirical risk minimizer
over $\cG_t$. Then
\begin{equation}
\norm{
\hat\blipadv_{t,L}^{\pi^e}
-
\blipadv_t^{\pi^e}
}_2
\lesssim
r_{t,L,n}
+n^{-1/2}
+
\max_{k\le K}
\norm{b_{t,L}^{\pi^e,-k}}_2,
\label{eq:diff-generic-main-theorem}
\end{equation}
where $n^{-1/2}$ absorbs the current-stage nuisance-product
remainder and $b_{t,L}^{\pi^e,-k}$ is the target shift.
\end{theorem}
See \Cref{thm-mse-rates-policy-evaluation-apx} for a more detailed statement. For the one-step endpoint $L=0$, \Cref{thm-policy-evaluation} applies
directly, with target shift
\[
b_{t,0}^{\pi^e,-k}
=
\gamma\cC_{t,t+1}
\left[
\hat V_{t+1}^{\pi^e,-k}-V_{t+1}^{\pi^e}
\right].
\]
Thus the one-step estimator depends on projected future-value estimation error,
modulo action-invariant errors. Notably, the projected error $b_{t,L}^{\pi^e,-k}$ depends on the \textit{difference in future $\blipadv$ and $V$ across actions}, so it can plausibly be $o_p(n^{-\frac 12})$ under the same structural conditions of simpler decision-centered abstractions that motivate our work and structure-adaptive estimation, like sparse learning. We give an example of a semiparametric model with \textit{noisy action-independent main effects.}  
\begin{corollary}[Uncontrollable continuation main effects]
\label{cor:diff-main-uncontrollable-continuation}
Suppose the next state admits a partition
$S_{t+1}=(S_{t+1}^{\bar a},S_{t+1}^{a})$, where \(\bar a\) and \(a\)
denote the uncontrollable and controllable blocks, respectively, and
$S_{t+1}^{\bar a}\independent A_t\mid S_t$. Suppose also that the true
and fitted value models follow the additive semiparametric model, $V_{t+1}^{\pi^e}(s)
=V_{t+1}^{\bar a,\pi^e}(s^{\bar a})
+V_{t+1}^{a,\pi^e}(s^{a})$. Then, under
\cref{asn-overlap},
\[
\max_{k\le K}\norm{b_{t,0}^{\pi^e,-k}}_2
\le
\frac{2\gamma}{\sqrt\epsilon}\max_{k\le K}
\norm{\hat V_{t+1}^{a,\pi^e,-k}-V_{t+1}^{a,\pi^e}}_2.
\]
Only the controllable component requires a rate; the uncontrollable main effect
requires none.
\end{corollary}
Full unrolling, $L_t=T-t$, recursively
substitutes future contrasts for future $Q$ and therefore
eliminates the terminal continuation-value fit. Off policy, estimation
error in the future contrasts still propagates through the
policy-difference terms in the target shift. On policy, this remaining
future-function target shift vanishes when the future behavior probabilities
are known, as summarized next.
\begin{corollary}[Reduced future-error dependence when $L=T-t$ and/or $\pi^e = \pi^b$]
\label{cor:full-unrolling-rate}
Fix an evaluation policy $\pi^e$, and use full unrolling at every stage,
so that $L_j=T-j$ at stage $j$. Suppose the conditions of
\Cref{thm-policy-evaluation}. Then, for fixed $T$, simultaneously for $t=1,\ldots,T$,
\begin{equation}
\norm{
\hat\blipadv_{t,T-t}^{\pi^e}
-
\blipadv_t^{\pi^e}
}_2
\lesssim
C(
  T,\gamma,\epsilon,
  \max_{t}
  \norm{\pi_t^e-\pi_t^b}_\infty
  )
  \max_{j=t,\ldots,T}
  \left\{r_{j,T-j,n}+n^{-1/2}\right\}.
\label{eq:full-unrolling-rate-propagation}
\end{equation}
The omitted constant is independent of $n$ and depends only on
$T$, $\gamma$, $\epsilon$, and
$\max_j\norm{\pi_j^e-\pi_j^b}_\infty$. 

Further, suppose $\pi^e=\pi^b$, use
$L=T-t$, and assume that the future behavior probabilities
$\{\pi_j^b(1\mid\cdot):j=t+1,\ldots,T\}$ are known. Then no consistency or
rate condition is required for $\widetilde\tau_j^{\pi^b,-k}$, $j>t$, and the
future-function target shift is identically zero: $b_{t,T-t}^{\pi^b,-k}=0$. Then
\begin{equation}
\norm{
\hat\blipadv_{t,T-t}^{\pi^b}
-
\blipadv_t^{\pi^b}
}_2
\lesssim
C(T,\gamma,\epsilon)
\left\{r_{t,T-t,n}+n^{-1/2}\right\}.
\label{eq:on-policy-full-unrolling-rate}
\end{equation}
\end{corollary}
\end{revision}

\looseness=-1
 Our theorem states that estimation of $\pi_b$ and the pseudo-outcome regressions $m_{t,L}$ needs to be only $n^{-\frac 14}$ convergent (in $L_4$ norm) to guarantee that the product error rate of \Cref{asn-producterrorrates} holds with rate $n^{-\frac 12}$. Naive FQE would require $n^{-\frac 12}$ convergence of $Q$. Consistent estimation is still required for $\pi^b$ and the pseudo-outcome regressions. When $L=0$, the future $Q_{t+1}$ model additionally enters through the target shift $b_{t,0}^{\pi^e,-k}$. Although $L=0$ results in lower-variance loss estimation than $L=T-t$, it is more sensitive to estimation error in future value functions. However, in our experiments we find the substantial variance improvement and weaker future projected error result in better performance than $L=T-t$.

To compare to prior methods, in general, methods based on direct regression such as FQE will inherit sample complexity estimation \textit{rates} from the empirical risk minimization subproblems. For example, \citep{duan2021risk} give a bound with Rademacher complexity on the Bellman error of $C \sum_{t=1}^T \mathcal{R}_n^{q_t}\left(\mathcal{F}_t\right)+O(n^{-1/2})$, where $\mathcal{R}_n^{q_t}\left(\mathcal{F}_t\right)$ is the Rademacher complexity of the function class $\mathcal{F}_t$, and $\mathcal{F}_t$ might be infinite or nonparametric. 
Nonparametric function classes are known to suffer curses of dimensionality, where for large $d$, nonparametric estimation is much slower, $p/(2p+d) \ll \frac 12;$ \citep{stone1980optimal}. Where our results improve is that we can leverage estimation of $\pi_b$ (which may be known in RL) in the orthogonal loss function.

\begin{revision}
\subsection{Screening decision-centered abstractions via sparse statistical learning}
Decision-centered abstractions can be learned by combining standard sparse statistical learning with our new $\blipadv$ estimand and orthogonal estimation. In particular, our loss minimization approach is modular and can support flexible regularizers to screen for the \textit{coordinate support} of $\tau$. 

We give a simple end-to-end analysis based on thresholded LASSO, although our analysis is modular and can be extended to nonlinear selectors. For a fixed timestep $t$,
use $L_t=T-t$ at $\pi^e=\pi^b$, and assume that the future behavior
probabilities used in the fully unrolled action-centering terms are known.
The contemporaneous behavior probability in the residualized loss may still
be estimated under \Cref{asn-producterrorrates}. Suppose the true $\blipadv$ is linear: 
\[
\blipadv_t^{\pi^b}(s)=s^\top\beta_t,
\qquad
K_{\Delta,t}=\operatorname{supp}(\beta_t).
\]
Let $\widehat{\mathcal L}_{t,\rm sel}(\beta)$ denote the normalized
cross-fitted loss in \eqref{eq:recursive-unrolled-loss-main}, specialized to
$L=T-t$, $\pi^e=\pi^b$, and the linear contrast
$s\mapsto s^\top\beta$. Thus, every fold-specific nuisance entering
$\widehat{\mathcal L}_{t,\rm sel}$ is trained on $\mathcal D_{-k}$
before evaluation on $\mathcal D_k$. We augment our orthogonal loss with the LASSO $\ell_1$ norm regularizer. Define the LASSO coefficient $\hat\beta_t$ and the estimated support $\hat K_{\Delta,t}$, obtained by thresholding at $v_{t,n}:$
\begin{equation*}\textstyle 
\hat\beta_t
\in
\argmin_{\beta\in\mathbb R^d}
\{
    \widehat{\mathcal L}_{t,\rm sel}(\beta)
    +2\lambda_{t,n}\norm{\beta}_1
\},
\qquad
\hat K_{\Delta,t}
=
\{j:|\hat\beta_{t,j}|>v_{t,n}\}.
\end{equation*}

\begin{proposition}[Thresholded-LASSO sure screening]
\label{prop:tl-support-main}
Assume the linear specification $\blipadv_t^{\pi^b}(s)=s^\top\beta_t, 
K_{\Delta,t}=\operatorname{supp}(\beta_t)$,
and suppose that:
\begin{enumerate}[label=(\roman*),leftmargin=*,nosep]
\item\label{eq:score-main} (Score bound) $\norm{
    \nabla\widehat{\mathcal L}_{t,\rm sel}(\beta_t)
}_\infty
\le \lambda_{t,n}$.
\item\label{eq:re-main} (Restricted-eigenvalue condition) For some $\kappa_t>0$ and every $u\in\mathbb R^d$ such that $\norm{u_{K_{\Delta,t}^c}}_1
\le
3\norm{u_{K_{\Delta,t}}}_1$,
$\frac12
u^\top
\nabla^2\widehat{\mathcal L}_{t,\rm sel}
u
\ge
\kappa_t\norm{u}_2^2.$
\item\label{asn-lasso-betamin} (Threshold calibration and beta-min condition) 

$v_{t,n}
\ge
3\kappa_t^{-1}\lambda_{t,n}$, and 
$\min_{j\in K_{\Delta,t}}|\beta_{t,j}|
>
v_{t,n}
+3\kappa_t^{-1}\sqrt{|K_{\Delta,t}|}\lambda_{t,n}$.
\end{enumerate}
Then we have the following coefficient-$\ell_2$ error bound and
sure-screening result:
\[
\norm{\hat\beta_t-\beta_t}_2
\le
3\kappa_t^{-1}\sqrt{\lvert K_{\Delta,t}\rvert}\lambda_{t,n},
\qquad
K_{\Delta,t}\subseteq\hat K_{\Delta,t},
\qquad
|\hat K_{\Delta,t}|\le 2|K_{\Delta,t}|.
\]
If these conditions hold simultaneously over
$t=1,\ldots,T$ with probability at least $1-\delta_{\rm sel,n}$, then
the sure-screening and size bounds hold simultaneously with the same probability.
\end{proposition}

Therefore, thresholded LASSO recovers a valid, possibly nonminimal, decision-centered coordinate abstraction. The selected set may contain
additional coordinates but has cardinality at most twice that of the true
support. The standard restricted-eigenvalue LASSO bound of \citet{bickel2009simultaneous} ensures sufficient curvature to ensure sparse dependence. The beta-min condition requires a large enough underlying signal, and is standard in the literature for preventing false exclusions \citep{meinshausen2009lasso,zhou2009thresholding,zhou2010thresholded,van2011adaptive}. While much of the analysis is similar to that of standard thresholded LASSO, our estimated loss differs from typical squared loss. Again, orthogonality helps: if the state coordinates are standardized so that $\norm{S_{t,j}}_2\le1$, \Cref{lemma-secondorderderivatives} shows it suffices to choose regularization parameter $\lambda$ of the order: 
\begin{equation*}
\lambda_{t,n}
\gtrsim
\sqrt{\frac{\log(dTK/\delta)}{n}}
+
\max_{k\le K}\rho_{t,T-t}^{\pi^b,-k}.
\end{equation*}
The product error assumption (\Cref{asn-producterrorrates}) renders the second term above of lower order, $o_p(n^{-1/2})$, so that we maintain the usual LASSO penalty and support-recovery rate. 

\begin{remark}[Horizon-union screening]
Our practical recommendation is to implement a horizon-union screening rule, inspired by the sure-screening objective of retaining all relevant
coordinates while permitting false inclusions \citep{fan2008sure}. Under time-homogeneous dynamics, we form
$\hat K_{\Delta}^{\mathrm{union}}=\bigcup_{t=1}^T\hat K_{\Delta,t}$ and use
this common set to refit every stage on a sample independent of support
selection. On the simultaneous
sure-screening event in \Cref{prop:tl-support-main},
$K_{\Delta,t}\subseteq\hat K_{\Delta}^{\mathrm{union}}$.
\end{remark}

\paragraph{Nonlinear selectors.}
We can combine our thresholding approach with coordinate selectors designed for nonlinear models, whether via thresholding nonlinear derivatives \citep{he2021efficient}, single-index models, or modern black-box deep representation learning \citep{liu2023}. 
One illustrative nonlinear option is derivative-thresholded KRR, which screens
coordinates using squared partial derivatives of a ridge-penalized RKHS
regression \citep{he2021efficient}. In the appendix (\Cref{apx-experiments-mi}), we illustrate extensions to deep reinforcement learning with mutual information regularization.

With a sure-screened support of controlled size, we now combine
our estimation results to provide \textit{end-to-end} guarantees on learning
$\blipadv$ with
constant-factor dependence on the coordinate support's cardinality rather
than the ambient dimension.

\begin{corollary}[Adaptation to the decision-centered dimension]
\label{cor:dimension-adaptation}
Fix $t$, $\pi^e=\pi^b$, and let
$K_{\Delta,t}$ be the support of
$\blipadv_t^{\pi^b}$.
Estimate $\hat K_{\Delta,t}$ on one sample split and refit the
Difference-of-$Q$ loss on an independent split. Suppose
\begin{equation*}
P\!\left(
K_{\Delta,t}\nsubseteq\hat K_{\Delta,t}
\ \text{or}\
|\hat K_{\Delta,t}|>2|K_{\Delta,t}|
\right)
\le \delta_{\mathrm{sel},n}.
\end{equation*}
Let $r_{t,T-t,n}(2\lvert K_{\Delta,t}\rvert,\delta)$ be a valid
critical-radius bound for each fixed refit class. 
Suppose the conditions of
\Cref{thm-policy-evaluation} hold for every fixed-support refit used below.
Then, where $C$ depends only on the constants in
\Cref{thm-policy-evaluation},
\begin{equation*}
P\left(
  \norm{\hat\blipadv_t^{\,\mathrm{refit}}
       -\blipadv_t^{\pi^b}}_2
 \le C\,r_{t,T-t,n}(2\lvert K_{\Delta,t}\rvert,\delta)
\right)
\ge 1-\delta_{\mathrm{sel},n}-\delta.
\end{equation*}
\end{corollary}

For example, for a regular linear refit,
$r_{t,T-t,n}(2\lvert K_{\Delta,t}\rvert,\delta)\asymp
\sqrt{\{2\lvert K_{\Delta,t}\rvert+\log(1/\delta)\}/n}$, yielding
$\sqrt{\lvert K_{\Delta,t}\rvert/n}$ rather than $\sqrt{d/n}$ dependence,
up to constants and logarithmic terms.

\end{revision}

\section{Policy optimization}\label{sec-policy-opt}
Next, we discuss leveraging $\blipadv$ estimation for local improvements of the behavior policy, as well as global policy optimization under stronger assumptions.

\begin{algorithm*}[t!]
\caption{Cross-fitted backward-greedy Difference-of-$Q$ optimization
($L=0$)}\label{alg-dyn-blipadv-optimization}
\begingroup
\makeatletter
\setboolean{ALC@noend}{true}
\makeatother
\begin{algorithmic}[1]
\REQUIRE Trajectory folds $\{\cD_k\}_{k=1}^K$ and contrast classes
$\{\cG_t\}_{t=1}^T$.
\FOR{$k=1,\ldots,K$}
\STATE On $\cD_{-k}$, estimate the behavior probabilities
$\hat\pi_t^{b,-k}$ for all $t$.
\ENDFOR
\STATE Set $\hatunderpi{T+1}:=\emptyset$ and, for every $k$,
$\hat Q_{T+1}^{\hatunderpi{T+1},-k}\equiv0$ and
$\hat V_{T+1}^{\hatunderpi{T+1},-k}\equiv0$.
\FOR{$t=T,T-1,\ldots,1$}
\FOR{$k=1,\ldots,K$}
\STATE For $t<T$, fit
$\hat Q_{t+1}^{\hatunderpi{t+1},-k}$ on $\cD_{-k}$ for the frozen suffix
$\hatunderpi{t+1}.$
\STATE Form the one-step outcome
$\hat Y_{it,0}^{\hatunderpi{t+1},-k}
=R_t^i
+\gamma
\hat V_{t+1}^{\hatunderpi{t+1},-k}(S_{t+1}^i)$,
and regress it on $S_t$ in $\cD_{-k}$ to obtain
$\hat m_{t,0}^{\hatunderpi{t+1},-k}$.
\ENDFOR
\STATE Estimate $\hat\blipadv_{t,0}^{\hatunderpi{t+1}}\in \argmin_{g\in\cG_t}
\widehat{\mathcal L}_{D,t,0}^{\hatunderpi{t+1}}(g)$, where the pooled
held-out loss is
\begin{equation*}
\widehat{\mathcal L}_{D,t,0}^{\hatunderpi{t+1}}(g)
:=
\sum_{k=1}^K\sum_{i\in\cD_k}
\Bigl[
\hat Y_{it,0}^{\hatunderpi{t+1},-k}
-\hat m_{t,0}^{\hatunderpi{t+1},-k}(S_t^i)
-\{A_t^i-\hat\pi_t^{b,-k}(1\mid S_t^i)\}g(S_t^i)
\Bigr]^2.
\end{equation*}
\STATE Set
$\hat\pi_t(1\mid s)
=\mathbb I[\hat\blipadv_{t,0}^{\hatunderpi{t+1}}(s)>0]$
and define $\hat\pi_{t:T}:=(\hat\pi_t,\hatunderpi{t+1})$.
\ENDFOR
\STATE Return $\hat\pi_{1:T}$.
\end{algorithmic}
\endgroup
\end{algorithm*}
\begin{revision}
\subsection{Local policy improvement}\label{sec-behavior-centered-trust-region}

Advantage functions (which can be recovered from our difference-of-$Q$ functions) are widely used in RL algorithms. Our improved estimation can therefore be used as a drop-in replacement where advantage functions are used in RL, such as \textit{local policy improvement}: optimizing $\blipadv^{\pi_b}_t$ and taking a local step, as in conservative policy iteration \citep{kakade2002approximately}. In particular, improving the behavior policy $\pi^b$ has two estimation benefits: we do not need to make strong concentrability assumptions, and, when the future behavior probabilities are known, our estimator with $L=T-t$ at $\pi^b$ has no future-function target shift. This local step is a greedy proposal $\pi_t^{\mathrm{gr}}(1\mid s)
:=\mathbb I[\blipadv_t^{\pi^b}(s)>0]$, and define the $w$-mixture policy:
\begin{equation}
\pi_t^w(\cdot\mid s)
:=(1-w)\pi_t^b(\cdot\mid s)
+w\pi_t^{\mathrm{gr}}(\cdot\mid s),
\qquad
\Delta_t^{\mathrm{gr}}(s)
:=\pi_t^{\mathrm{gr}}(1\mid s)-\pi_t^b(1\mid s),
\label{eqn-behavior-centered-mixture}
\end{equation}
where $w\in[0,1]$. A standard performance difference lemma reveals the role of $\blipadv$ in local policy improvement.


\begin{proposition}[Finite-step local value bound around the behavior policy]
\label{prop-behavior-centered-improvement}
Suppose $\norm{\blipadv_t^{\pi^b}}_\infty\leq\boundadv$ for every $t$.
Then, for every $w\in[0,1]$,
\begin{equation}
\E[V_1^{\pi^w}(S_1)]-\E[V_1^{\pi^b}(S_1)]
\geq w
\sum_{t=1}^T\gamma^{t-1}
\E_{d_t^{\pi^b}}
\left[\Delta_t^{\mathrm{gr}}(S_t)
\blipadv_t^{\pi^b}(S_t)\right]
-2\boundadv w^2
\sum_{t=1}^T\gamma^{t-1}(t-1).
\label{eqn-behavior-centered-improvement}
\end{equation}
If
$\sum_{t=1}^T\gamma^{t-1}\E_{d_t^{\pi^b}}
[\Delta_t^{\mathrm{gr}}(S_t)\blipadv_t^{\pi^b}(S_t)]>0$, there is a
$\bar w>0$ such that every $w\in(0,\bar w]$ improves $\pi^b$.
\end{proposition}

For the greedy proposal,
  $\Delta_t^{\mathrm{gr}}(s)\blipadv_t^{\pi^b}
  (s)\geq0$ at every state.
For sufficiently small $w>0,$ the first-order linear
  gain dominates the
  quadratic state-distribution-shift penalty.
\Cref{prop-behavior-centered-improvement} motivates local policy improvement from $\pi_b$: for small $w$, we are guaranteed improvement \textit{without} making strong concentrability assumptions required for full policy optimization. And, evaluating $\pi^e=\pi^b$ is lower-variance, as in \Cref{prop:behavior-policy-control-variates}.
\end{revision}

\subsection{Full greedy optimization}
\label{sec-global-policy}

The sequential loss also yields a backward-greedy policy-optimization
procedure. The displayed implementation,
\Cref{alg-dyn-blipadv-optimization}, uses the one-step endpoint $L=0$ at each
stage, which can easily be replaced by $L=T-t$. Both procedures estimate
the action-value contrast for the already constructed continuation suffix and
then take the greedy action.

\paragraph{Cross-fitting under a data-dependent continuation policy.}
For the global algorithm, $\cG_j$ denotes the stage-$j$ candidate $\blipadv$ function
class. For $t<T$, define the
greedy-suffix class
\[
\Pi_{t+1:T}^{\mathrm{gr}}
:=
\left\{
\pi_{t+1:T}:
\pi_j(1\mid s)=\mathbb I[g_j(s)>0]
\text{ for some }g_j\in\cG_j,
\quad j=t+1,\ldots,T
\right\},
\]
and set $\Pi_{T+1:T}^{\mathrm{gr}}:=\{\varnothing\}$. 
Algorithm~\ref{alg-dyn-blipadv-optimization} produces optimized policy suffixes in this class. In the appendix, 
\Cref{prop:uniform-one-step} gives sufficient conditions for a uniform oracle
inequality over $\Pi_{t+1:T}^{\mathrm{gr}}$ that enables \Cref{alg-dyn-blipadv-optimization} to use ordinary
trajectory-level $K$-fold cross-fitting at every stage and reuse the same
folds across stages. 

\textbf{Margin assumption transfers convergence of $\blipadv_t$ to policy value. }Convergence of $\blipadv_t$ implies convergence in policy value. We quantify this with the \textit{margin} assumption, a low-noise condition that quantifies the gap between regions of different optimal action \citep{tsybakov2004optimal}, often used to relate plug-in estimation error  to decision risk. 
\begin{assumption}[Margin on observational distribution]\label{asn-margin-observational-pib}
Let $Q^{*}_t(s, \pi^*)$ be the optimal $Q$ function at the optimal action, and $a^{\prime}$ the second-best option, $a^{\prime} \in \mathcal{A}\setminus \arg \max _a Q^{*}_t(s, a).$
Assume there exist constants $\alpha\geq0$ (the margin exponent) and
$\delta_0>0$ such that
$$
\begin{aligned}
\textstyle
P_{ 
}
\left( Q^{*}_t(S_t, \pi^*)
-Q^{*}_t\left(S_t, a^{\prime}\right) \leq \epsilon\right)
\leq (\epsilon/\delta_0)^\alpha
, \forall t \in 1, \dots, T
\end{aligned}
$$
\end{assumption}
The above probability is over the observational data distribution, $S_t \sim \mu_{\pi^b_t}$.
Note that \citep{hu2024fast} establishes margin constants for linear and tabular MDPs; {$\alpha = \infty$ for tabular MDPs, $\alpha=1$ for linear MDPs where $Q^*$ is linear, and for nonlinear $Q^*$ under structural assumptions}. A uniform gap between optimal and next-optimal action implies $\alpha=\infty$. {The margin assumption translates advantage-function convergence rates to the policy value, summarized in the following Lemma.}

\begin{lemma}[Contrast error to decision regret under a margin condition]
\label{lemma-adverror-to-value}
Fix a stage $t$ and binary actions. Suppose \Cref{asn-margin-observational-pib} holds with
exponent $\alpha\geq0$. For any event $\mathcal E$ on which
$\norm{\hat\blipadv_t-\blipadv_t}_2\leq\epsilon_t$, the following bounds hold
on the same event:
\begin{equation*}
\E\!\left[
Q_t^*(S_t,\pi_t^*(S_t))-Q_t^*(S_t,\hat\pi_t(S_t))
\right]
\lesssim
\epsilon_t^{\frac{2+2\alpha}{2+\alpha}}.
\end{equation*}
\begin{equation*}
\norm{
Q_t^*(S_t,\pi_t^*(S_t))-Q_t^*(S_t,\hat\pi_t(S_t))
}_2
\leq\epsilon_t.
\end{equation*}
\end{lemma}
The first conclusion is applied stage by stage in the final
performance-difference sum. The second conclusion is the ordinary $L_2$
bound used in the backward induction.

Next we study policy optimization. {\Cref{lemma-adverror-to-value} relies on convergence of estimated difference-of-Q under the estimated-optimal policy to the true difference-of-Q under the optimal policy, while earlier guarantees in \Cref{thm-policy-evaluation} just grant convergence for the estimated-optimal policy. This is the key technical difficulty that we overcome with uniform estimation of $\blipadv^{\pi}$ over the greedy-policy suffixes,  induction/concentrability and \Cref{asn-margin-observational-pib}. We state a simplified version of the full theorem in the main text; the full statement is in \Cref{apx-full-statements}.}

For the displayed one-step algorithm, set $L_t=0$. For its full-unrolling
analogue, set $L_t=T-t$.
\begin{assumption}[Uniform endpoint rate over greedy suffixes]
\label{asn-uniform-global}
There is a deterministic sequence $\varepsilon_n\downarrow0$ such that
\[
\max_{1\le t\le T}
\sup_{\pi_{t+1:T}\in\Pi_{t+1:T}^{\mathrm{gr}}}
\left\norm{
\hat\blipadv_{t,L_t}^{\pi_{t+1:T}}
-\blipadv_t^{\pi_{t+1:T}}
\right}_{2,d_t^{\pi^b}}
=O_p(\varepsilon_n).
\]
\end{assumption}

\begin{corollary}[Root-$n$ uniform rate for fixed-complexity threshold policies]
\label{cor:uniform-endpoint-rootn}
Suppose 
\Cref{asn-sequential-unconf,asn-iid-trajs,asn-bounded values,asn-overlap}
hold; and \Cref{asn-producterrorrates,asn-wellspecified} hold uniformly over
stages, folds, and greedy future policies. Suppose also that
the policy, contrast, and policy-indexed nuisance classes satisfy the
fixed-complexity covering-number and Lipschitz-composition conditions in
\Cref{rem:pdim-global}. When $L_t=T-t$,
\Cref{asn-uniform-global} holds with $\varepsilon_n=\widetilde O(n^{-1/2}).$ When $L_t=0$, the same is true under the additional assumption
\[
\max_{1\leq t\leq T}
\sup_{\pi_{t+1:T}\in\Pi_{t+1:T}^{\mathrm{gr}}}
\max_{k\leq K}\norm{b_{t,0}^{\pi_{t+1:T},-k}}_2
=\widetilde O_p(n^{-1/2}).
\]
\end{corollary}

\begin{theorem}[Policy optimization bound]\label{thm-policy-optimization}
Suppose
\Cref{asn-sequential-unconf,asn-iid-trajs,asn-bounded values,asn-concentrability,asn-margin-observational-pib,asn-uniform-global}
hold, with margin exponent $\alpha\geq0$. Apply the backward-greedy procedure
with the endpoint schedule $\{L_t\}_{t=1}^T$ fixed above. Then, for binary
actions,
\ifonecolumnequations
\begin{equation*}
\norm{
\hat\blipadv_{t,L_t}^{\hat\pi_{t+1:T}}
-\blipadv_t^{\pi^*_{t+1:T}}
}_2
\lesssim \varepsilon_n,
\qquad
\abs{\E[V_1^{\pi^*}(S_1)-V_1^{\hat\pi_{\hat\blipadv}}(S_1)]}
\lesssim C_\infty \varepsilon_n^{\frac{2+2\alpha}{2+\alpha}}.
\end{equation*}
\else
\begin{align*}
\norm{
\hat\blipadv_{t,L_t}^{\hat\pi_{t+1:T}}
-\blipadv_t^{\pi^*_{t+1:T}}
}_2
&\lesssim \varepsilon_n,\\
\abs{\E[V_1^{\pi^*}(S_1)-V_1^{\hat\pi_{\hat\blipadv}}(S_1)]}
&\lesssim
C_\infty \varepsilon_n^{\frac{2+2\alpha}{2+\alpha}}.
\end{align*}
\fi
In particular, if $\varepsilon_n=O(n^{-1/2})$, then
\begin{equation*}
\E[V_1^{\pi^*}(S_1)-V_1^{\hat\pi_{\hat\blipadv}}(S_1)]
=
O_p\!\left(n^{-\frac{1+\alpha}{2+\alpha}}\right).
\end{equation*}
Under the conditions of \Cref{cor:uniform-endpoint-rootn}, the same conclusion
holds up to logarithmic factors:
$\E[V_1^{\pi^*}(S_1)-V_1^{\hat\pi_{\hat\blipadv}}(S_1)]
=\widetilde O_p(n^{-(1+\alpha)/(2+\alpha)})$.

\end{theorem}
The main takeaway is that uniform estimation over the greedy future
policy class yields convergent policy optimization. When the contrast rate is
root-$n$, $\alpha=0$ gives the standard root-$n$ policy-value rate, while
$\alpha>0$ gives a faster rate. To our knowledge, this is a new approach for
establishing policy optimization with policy-dependent nuisance functions. It
combines a margin condition with policy classes induced by a learned
$\hat\tau$ and backward-greedy optimization.

 %
\section{Experiments}\label{sec-experiments}


\begin{figure*}[t!]\label{fig:interacted}
    \centering
\begin{subfigure}[t]{0.33\textwidth}
    \includegraphics[width=\textwidth]{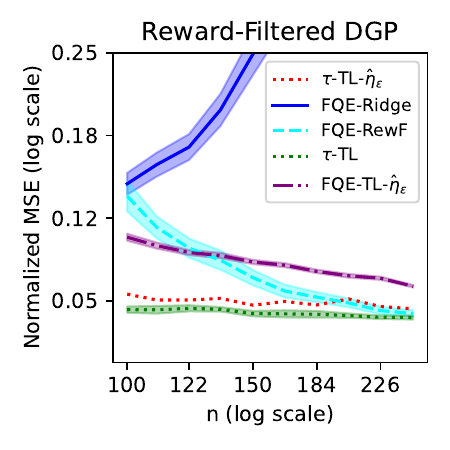}
    \end{subfigure}\begin{subfigure}[t]{0.33\textwidth}
    \includegraphics[width=\textwidth]{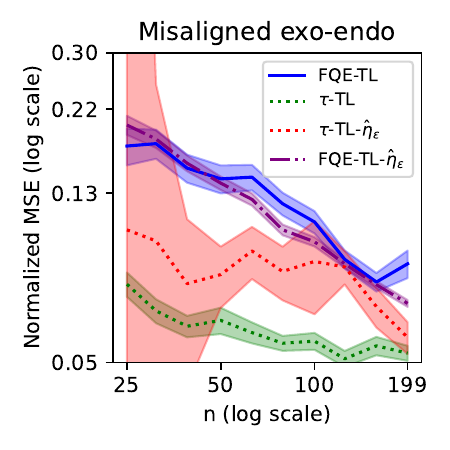}
    \end{subfigure}\begin{subfigure}[t]{0.33\textwidth}
    \includegraphics[width=\textwidth]{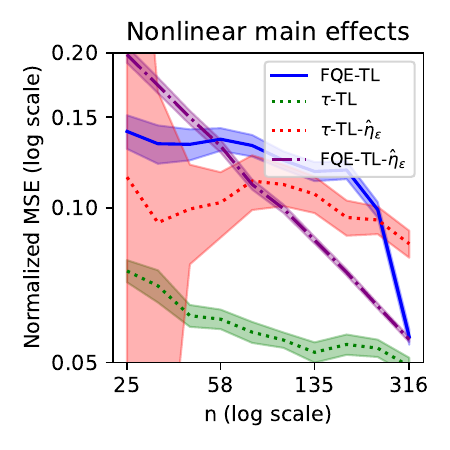}
    \end{subfigure}\begin{subfigure}[t]{0.33\textwidth}
    \end{subfigure}
    \caption{Adapting to structure. Interacted setting where $E[M_1 - M_0] = 0.1 \cdot I.$}
    \label{fig:adapting-to-structure}
\end{figure*}

\looseness=-1
\textbf{Adapting to structure in $\tau(s)$. }
 \Cref{fig:adapting-to-structure} illustrates that
  estimating the difference
  of $Q$ adapts across data-generating processes with
  different graphical and
  main-effect structures.  We compare ridge FQE (blue), reward-filtered FQE following \citep{zhou2024rewardrelevancefiltered} (dotted cyan), and
  thresholded-LASSO FQE (blue), with
  our thresholded-LASSO difference-of-$Q$ estimator (dotted green); and a sample-split variant with nuisance predictions
  perturbed at the
  $n^{-1/4}$ rate (dot-dashed purple). Each DGP has a $d=100$ state
  and a sparse
  difference-of-$Q$ function. The figure reports normalized MSE of estimating $\blipadv$
  against the number of
  episodes over 100 replications, using the median for
  the sample-split
  variants. The left panel uses the
  reward-filtered
  factorization of
  \citet{zhou2024rewardrelevancefiltered}, with
  approximately
  15 reward- and transition-relevant coordinates. The
  center panel modifies
  the conditional independences to follow the
  exogenous--endogenous structure
  in \Cref{fig:exoendo} and adds dense rewards, so
  reward filtering is
  misspecified. The right panel adds dense nonlinear
  main effects that are
  disjoint from the sparse difference-of-$Q$ function.
  Exact transition,
  reward, policy, and noise specifications appear in
  \Cref{sec-experimental-details}.

Therefore, difference-of-$Q$ estimation can adapt to favorable decision-centered abstractions, without relying on a priori graphical knowledge about exactly how they arise. In the left panel,
  reward-filtered estimation
  is well specified, whereas ridge FQE is destabilized
  by the full
  high-dimensional state. In the center panel, reward
  filtering is
  misspecified and thresholded-LASSO FQE retains too
  many extraneous
  coordinates, while sparse difference-of-$Q$
  estimation remains stable. The
  right panel shows the same advantage when introducing complex
  nonlinear main effects. 

\begin{revision}
\paragraph{Sparse support recovery in a linear-Gaussian MDP.}
Next we isolate the small-sample benefit of sparse difference-of-$Q$
estimation in a synthetic eight-period linear-Gaussian MDP. The state contains
120 coordinates with dense action-invariant prognostic effects and 30
candidate action-effect modifiers, of which three are active. The behavior
policy assigns probability $1/2$ to each binary action. We compare DiffQ with one-step ($L=0$) estimation with full-Q ridge and the randomized behavior policy. \Cref{fig:hospital-sparse-l0-validation} shows a uniform small-sample
advantage due to more efficient screening, illustrated on the right subpanel, while the horizon union is more robust at small samples for exact recovery. Direct full-Q
estimation continues to spend statistical capacity on dense prognostic
structure, while our method improves. Further details are in
\Cref{app:linear-gaussian-policy-details}.

\begin{figure}[t]
  \centering
  \includegraphics[width=0.98\linewidth]{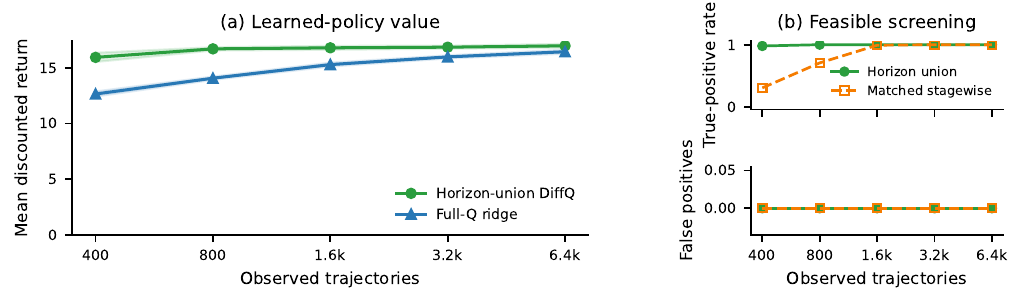}
  \caption{\rev{\textbf{Feasible sparse policy learning in the linear-Gaussian
  experiment.} Panel (a) reports the mean discounted return of horizon-union
  DiffQ and full-Q ridge, with two-sided 95\% Student-$t$ intervals across 32
  replications. Panel (b) reports the mean stage-level true-positive rate
  (top) and mean number of false-positive state modifiers (bottom) for the
  horizon union and the matched stage-specific screens.}}
  \label{fig:hospital-sparse-l0-validation}
\end{figure}

\end{revision}

\begin{revision}
\paragraph{Nonlinear decision-variable screening with derivative-thresholded KRR.}

Next, we illustrate the empirical use of derivative-thresholded KRR as an
alternative nonlinear selector \citep{he2021efficient}. In this sparse
set-up, we find that the $L=0$ estimator achieves greater variance reduction
and projected orthogonality results in better overall performance than the
higher-variance $L=T-t$ estimator.

\begin{figure*}[t]
  \centering
    \includegraphics[width=\textwidth]{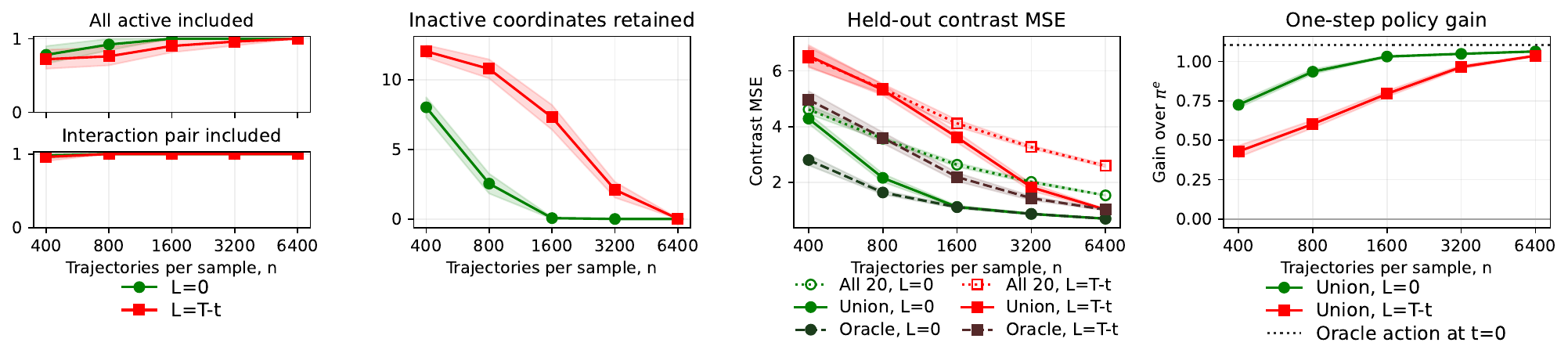}%
  \caption{\textbf{Nonlinear decision-variable screening under fixed
  off-policy evaluation.} From left to right, we compare screening and estimation properties for our methods: active-coordinate inclusion, inactive coordinates retained, MSE for off-policy evaluation, and one-step policy improvement. Green circles use the one-step endpoint $L=0$, while red squares use
  full unrolling $L=T-t$. Each
  replication uses two independent samples, each containing $n$ trajectories:
  one for screening and one for refitting. Standard errors are calculated over $50$
 replications with two-sided $95\%$ Student-$t$ intervals. }
  \label{fig:nonlinear-screening-refit}
\end{figure*}

\Cref{fig:nonlinear-screening-refit} shows the favorable screening properties and estimation benefits of $L=0$. Throughout, we use a union screening rule which includes the union of recovered supports, reflecting asymmetric harms of approximation bias from underinclusion vs. mild losses of efficiency from overinclusion.
The first and second subpanels report true-positive active-coordinate inclusion vs. false positives. The lower variance of $L=0$ also reduces the number of inactive coordinates retained. The third subpanel compares the MSE for $\blipadv$ and includes, for each $L=0$ or $L=T-t$ variant, a baseline using all $d=20$ variables and an unachievable oracle ``skyline'' that knows the true sparse support. The last panel illustrates how better estimation of $\blipadv$ translates into better policy gains via one-step policy improvement at $t=0$ only. The lower-variance $L=0$ method also converges more quickly to its oracle-optimal performance. Further experimental details are in \Cref{app:nonlinear-krr-details}.

\end{revision}

\begin{revision}
\subsection{Improving Estimation Under Markovian interference: Ridesharing}
\label{app:tpg-results}

Although our paper focuses on estimating the difference-of-$Q$ to learn decision-centered abstractions, our improvements in estimation can be of independent interest. Prior works study Markovian interference and showed that on-policy estimation of the difference-of-$Q,$ $\E[Q^{\frac{1}{2}}(S_1,1)-Q^{\frac{1}{2}}(S_1,0)]$ improves upon naive reward differencing \citep{farias2022markovian}, including refinements via truncation for nonstationary settings (truncated policy gradient, or TPG) \citep{johari2025estimation}. We leverage our orthogonal estimation to estimate $\blipadv^{\frac 12}$ directly with lower variance, which we demonstrate by combining improvements from combining our orthogonal estimation with the $k=2$ truncated reward signal and ridesharing experiment in \citet{johari2025estimation}. \Cref{fig:tpg-diffq-learning-curves} demonstrates the results. All estimators achieve the expected
\(n^{-1/2}\) pattern across 200 independent replications, but our estimators improve the constant-factor terms due to variance reduction. We compare to TPG as the baseline (blue) and  consider two
DiffQ variants that use TPG's \(k=2\) truncated reward signal: one without control variates (\(h=0\)) (orange), and the other estimates the population-optimal
control variate \(m_t^{\pi_b}\) (green). The RMSE of estimated-CV DiffQ is 64\% that of DiffQ (\(h=0\))
and 34\% that of TPG.

\paragraph{State-dependent policy learning.}
We next modify the simulator by introducing state-dependent effects to illustrate benefits of one policy improvement step. The treatment effect varies with a few states: cyclic time, the available
fleet fraction, remaining busy time, and a time-by-availability interaction.
The outcome also contains an action-independent main effect on either $p=3$ or 50 correlated variables. DiffQ uses a well-specified depth-three policy-tree class, while direct $Q$ estimation uses the same for each action and fits the contrast to the estimated $Q$. TPG can only guide policy decisions between all-treat and all-control.
The right panel of \Cref{fig:tpg-diffq-learning-curves} shows DiffQ quickly learns the policy improvement. Naively differencing direct $Q$ estimation suffers because the approximation errors for the simple decision trees compound.
\begin{figure}[t]
\centering
\begin{minipage}[t]{0.51\textwidth}
\vspace{0pt}
\centering
\includegraphics[width=\linewidth]{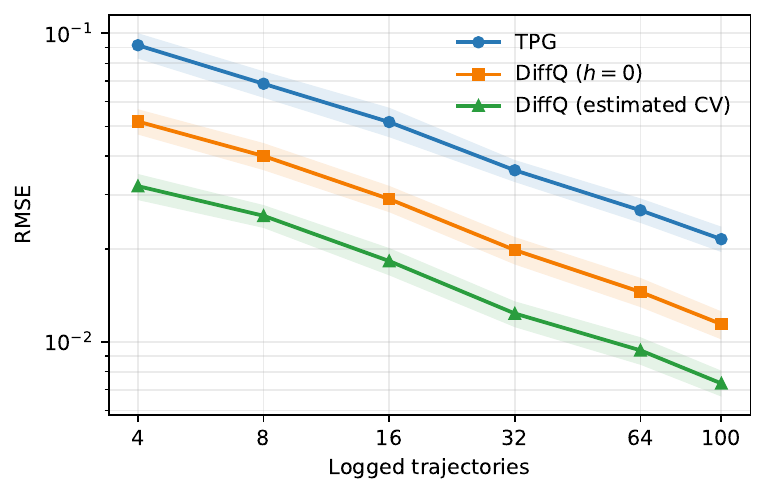}
\end{minipage}\hfill
\begin{minipage}[t]{0.47\textwidth}
\vspace{0pt}
\centering
\includegraphics[width=\linewidth]{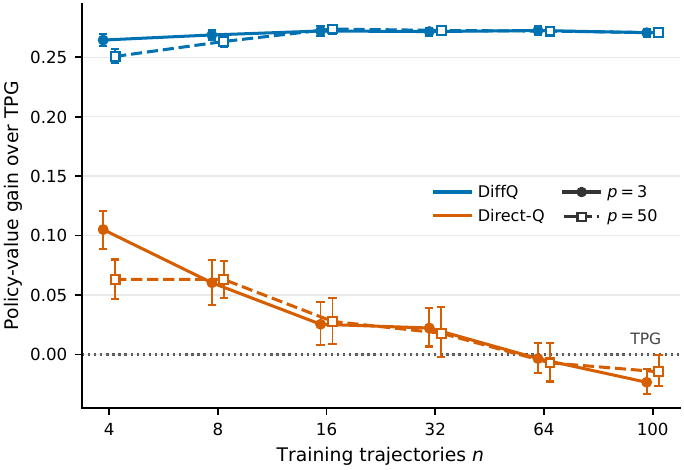}
\end{minipage}
\caption{Improving estimation for Markovian interference; ridesharing simulator. Left: Absolute RMSE for TPG, DiffQ with $h=0$, and DiffQ with
estimated control variates across 200 independent replications. Right:
policy-value gain over the TPG-guided constant-action policy for DiffQ and
direct-Q across 100 independent replications, where $p$ is the number of
action-independent main-effect variables. Error bars in both panels are
paired-bootstrap 95\% confidence intervals.}
\label{fig:tpg-diffq-learning-curves}
\end{figure}

\end{revision}

\textbf{Conclusions and limitations.} We bridged offline RL and causal ML by introducing decision-centered abstractions that preserve the difference-of-$Q$ functions, a dynamic generalization of well-studied heterogeneous treatment effects. We introduced orthogonal estimation that reduces variance. We can improve sequential data-driven decision-making by learning the difference that makes a difference for future rewards, not just about all future state dynamics. 

 \section*{Data and Code Accessibility}

  Data, code and documentation needed to
  reproduce the
  computational results in this paper are provided in a
  separate
  Code and Data Supplement accompanying this submission. Note that the ridesharing experiments use two
  publicly accessible
  New York City--derived input files and the simulator from \citet{johari2025estimation}, which are not redistributed here. No confidential or
  proprietary data were
  used.

\clearpage
\bibliographystyle{informs2014}
\bibliography{for-arxiv}

\ECSwitch
\ECHead{Supplementary Materials}

Appendix table of contents
\begin{itemize}
    \item \Cref{sec-addl-discussion} includes additional discussion on related work. 
    \item \Cref{sec-apx-deferred-fullstatement} includes full statements of some policy evaluation theorems given in the main text, which were simplified for a smoother presentation. 
    \item \Cref{apx-decision-centered-proofs} contains proofs of our characterization of decision-centered abstractions (the results in \Cref{sec-abstractions}).
    \item \Cref{apx-sec-proofs-estimation} contains proofs of estimation and analysis (the results in \Cref{sec-method}).
    \item \Cref{sec-pol-opt} contains proof of policy optimization results (the results in \Cref{sec-policy-opt}).
    \item \Cref{sec-experimental-details} contains additional details about the experiments in \Cref{sec-experiments}.
\end{itemize}

\section{Additional discussion}\label{sec-addl-discussion}

\paragraph{Additional discussion on related work.}

\citet{pan2023learning} note the analogy of the advantage function with causal contrast estimation and derive a Q-function independent estimator, but in the online setting. After preparing an initial version of this paper, we became aware of the recent work of \citet{pan2024skill} in the offline setting. {While their method elegantly avoids estimating future $Q$ functions, it requires a nonconvex constraint on the action-average of advantages, which is computationally and statistically difficult. The motivations are different; the methods are complementary.} Our identification is different and we focus on improved statistical guarantees.

\paragraph{Methods for learning state abstractions and state aggregations}

State aggregations have been widely studied in approximate dynamic programming \citep{van2006performance}, but it remains unclear how best to estimate them. (In either case, our decision-centered abstractions make poor state aggregations --- where one conducts backwards reduction on the reduced decision-centered state itself --- although they can support policy optimization based on learned difference-of-$Q$ functions, as we do in our paper). Learning state abstractions is also nontrivial, often requiring bespoke estimation techniques that embed the Markovian desiderata of state abstractions within the estimation process \citep{bennouna2025learning,kemertas2021towards}. This can attempt to embed difficult and statistically/computationally intractable conditional independence requirements within the estimation process itself \citep{wang2021provably,wang2022causal,wang2022denoised,liu2023}. 

In contrast, our approach is first centered on \textit{changing the estimand} to one whose simple sparsity achieves abstraction desiderata and with rich estimation theory from causal inference and machine learning. Then, we can reduce the problem of learning decision-centered abstractions to the more standard problem of sparsity or support recovery in plain statistical learning, albeit on a nonstandard loss function. Importantly, the structural desiderata then do not enter our empirical loss minimization procedure. 

\paragraph{Minimal and optimal adjustment sets }
A rich literature from causal inference focuses on confounder selection and minimal or efficient adjustment sets \citep{witte2020efficient}. Crucially, minimal decision-centered abstractions are not properties of the nonparametric structural causal model/DAG alone, but require additional functional form information. Our graphical criterion requires augmentation with \textit{additive reward scopes} describing the across-action cancellation of within-timestep rewards. 

\section{Deferred full statements}\label{sec-apx-deferred-fullstatement}
\subsection{Full statements of algorithms}\label{app:recursive-unrolling}
Given a training trajectory set $\cD_{\mathrm{tr}}$, the training-only
backward routine returns a sequence of future-contrast estimates using only
$\cD_{\mathrm{tr}}$. The resulting
functions are treated as fold-specific nuisance fits in the outer procedure.

\begin{algorithm*}[t!]
\caption{Dynamic Residualized Difference-of-Q Evaluation}
\label{alg-dyn-blipadv-evaluation}
\begin{algorithmic}[1]
\REQUIRE Evaluation policy $\pi^e$; trajectory folds
$\{\cD_k\}_{k=1}^K$; contrast classes $\{\cG_{t,n}\}_{t=1}^T$;
endpoint schedule $L_t\in\{0,T-t\}$.
\FOR{$k=1,\ldots,K$}
\STATE Set $\cD_{-k}=\cD\setminus\cD_k$.
\STATE On $\cD_{-k}$, estimate the behavior probabilities
$\{\hat\pi_t^{b,-k}\}_{t=1}^T$.
\IF{the endpoint schedule contains at least one full-unrolling stage}
\STATE Use only $\cD_{-k}$ to run the training-only backward routine and
obtain $\{\widetilde\tau_j^{\pi^e,-k}\}_{j=1}^T$.
\ENDIF
\FOR{$t=T,T-1,\ldots,1$}
\IF{$L_t=0$}
\STATE On $\cD_{-k}$, estimate $\hat Q_{t+1}^{\pi^e,-k}$, form
$\hat V_{t+1}^{\pi^e,-k}$ by averaging under $\pi^e$, and construct
$\hat Y_{t,0}^{\pi^e,-k}$.
\ENDIF
\IF{$L_t=T-t$}
\STATE Construct $\hat Y_{t,T-t}^{\pi^e,-k}$ using
$\{\widetilde\tau_j^{\pi^e,-k}\}_{j=t+1}^T$.
\ENDIF
\STATE Regress the selected constructed outcome on $S_t$ in $\cD_{-k}$
to obtain $\hat m_{t,L_t}^{\pi^e,-k}$.
\ENDFOR
\ENDFOR
\FOR{$t=T,\ldots,1$}
\STATE Estimate $\hat\blipadv_t^{\pi^e}$ by minimizing the pooled
held-out residualized loss over $\cG_{t,n}$ based on
$\hat Y_{t,L_t}^{\pi^e,-k}$ and $\hat m_{t,L_t}^{\pi^e,-k}$.
\ENDFOR
\STATE Return $\{\hat\blipadv_t^{\pi^e}\}_{t=1}^T$.
\end{algorithmic}
\end{algorithm*}

\clearpage
\subsection{Full statements of theorems}\label{apx-full-statements}
\subsubsection{Full statements of policy evaluation theorems}
\begin{remark}[Relation between the policy-evaluation theorems]
The main text theorem, \Cref{thm-policy-evaluation}, is the well-specified specialization of
\Cref{thm-mse-rates-policy-evaluation-apx}.
\Cref{thm-mse-rates-policy-evaluation-apx} retains an explicit approximation
term for misspecified classes. 
\end{remark}
The main text defines the local Rademacher complexity and critical radius used
below. The critical radius quantifies the statistical complexity of a function
class. Bounds on the critical radius of common function classes like linear
and polynomial models, deep neural networks, etc. can be found in standard
references on statistical learning theory, e.g. \citet{wainwright2019high}.
The resulting $L_2$-error bound leverages \citet{foster2019orthogonal}.

\begin{assumption}[{Current-stage product remainder rate}]\label{asn-producterrorrates-full-apx}
Fix an evaluation policy $\pi^e$ and an unrolling depth $L$. Conditional on
the cross-fitted future functions used to construct
$\hat Y_{t,L}^{\pi^e,-k}$, suppose
$\max_{k\leq K}\rho_{t,L}^{\pi^e,-k}=o_p(n^{-1/2})$.
\end{assumption}
\Cref{asn-producterrorrates-full-apx} controls only the current-stage
product remainder for the contemporaneous
conditional-mean regression and behavior policy.
The product error terms $\rho_{t,L}^{\pi^e,-k}$ highlight the reduced dependence
on individual nuisance error rates.

For exact fold-size bookkeeping, let
$\underline n:=\min_{k\le K}|\cD_k|$.
Let $r_{t,L,\underline n}$ denote the largest conditional critical-radius
bound for the fold-specific losses $\ell_{t,L}^{-k}$.
For each fixed $\delta\in(0,1)$, require the foldwise confidence floor
\[
r_{t,L,\underline n}^{2}\gtrsim
\frac{\log(K/\delta)}{\underline n}.
\]

\begin{theorem}[$L_2$-error rates for policy evaluation]\label{thm-mse-rates-policy-evaluation-apx}
Suppose
\Cref{asn-iid-trajs,asn-sequential-unconf,asn-bounded values,asn-overlap}
hold, as does \Cref{asn-producterrorrates-full-apx}.
Fix a stage $t$, a policy $\pi^e$, and an unrolling depth $L$.
Let $\hat\blipadv_{t,L}^{\pi^e}$ be the exact empirical risk minimizer,
$
\hat\blipadv_{t,L}^{\pi^e}
\in\argmin_{f\in\cG_{t,n}}
\widehat{\mathcal L}_{D,t,L}^{\pi^e}(f)$,
where $\widehat{\mathcal L}_{D,t,L}^{\pi^e}$ is the pooled held-out
residualized loss in \Cref{alg-dyn-blipadv-evaluation}. Then
\begin{equation}
\norm{\hat\blipadv_{t,L}^{\pi^e}-\blipadv_t^{\pi^e}}_2
=O_p\!\left(
r_{t,L,\underline n}
+\inf_{g\in\cG_{t,n}}\norm{g-\blipadv_t^{\pi^e}}_2
+\underline n^{-1/2}
+\max_{k\le K}\norm{b_{t,L}^{\pi^e,-k}}_2
\right).
\label{eq:diff-full-main-theorem}
\end{equation}
\end{theorem}

The bound includes separate contributions from critical radius, approximation error, current-stage
nuisance-product term, and future-function target shift.
The bias term $\inf_{g\in\cG_{t,n}}\norm{g-\blipadv_t^{\pi^e}}_2$ describes
the model misspecification bias of the function class parametrizing
$Q$-function contrasts, $\cG_{t,n}$.
Estimated future contrasts enter only through $b_{t,L}^{\pi^e,-k}$.
At $L=0$, Bellman completeness and an appropriate
rate for the future $Q_{t+1}$ model are one sufficient route for controlling
$b_{t,L}^{\pi^e,-k}$, although choosing $L_t =T-t$ does not require Bellman completeness. 
\section{Proofs for Decision-Centered Abstractions}
\label{apx-decision-centered-proofs}

\subsection{Position of optimal contrasts in the abstraction hierarchy}
\label{sec:ec-contrast-hierarchy}

The broader hierarchy of exact state abstractions is given by
\citet{li2006towards}; here we only state and prove results that determine the placement of the optimal
action-value contrast between preservation of the full optimal \(Q\)-vector
and preservation of the tie-broken optimal action.

We define the following optimal contrast function adn policy (under any consistent deterministic tie-breaking rule). Fix a stage \(t\), a reference action \(a_0\), and a deterministic
tie-breaking selector.
Define
\[
\blipadv_t^*(s;a,a_0)
:=Q_t^*(s,a)-Q_t^*(s,a_0),
\qquad
\pi_t^*(s)
:=\argmax_{a\in\mathcal A}Q_t^*(s,a)
\]

\begin{proposition}[Position of optimal contrasts in the hierarchy]
\label{prop:contrast-abstraction-hierarchy}
For any states \(s,\tilde s\) in the relevant stage-\(t\) support,
\[
Q_t^*(s,\cdot)=Q_t^*(\tilde s,\cdot)
\quad\Longrightarrow\quad
\blipadv_t^*(s;\cdot,a_0)
=
\blipadv_t^*(\tilde s;\cdot,a_0)
\quad\Longrightarrow\quad
\pi_t^*(s)=\pi_t^*(\tilde s).
\]
Neither converse holds in general. 
\end{proposition}

\begin{proof}{Proof of \Cref{prop:contrast-abstraction-hierarchy}}
The first implication follows by subtracting the common reference
coordinate \(Q_t^*(\cdot,a_0)\).  For the second,
\(
\argmax_{a\in\mathcal A}Q_t^*(s,a)
=
\argmax_{a\in\mathcal A}\blipadv_t^*(s;a,a_0)\).

 Note that both implications can be strict.  For two
actions ordered as \((a_0,a_1)\), the one-stage optimal \(Q\)-vectors
\(Q_t^*(s,\cdot)=(4,7)\) and
\(Q_t^*(\tilde s,\cdot)=(0,3)\)
have identical contrasts but different value levels.  Conversely,
\(Q_t^*(s,\cdot)=(0,2)\) and
\(Q_t^*(\tilde s,\cdot)=(0,5)\)
have the same unique optimal action but different contrasts. \qed
\end{proof}


\begin{proof}{Proof of \Cref{prop:no-abstract-planning}}
Consider two states at time \(t+1\) with $Q_{t+1}^*(s_H,\cdot)=(100,101),
Q_{t+1}^*(s_L,\cdot)=(0,1)$.
They have the same action contrast and can therefore be merged by an exact
decision-centered abstraction. However, $V_{t+1}^*(s_H)=101, 
V_{t+1}^*(s_L)=1$.
At an upstream state \(s_0\), let action \(a\) give immediate reward \(0\)
and transition deterministically to \(s_H\). Let action \(a'\) give immediate
reward \(50\) and transition deterministically to \(s_L\). With
\(\gamma=0.9\), \(Q_t^*(s_0,a)=0+0.9(101)=90.9\) and
\(Q_t^*(s_0,a')=50+0.9(1)=50.9\).
The ground-optimal action is \(a\).

Now merge \(s_H\) and \(s_L\) into one abstract state \(\bar s\). Both
upstream actions then have the same abstract continuation value, regardless
of the value assigned to \(\bar s\). Thus
\(\bar Q_t(s_0,a)=\gamma\bar V_{t+1}(\bar s)\) and
\(\bar Q_t(s_0,a')=50+\gamma\bar V_{t+1}(\bar s)\).
The induced abstract MDP selects \(a'\), which is suboptimal in the ground
MDP. Therefore, contrast preservation alone does not justify Bellman
planning after state aggregation. \qed
\end{proof}

\subsection{Proof of graphical characterization (\Cref{prop:reward-scope-characterization})}
\begin{proof}{Proof of \Cref{prop:reward-scope-characterization}}
By the additive reward representation \eqref{eq:additive-reward-scopes},
the termwise definition \eqref{eq:termwise-contrast}, and linearity of
conditional expectation,
\[
\blipadv_t^\pi(s;a,a_0)
=
\sum_{k=t}^{T}\sum_{H\in\mathcal H_k}
d_{t,k,H}^{\pi,a,a_0}(s).
\]
For the necessary direction, fix a coordinate \(j\) that fails the graphical
condition in \eqref{eq:graphical-decision-support}, a required continuation
policy \(\pi\), and an allowed action pair \((a,a_0)\). Consider one term node
\(r_{k,H}\).

If \(r_{k,H}\notin\operatorname{De}^{\pi}(A_t)\), then changing the current
action from \(a_0\) to \(a\) cannot change that reward term under the Markov
structural model. Hence $d_{t,k,H}^{\pi,a,a_0}=0$.

If \(r_{k,H}\in\operatorname{De}^{\pi}(A_t)\), then failure of the graphical
condition implies
$r_{k,H}\notin\operatorname{De}^{\pi}(S_t^j)$.
Under the Markov structural model, the corresponding termwise action
contrast is therefore invariant to \(s_j\) when the other current-state
coordinates are held fixed. Thus every summand in the displayed
decomposition is constant in \(s_j\). This holds for every required policy
and allowed action pair, so \(j\notin K_{\Delta,t}\). Therefore
$K_{\Delta,t}$ is contained in the right-hand side of
\eqref{eq:graphical-decision-support}.

Conversely, let \(j\) satisfy the graphical condition. Under the faithfulness
condition in the proposition, some required continuation policy and allowed
action pair has a termwise contribution that varies with \(s_j\), and this
variation does not cancel in the sum of termwise contributions. The displayed
decomposition is therefore nonconstant in \(s_j\), so \(j\in K_{\Delta,t}\).
This proves equality under the stated faithfulness condition. \qed
\end{proof}

\section{Proofs for Estimation }\label{apx-sec-proofs-estimation}

\subsection{Justifying identification of the method}
\begin{proposition}[Equivalent $L$-step moments]
\label{prop:diff-equivalent-L-moments}
For every $L\in\{0,\ldots,T-t\}$,
$\E[Y_{t,L}^{\pi^e}\mid S_t,A_t]=Q_t^{\pi^e}(S_t,A_t)$.
\end{proposition}

\begin{proof}{Proof of \Cref{prop:diff-equivalent-L-moments}}
For any $j$,
\[
V_j^{\pi^e}(S_j)
=
Q_j^{\pi^e}(S_j,A_j)
+
\{\pi_j^e(1\mid S_j)-A_j\}\blipadv_j^{\pi^e}(S_j),
\]
and
\[
Q_j^{\pi^e}(S_j,A_j)
=
\E[R_j+\gamma V_{j+1}^{\pi^e}(S_{j+1})\mid S_j,A_j].
\]
One substitution replaces the terminal value at stage $j$ by one realized
reward, one future action correction, and the next terminal value without
changing the conditional mean. Iterate $L$ times. \qed
\end{proof}

\paragraph{Proofs of the main-text estimation results.}

\begin{proof}{Proof of \Cref{prop:oracle-unrolling-variance}}
Let $j=t+L+1$. By the Markov property and the Bellman equation,
$\E[\xi_j^{\pi^e}\mid S_j,A_j]=0$, and iterated expectations therefore gives
$\operatorname{Cov}\!\left(Y_{t,L}^{\pi^e},\xi_j^{\pi^e}\mid S_t,A_t\right)=0$.
Using \cref{eq:innovation-representation},
\[
\operatorname{Var}\!\left(Y_{t,L+1}^{\pi^e}\mid S_t,A_t\right)
=\operatorname{Var}\!\left(Y_{t,L}^{\pi^e}\mid S_t,A_t\right)
+\gamma^{2(L+1)}\operatorname{Var}\!\left(\xi_j^{\pi^e}\mid S_t,A_t\right).
\]
\qed
\end{proof}

\begin{proof}{Proof of \Cref{prop:behavior-policy-control-variates}}
Condition on the sample used to construct $h$.  For every $j>t$,
$\E[\{\pi_j^b(1\mid S_j)-A_j\}h_j(S_j)\mid S_j]=0$.
Iterated expectation proves part~\textup{(i)}.  Write
$p_t=\pi_t^b(1\mid S_t)$ and
$\omega_t^b=p_t(1-p_t)$.  Part~\textup{(i)} gives
\[
\E[(A_t-p_t)\{Y_t^{b,\mathrm{cv}}(h)-m(S_t)\}\mid S_t]
=\omega_t^b(S_t)\blipadv_t^{\pi^b}(S_t),
\qquad \E[(A_t-p_t)^2\mid S_t]=\omega_t^b(S_t).
\]
Overlap makes the conditional quadratic strictly convex, so its unique
minimizer is $\blipadv_t^{\pi^b}$.  This proves part~\textup{(ii)}.
For fixed $f$ and $h$, the optimal regression main effect is
$
\E[Y_t^{b,\mathrm{cv}}(h)-(A_t-p_t)f(S_t)\mid S_t]
=V_t^{\pi^b}(S_t)$,
which proves the first claim in part~\textup{(iii)}.

For the variance claim, condition at a future stage $j$.  The
action-dependent part of the conditional mean of the remaining return is
$
\{A_j-\pi_j^b(1\mid S_j)\}
\{\blipadv_j^{\pi^b}(S_j)-h_j(S_j)\}$.
Its conditional action variance is
$
\omega_j^b(S_j)
\{\blipadv_j^{\pi^b}(S_j)-h_j(S_j)\}^2$,
which is minimized by $h_j=\blipadv_j^{\pi^b}$.  The remaining
conditional-variance term does not depend on $h_j$.  Apply this decomposition
backward from $j=T$ through $j=t+1$.  The oracle sequence therefore weakly
minimizes the conditional variance given $(S_t,A_t)$. \qed
\end{proof}


\subsection{Orthogonality}\label{apx-orthogonality}
\paragraph{Preliminaries}
Throughout EC.4.2--EC.4.3, recall that
$\eta_{t,L}:=(m_{t,L},\pi_t^b)$.
Every future function used to construct
$\hat Y_{t,L}^{\pi^e,-k}$ is held fixed, and we track resulting error therein through $b_{t,L}^{\pi^e,-k}$.

Below we will omit the $\pi$ superscript; the analysis below holds for any valid $\pi$.
We define for any functional $L(f)$ the Frechet derivative as
\[
D_f L(f)[\nu]=\left.\frac{\partial}{\partial t} L(f+t \nu)\right|_{t=0}.
\]

Higher order derivatives are denoted as $D_{g, f} L(f, g)[\mu, \nu]$.
\begin{lemma}[Action-contrast operator bound under overlap]
\label{lem:diff-overlap-action-contrast}
Under \cref{asn-sequential-unconf,asn-concentrability}, for every $j>t$,
$\norm{\E[f(S_j)\mid S_t,A_t=a]}_2
\leq\sqrt{C_\infty}\norm f_2$ and
$\norm{\cC_{t,j}f}_2\leq2\sqrt{C_\infty}\norm f_2$.
(All norms are evaluated under the observational behavior marginals of the respective timestep's $S_t$).
Equivalently, if $\pi_t^b(a\mid s)\geq\epsilon$ for both actions, the two
constants are $1/\sqrt\epsilon$ and $2/\sqrt\epsilon$.
\end{lemma}

\begin{proof}{Proof of \Cref{lem:diff-overlap-action-contrast}}
Apply \cref{asn-concentrability} to the policy that follows behavior before
time $t$ and selects $a$ at time $t$ to obtain
$\pi_t^b(a\mid s)\geq C_\infty^{-1}$. Jensen's inequality and the behavior
mixture over $A_t$ give
$\norm{\E[f(S_j)\mid S_t,A_t=a]}_2^2
\leq C_\infty\norm f_2^2$. The triangle inequality gives the contrast bound. \qed
\end{proof}

\begin{lemma}[Universal orthogonality in the current-timestep nuisances]
\label{lemma-universal-ortho}
\label{lem:diff-contemporaneous-orthogonality}
Conditional on $\cD_{-k}$, define
\[
\mathcal L_{D,t,L}^{\pi^e,-k}(\blipadv;m,\pi^b)
:=\E\big[\{\hat Y_{t,L}^{\pi^e,-k}-m(S_t)
-\{A_t-\pi^b(1\mid S_t)\}\blipadv(S_t)\}^2\mid\cD_{-k}\big].
\]
At $m=m_{t,L}^{\pi^e,-k}$ and $\pi^b=\pi_t^b$,
$D_{m,\blipadv}\mathcal L_{D,t,L}^{\pi^e,-k}[h_m,h_{\blipadv}]=0$ and
$D_{\pi^b,\blipadv}\mathcal L_{D,t,L}^{\pi^e,-k}[h_{\pi},h_{\blipadv}]=0$
for every candidate $\blipadv$ and admissible directions. 
\end{lemma}
\begin{proof}{Proof of \Cref{lemma-universal-ortho}}
Condition on $\cD_{-k}$. At
$m=m_{t,L}^{\pi^e,-k}$ and $\pi^b=\pi_t^b$, direct differentiation
gives
\begin{align*}
D_{m,\blipadv}\mathcal L_{D,t,L}^{\pi^e,-k}[h_m,h_{\blipadv}]
&=
2\E[ h_m(S_t)\{A_t-\pi_t^b(1\mid S_t)\}
h_{\blipadv}(S_t)\mid\cD_{-k}]
=0,\\
D_{\pi^b,\blipadv}\mathcal L_{D,t,L}^{\pi^e,-k}[h_{\pi},h_{\blipadv}]
&=
2\E[ h_{\pi}(S_t)h_{\blipadv}(S_t)
\{\hat Y_{t,L}^{\pi^e,-k}-m_{t,L}^{\pi^e,-k}(S_t)
-\{A_t-\pi_t^b(1\mid S_t)\}\blipadv(S_t)\}
\mid\cD_{-k}]\\
&\quad-
2\E[\{A_t-\pi_t^b(1\mid S_t)\}
h_{\pi}(S_t)\blipadv(S_t)h_{\blipadv}(S_t)
\mid\cD_{-k}]
=0.
\end{align*}
Both equalities use
$\E[A_t-\pi_t^b(1\mid S_t)\mid S_t]=0$; the second also uses the
definition of $m_{t,L}^{\pi^e,-k}$. \qed
\end{proof}

\begin{lemma}[Current-timestep plug-in remainder]
\label{lemma-secondorderderivatives}
Conditional on $\cD_{-k}$, let
$
h_\pi:=\hat\pi_t^{b,-k}-\pi_t^b,
\qquad
h_m:=\hat m_{t,L}^{\pi^e,-k}-m_{t,L}^{\pi^e,-k}$,
where
$
m_{t,L}^{\pi^e,-k}(s)
:=\E[\hat Y_{t,L}^{\pi^e,-k}\mid S_t=s,\cD_{-k}]$.
For any candidate $\blipadv$ with $\norm{\blipadv}_\infty\leq\boundadv$
and any contrast direction $\nu_t$,
\begin{align*}
&D_{\blipadv}\mathcal L_{D,t,L}^{\pi^e,-k}
(\blipadv;\hat m_{t,L}^{\pi^e,-k},\hat\pi_t^{b,-k})[\nu_t]
-D_{\blipadv}\mathcal L_{D,t,L}^{\pi^e,-k}
(\blipadv;m_{t,L}^{\pi^e,-k},\pi_t^b)[\nu_t]\\
&\qquad=2\E[\blipadv(S_t)h_\pi^2(S_t)\nu_t(S_t)
-h_\pi(S_t)h_m(S_t)\nu_t(S_t)\mid\cD_{-k}].
\end{align*}
Suppose $\boundadv\geq1$. Then
\[
\left|
D_{\blipadv}\mathcal L_{D,t,L}^{\pi^e,-k}
(\blipadv;\hat m_{t,L}^{\pi^e,-k},\hat\pi_t^{b,-k})[\nu_t]
-D_{\blipadv}\mathcal L_{D,t,L}^{\pi^e,-k}
(\blipadv;m_{t,L}^{\pi^e,-k},\pi_t^b)[\nu_t]
\right|
\leq2\rho_{t,L}^{\pi^e,-k}\norm{\nu_t}_2.
\]
Future-function error is not part of $\rho_{t,L}^{\pi^e,-k}$; it is
represented by $b_{t,L}^{\pi^e,-k}$.
\end{lemma}
\begin{proof}{Proof of \Cref{lemma-secondorderderivatives}}
Condition on $\cD_{-k}$, and write
$Z_t=A_t-\pi_t^b(1\mid S_t)$ and
$
r_t=\hat Y_{t,L}^{\pi^e,-k}-m_{t,L}^{\pi^e,-k}(S_t)
-Z_t\blipadv(S_t)$.
The score at the current population nuisances is
$D_{\blipadv}\mathcal L_{D,t,L}^{\pi^e,-k}
(\blipadv;m_{t,L}^{\pi^e,-k},\pi_t^b)[\nu_t]
=-2\E[r_tZ_t\nu_t(S_t)\mid\cD_{-k}]$.
At the estimated current nuisances, the two factors in the score are
$r_t-h_m(S_t)+h_\pi(S_t)\blipadv(S_t)$ and
$Z_t-h_\pi(S_t)$. Subtract the population score and expand. The terms that
are linear in $h_m$ or $h_\pi$ vanish because
$\E[Z_t\mid S_t,\cD_{-k}]=0$ and
$\E[r_t\mid S_t,\cD_{-k}]=0$. The two remaining terms give the exact
identity in the lemma. Cauchy--Schwarz gives
\[
2\{\boundadv\norm{h_\pi^2}_2+\norm{h_\pi h_m}_2\}\norm{\nu_t}_2
\leq2\rho_{t,L}^{\pi^e,-k}\norm{\nu_t}_2,
\]
where the last inequality uses $\boundadv\geq1$. \qed
\end{proof}

\subsection{Proof of sample complexity bounds}

\begin{proof}{Proof of \Cref{thm-mse-rates-policy-evaluation-apx}}
\textbf{Overview}: 
We prove policy evaluation sample complexity bounds for a fixed stage $t$, an evaluation policy $\pi^e$, and an unrolling depth $L$. The proof strategy is overall similar to that of \citet{foster2019orthogonal,lewis2021double}, but specializing to the squared loss.
We first compare the estimator's empirical loss with that of a
candidate approximating the true contrast. We then control the difference
between empirical and population losses on each fold, and use population
curvature to bound the estimation error by exact expansion of the squared loss. The first-order expansion leverages the Neyman-orthogonality of the loss to expose the product error rates, and projected orthogonality and error rates from future nuisance functions.

\paragraph{Step 1: Compare pooled empirical losses.}
To handle potential misspecification, we introduce a comparator $g_t^*$ that is an approximate projection of the true
$Q$-function contrast onto $\cG_{t,n}$. For arbitrary $\varepsilon_{\mathrm{app}}>0$, choose a deterministic comparator $g_t^*\in\cG_{t,n}$ such that
$
\norm{g_t^*-\blipadv_t^{\pi^e}}_2
\leq\inf_{g\in\cG_{t,n}}
\norm{g-\blipadv_t^{\pi^e}}_2+\varepsilon_{\mathrm{app}}$.
 Define the error of the estimate
relative to this approximate projection by
\[
\nu_t:=\hat\blipadv_{t,L}^{\pi^e}-g_t^*.
\]
This comparator is fixed under every fold conditioning.
Write $O_i$ for observed trajectory $i$ and $O$ for an independent
trajectory from the behavior distribution. Recall the fold-specific loss
\[
\ell_{t,L}^{-k}(O;g)
:=\left\{\hat Y_{t,L}^{\pi^e,-k}
-\hat m_{t,L}^{\pi^e,-k}(S_t)
-\{A_t-\hat\pi_t^{b,-k}(1\mid S_t)\}g(S_t)\right\}^2.
\]
Set $w_k:=|\cD_k|/n$, so $\sum_k w_k=1$.
Because the estimator minimizes the pooled empirical loss,
\[
\sum_{k=1}^K\frac{w_k}{|\cD_k|}\sum_{i\in\cD_k}
\left\{\ell_{t,L}^{-k}(O_i;\hat\blipadv_{t,L}^{\pi^e})
-\ell_{t,L}^{-k}(O_i;g_t^*)\right\}\leq0.
\]

\paragraph{Step 2: Control empirical-to-population deviations.}
We next control each fold's empirical loss difference uniformly over the
candidate class. For each $k$, condition on $\cD_{-k}$.
All fitted functions entering $\ell_{t,L}^{-k}$ use only $\cD_{-k}$,
including the future functions in the constructed outcome. Thus the
held-out trajectories remain independent of these fits, and the class
\[
\mathcal H_{t,L,k}^{*}
:=
\left\{
\ell_{t,L}^{-k}(\mathord\cdot;g)
-\ell_{t,L}^{-k}(\mathord\cdot;g_t^*)
:
g\in\cG_{t,n}
\right\}
\]
is fixed. Factoring the difference of squares and using
\Cref{asn-bounded values} bounds the absolute loss difference by a
constant times $|g(S_t)-g_t^*(S_t)|$. Consequently,
\[
\norm{
\ell_{t,L}^{-k}(\mathord\cdot;g)
-\ell_{t,L}^{-k}(\mathord\cdot;g_t^*)
}_2
\leq
C_{\mathrm{env}}\norm{g-g_t^*}_2,
\]
where $C_{\mathrm{env}}$ depends only on the fixed envelope bounds.
Fix $\delta\in(0,1)$. Apply \Cref{lemma-lemma14-foster2019}, the localized Rademacher complexity bound, conditionally to
$\mathcal H_{t,L,k}^{*}\subseteq\mathcal H_{t,L,k}$, the pairwise
loss-difference class defining the critical radius. With conditional
failure probability at most $\delta/K$, it gives, for every $g\in\cG_{t,n}$,
\[
\begin{aligned}
\Bigg|&\frac{1}{|\cD_k|}\sum_{i\in\cD_k}
\left\{\ell_{t,L}^{-k}(O_i;g)-\ell_{t,L}^{-k}(O_i;g_t^*)\right\}-\E\!\left[
\ell_{t,L}^{-k}(O;g)-\ell_{t,L}^{-k}(O;g_t^*)
\right]\Bigg|\\
&\lesssim r_{t,L,\underline n}
\{\norm{g-g_t^*}_2+r_{t,L,\underline n}\},
\end{aligned}
\]
where $\underline n=\min_{k\leq K}|\cD_k|$. The confidence-floor condition ensures
the stated conditional failure probability. Averaging over each training
sample and taking a union bound gives a simultaneous event of probability
at least $1-\delta$. 

\paragraph{Step 3: Strong convexity via overlap gives quadratic curvature.}
We now bound each fold's population loss difference from below.
The squared loss is exactly quadratic in its $\blipadv$ contrast argument. Expanding
around $g_t^*$ gives a linear term, written as the directional derivative
in direction of estimation error $\nu_t$, and a quadratic term:
\begin{align*}
&\E\!\left[
\ell_{t,L}^{-k}(O;\hat\blipadv_{t,L}^{\pi^e})
-\ell_{t,L}^{-k}(O;g_t^*)\right]\\
&\qquad=
D_{\blipadv}\mathcal L_{D,t,L}^{\pi^e,-k}
(g_t^*;\hat m_{t,L}^{\pi^e,-k},
\hat\pi_t^{b,-k})[\nu_t]
+\E[(A_t-\hat\pi_t^{b,-k}(1\mid S_t))^2
\nu_t(S_t)^2].
\end{align*}
The quadratic term controls $\norm{\nu_t}_2^2$ by overlap.
Recall $\omega_t^b(s):=\pi_t^b(1\mid s)\{1-\pi_t^b(1\mid s)\}$
and $\lambda:=\epsilon(1-\epsilon)>0$. Then
\[
\E[(A_t-\hat\pi_t^{b,-k}(1\mid S_t))^2
\mid S_t]
=\omega_t^b(S_t)
+\{\hat\pi_t^{b,-k}(1\mid S_t)
-\pi_t^b(1\mid S_t)\}^2
\geq\lambda,
\]
so the quadratic term is at least $\lambda\norm{\nu_t}_2^2$.
\paragraph{Step 4: Bound the linear term using orthogonality and the target shift.}
To bound the linear term, first replace the current nuisance estimates by
$m_{t,L}^{\pi^e,-k}(s)=\E[\hat Y_{t,L}^{\pi^e,-k}\mid S_t=s,\cD_{-k}]$
and $\pi_t^b$. With these current nuisances and the fitted future functions,
\cref{eq:diff-target-shift} identifies the contrast target as
$\blipadv_t^{\pi^e}+b_{t,L}^{\pi^e,-k}$. Therefore,
\[
D_{\blipadv}\mathcal L_{D,t,L}^{\pi^e,-k}
(g_t^*;m_{t,L}^{\pi^e,-k},\pi_t^b)[\nu_t]
=2\E[\omega_t^b(S_t)
\{g_t^*(S_t)-\blipadv_t^{\pi^e}(S_t)
-b_{t,L}^{\pi^e,-k}(S_t)\}
\nu_t(S_t)].
\]
For binary actions,
$\omega_t^b(S_t)
=\pi_t^b(1\mid S_t)\{1-\pi_t^b(1\mid S_t)\}\leq1/4$.
Thus Cauchy--Schwarz bounds the absolute value of this derivative by
$\frac12\norm{g_t^*-\blipadv_t^{\pi^e}
-b_{t,L}^{\pi^e,-k}}_2\norm{\nu_t}_2$.
Restoring the fitted current nuisances changes this derivative by at most
$2\rho_{t,L}^{\pi^e,-k}\norm{\nu_t}_2$, by
\Cref{lemma-secondorderderivatives}. This is where orthogonality enters:
the current nuisance errors contribute through the products in $\rho$.
Combining the linear and quadratic bounds gives
\[
\E\!\left[\ell_{t,L}^{-k}(O;\hat\blipadv_{t,L}^{\pi^e})
-\ell_{t,L}^{-k}(O;g_t^*)\right]
\geq \lambda\norm{\nu_t}_2^2
-2\left\{\frac14\norm{g_t^*-\blipadv_t^{\pi^e}
-b_{t,L}^{\pi^e,-k}}_2
+\rho_{t,L}^{\pi^e,-k}\right\}\norm{\nu_t}_2.
\]
\paragraph{Step 5: Combine the bounds to obtain the estimation rate.}
Sum this lower bound with weights $w_k$. The earlier uniform deviation
bound and pooled empirical loss comparison bound the same sum from above
by the critical-radius term. Since the weights sum to one, each weighted
nuisance error is at most its maximum over folds. The triangle inequality
separates approximation error from the target shifts, giving
\[
\lambda\norm{\nu_t}_2^2
\lesssim r_{t,L,\underline n}
\{\norm{\nu_t}_2+r_{t,L,\underline n}\}
+\left\{\frac14\norm{g_t^*-\blipadv_t^{\pi^e}}_2
+\frac14\max_{k\leq K}\norm{b_{t,L}^{\pi^e,-k}}_2
+\max_{k\leq K}\rho_{t,L}^{\pi^e,-k}\right\}\norm{\nu_t}_2.
\]
The AM--GM inequality implies that for $x,y\geq0$ and $\sigma>0$,
\[
xy\leq\frac12\left(\frac{2}{\sigma}x^2
+\frac{\sigma}{2}y^2\right).
\]
Apply this inequality to each product involving $\norm{\nu_t}_2$,
choosing $\sigma$ small enough to absorb their quadratic contributions
into $\lambda\norm{\nu_t}_2^2$. Taking square roots gives
\[
\norm{\nu_t}_2
\lesssim r_{t,L,\underline n}
+\norm{g_t^*-\blipadv_t^{\pi^e}}_2
+\max_{k\leq K}\norm{b_{t,L}^{\pi^e,-k}}_2
+\max_{k\leq K}\rho_{t,L}^{\pi^e,-k},
\]
with constants depending on the fixed overlap and envelope bounds.
Finally,
$\norm{\hat\blipadv_{t,L}^{\pi^e}-\blipadv_t^{\pi^e}}_2
\leq\norm{\nu_t}_2+\norm{g_t^*-\blipadv_t^{\pi^e}}_2$.
Use the comparator's approximation bound and
\Cref{asn-producterrorrates-full-apx}; since $\underline n\leq n$,
the current nuisance remainder is also $o_p(\underline n^{-1/2})$.
We obtain
\[
\norm{\hat\blipadv_{t,L}^{\pi^e}
-\blipadv_t^{\pi^e}}_2
=O_p\!\left(
r_{t,L,\underline n}
+\inf_{g\in\cG_{t,n}}\norm{g-\blipadv_t^{\pi^e}}_2
+\underline n^{-1/2}
+\max_{k\leq K}\norm{b_{t,L}^{\pi^e,-k}}_2
\right),
\]
after letting $\varepsilon_{\mathrm{app}}\downarrow0$. Thus future-function errors enter only through
the fold target shifts. \qed
\end{proof}

\begin{proof}{Proof of
\Cref{cor:diff-main-uncontrollable-continuation}}
At $L=0$, decompose the continuation error into its uncontrollable and
controllable components.  The conditional independence assumption gives
$
\cC_{t,t+1}
[\hat V_{t+1}^{\bar a,\pi^e,-k}
-V_{t+1}^{\bar a,\pi^e}]=0$.
Thus only the controllable component remains in the one-step target shift.
Apply \Cref{lem:diff-overlap-action-contrast} to that component and multiply
by $\gamma$ to obtain the displayed bound. \qed
\end{proof}

\begin{proof}{Proof of \Cref{cor:full-unrolling-rate}}
For each stage $t$, let $e_t$ be the maximum $L_2$ error over the finite
collection of cross-fitted stage-$t$ estimators used by the full-unrolling
procedure, including the pooled estimator.
At $L=T-t$, \Cref{thm-mse-rates-policy-evaluation-apx} and
\cref{eq:diff-target-shift} imply the following common-event statement. For
every fixed $\delta\in(0,1)$, there is a constant $C_0(\delta)$, independent of $n$,
such that with probability at least $1-\delta$, simultaneously over the finite
collection of stages and fold-specific estimators,
\[
e_t\leq C_0(\delta)
\left\{
r_{t,T-t,\underline n}+\underline n^{-1/2}
\right\}
+C_0(\delta)\sum_{j=t+1}^T\gamma^{j-t}
\frac{2}{\sqrt\epsilon}
\norm{\pi_j^e-\pi_j^b}_\infty e_j,
\]
where \Cref{lem:diff-overlap-action-contrast} bounds each action-contrast
operator.
Because $T$ is fixed, we can solve the strictly upper triangular recursion via backward substitution, which gives
\[
e_t\leq C(\delta)
\max_{j\geq t}\left\{r_{j,T-j,\underline n}+\underline n^{-1/2}\right\}.
\]
Because the folds are approximately balanced and $K$ is fixed,
$\underline n\asymp n/K\asymp n$. Thus replacing $\underline n$ by $n$
in the rate notation changes only constants.
This proves \cref{eq:full-unrolling-rate-propagation}. Finally, when
$\pi^e=\pi^b$ and the known
future behavior probabilities are used in the action-centering terms, every
future-policy multiplier is zero and full unrolling has no terminal value
term, so the target shift vanishes. \qed
\end{proof}

\begin{proof}{Proof of \Cref{cor:dimension-adaptation}}
Let $\mathcal E_{\mathrm{sel}}
=\{K_{\Delta,t}\subseteq\hat K_{\Delta,t},
|\hat K_{\Delta,t}|\le 2|K_{\Delta,t}|\}$. Conditional on the
independent selection sample and on $\mathcal E_{\mathrm{sel}}$, the selected
set is fixed, contains the true support, and has at most
$2|K_{\Delta,t}|$ coordinates. Thus the true contrast remains in
the refit class. The fixed-set oracle guarantee implies the stated refit
bound except on an event of
conditional probability at most $\delta$. Since
$P(\mathcal E_{\mathrm{sel}}^c)\leq\delta_{\mathrm{sel},n}$, a union bound
gives failure probability at most $\delta_{\mathrm{sel},n}+\delta$, which is
the claimed result. \qed
\end{proof}

\subsubsection{Results used from other works}
Here we collect technical lemmas from other works, stated without proof. 

\begin{lemma}[Localized deviation for a bounded scalar class]
\label{lemma-lemma14-foster2019}
Let $O_1,\ldots,O_n$ be independent trajectories from the behavior
distribution, and let $O$ be an independent trajectory with the same law.
Let $\mathcal H$ be a class of scalar measurable functions with envelope
$B<\infty$. Let $\delta_n$ satisfy
\[
\mathscr R_n\!\left(\op{star}(\mathcal H/B);\delta_n\right)\leq\delta_n^2,
\qquad \delta_n^2\geq c\frac{\log(1/\xi)}{n},
\]
for a sufficiently large universal constant $c$. Then, with probability at
least $1-\xi$, simultaneously for every $h\in\mathcal H$,
\begin{equation*}
\left|\frac1n\sum_{i=1}^n h(O_i)-\E[h(O)]\right|
\leq
CB\delta_n
\left\{
\frac{\norm{h}_2}{B}+\delta_n
\right\},
\end{equation*}
where $C$ is universal. The same statement holds conditionally on an
independent training sample when $\mathcal H$ is fixed after conditioning.
\end{lemma}
This is a standard localized-Rademacher deviation consequence; see
\citet[Lemma~14]{foster2019orthogonal} and \citet{wainwright2019high}.

\begin{revision}
    
\subsection{Support recovery for decision-centered coordinate support}

\subsubsection{Thresholded LASSO}

Fix $t$, use full unrolling at $\pi^e=\pi^b$, and suppose
$\blipadv_t^{\pi^b}(s)=s^\top\beta_t$. On the independent selection sample,
let $n_{\rm sel}:=\sum_{k=1}^K|\mathcal D_k|$ and define
\[
\hat Z_{it,T-t}^{\pi^b,-k}
:=
\hat Y_{it,T-t}^{\pi^b,-k}
-
\hat m_{t,T-t}^{\pi^b,-k}(S_{it}),
\qquad
\hat D_{it}^{-k}
:=
A_{it}-\hat\pi_t^{b,-k}(1\mid S_{it}).
\]
Define the normalized selection loss locally as
\[
\widehat{\mathcal L}^{b}_{t,\rm sel}(\beta)
:=
\frac{1}{n_{\rm sel}}
\sum_{k=1}^{K}\sum_{i\in\mathcal D_k}
\left\{
\hat Z_{it,T-t}^{\pi^b,-k}
-
\hat D_{it}^{-k}S_{it}^{\top}\beta
\right\}^{2},
\]
and set
\[
\hat\beta_t
\in
\argmin_{\beta\in\mathbb R^d}
\left\{
\widehat{\mathcal L}^{b}_{t,\rm sel}(\beta)
+2\lambda_{t,n}\norm{\beta}_1
\right\},
\qquad
\hat K_{\Delta,t}
=
\{j:|\hat\beta_{t,j}|>v_{t,n}\},
\]
as in Section~4.3.

\begin{assumption}[Selection score and restricted curvature]\label{ass:tl-score-re}
For some $\kappa_t>0$,
\begin{equation}
\left\norm{
\frac{1}{n_{\rm sel}}
\sum_{k=1}^K\sum_{i\in\mathcal D_k}
\hat D_{it}^{-k}S_{it}
\bigl(\hat Z_{it,T-t}^{\pi^b,-k}-\hat D_{it}^{-k}S_{it}^\top\beta_t\bigr)
\right}_\infty
\le \frac{\lambda_{t,n}}{2},
\label{eq:tl-score-ec}
\end{equation}
and
\begin{equation}
\frac{1}{n_{\rm sel}}
\sum_{k=1}^K\sum_{i\in\mathcal D_k}
(\hat D_{it}^{-k})^2(S_{it}^\top u)^2
\ge \kappa_t\norm{u}_2^2
\label{eq:tl-re-ec}
\end{equation}
for every $u\in\mathbb R^d$ satisfying
$\norm{u_{K_{\Delta,t}^c}}_1\le3\norm{u_{K_{\Delta,t}}}_1$.
\end{assumption}

\begin{assumption}[Threshold calibration and beta-min]\label{ass:tl-betamin}
Fix a support-inflation factor $C_{\rm sel}>1$. The threshold and minimum active coefficient satisfy
\[
v_{t,n}
\ge
\frac{3\lambda_{t,n}}{\kappa_t\sqrt{C_{\rm sel}-1}},
\qquad
\min_{j\in K_{\Delta,t}}|\beta_{t,j}|
>
v_{t,n}
+\frac{3\sqrt{|K_{\Delta,t}|}\lambda_{t,n}}{\kappa_t}.
\]
\end{assumption}

\begin{proposition}[Thresholded-LASSO selector]\label{prop:tl-support-ec}
Under Assumptions~\ref{ass:tl-score-re}--\ref{ass:tl-betamin},
$\norm{\hat\beta_t-\beta_t}_2\le3\sqrt{\lvert K_{\Delta,t}\rvert}\lambda_{t,n}/\kappa_t$,
$K_{\Delta,t}\subseteq\hat K_{\Delta,t}$, and
$|\hat K_{\Delta,t}|\le C_{\rm sel}|K_{\Delta,t}|$.
If the assumptions hold simultaneously over $t=1,\ldots,T$ with probability at least $1-\delta_{\rm sel,n}$, then
\[
P\!\left(
K_{\Delta,t}\subseteq\hat K_{\Delta,t},
|\hat K_{\Delta,t}|\le C_{\rm sel}|K_{\Delta,t}|
\text{  for every }t
\right)
\ge1-\delta_{\rm sel,n}.
\]
\end{proposition}

\Cref{prop:tl-support-main} is the specialization of this result to
$C_{\rm sel}=2$.

\begin{proof}{Proof of \Cref{prop:tl-support-ec}}
Set $u=\hat\beta_t-\beta_t$. The basic inequality for the penalized loss and \eqref{eq:tl-score-ec} give
\begin{equation}
\frac{1}{n_{\rm sel}}
\sum_{k=1}^K\sum_{i\in\mathcal D_k}
(\hat D_{it}^{-k})^2(S_{it}^\top u)^2
+
\lambda_{t,n}\norm{u_{K_{\Delta,t}^c}}_1
\le
3\lambda_{t,n}\norm{u_{K_{\Delta,t}}}_1.
\label{eq:tl-basic-ec}
\end{equation}
Hence $u$ lies in the cone in Assumption~\ref{ass:tl-score-re}. Using \eqref{eq:tl-re-ec} and $\norm{u_{K_{\Delta,t}}}_1\le\sqrt{\lvert K_{\Delta,t}\rvert}\norm{u}_2$ in \eqref{eq:tl-basic-ec} yields
$\kappa_t\norm{u}_2^2\le3\sqrt{\lvert K_{\Delta,t}\rvert}\lambda_{t,n}\norm{u}_2$,
which proves the coefficient bound. Write
$B_{t,n}:=3\sqrt{|K_{\Delta,t}|}\lambda_{t,n}/\kappa_t$. For each
$j\in K_{\Delta,t}$,
\[
|\hat\beta_{t,j}|
\ge
|\beta_{t,j}|-|\hat\beta_{t,j}-\beta_{t,j}|
>
v_{t,n}+B_{t,n}-\norm{u}_2
\ge v_{t,n},
\]
so $K_{\Delta,t}\subseteq\hat K_{\Delta,t}$. Let
$F_t:=\hat K_{\Delta,t}\setminus K_{\Delta,t}$. Since $\beta_{t,j}=0$
and $|\hat\beta_{t,j}|>v_{t,n}$ for every $j\in F_t$,
\[
|F_t|v_{t,n}^2
<
\norm{u_{F_t}}_2^2
\le
\norm{u}_2^2
\le
B_{t,n}^2
\le
(C_{\rm sel}-1)|K_{\Delta,t}|v_{t,n}^2.
\]
Therefore $|\hat K_{\Delta,t}|\le C_{\rm sel}|K_{\Delta,t}|$. The
simultaneous statement follows directly from the joint event. This is the
standard restricted-eigenvalue plus beta-min sure-screening argument; see
\citet{bickel2009simultaneous},
\citet{zhou2009thresholding,zhou2010thresholded}, and
\citet{van2011adaptive}. \qed
\end{proof}

\paragraph{How the orthogonal analysis verifies \eqref{eq:tl-score-ec}.}
Assume the coordinates are standardized so that $\norm{S_{t,j}}_2\le1$.
Fix a fold $k$ and condition on $\mathcal D_{-k}$. At full unrolling under
$\pi^e=\pi^b$, with the known future behavior probabilities used in the
action-centering terms, the continuation target shift is zero. Apply
\Cref{lemma-secondorderderivatives} at the true linear contrast with the
coordinate direction $s\mapsto s_j$. Then
\begin{equation}
\left\norm{
\E\!\left[
\hat D_t^{-k}S_t
\bigl(\hat Z_{t,T-t}^{\pi^b,-k}-\hat D_t^{-k}S_t^\top\beta_t\bigr)
\,\middle|\,\mathcal D_{-k}
\right]
\right}_\infty
\le
\rho_{t,T-t}^{\pi^b,-k}.
\label{eq:tl-pop-score-ec}
\end{equation}
Suppose, in addition, that conditional maximal concentration on the held-out fold gives, simultaneously over $t$ and $k$,
\begin{equation}
\begin{aligned}
&\left\norm{
\frac{1}{|\mathcal D_k|}
\sum_{i\in\mathcal D_k}
\hat D_{it}^{-k}S_{it}
\bigl(\hat Z_{it,T-t}^{\pi^b,-k}-\hat D_{it}^{-k}S_{it}^\top\beta_t\bigr)
-
\E\!\left[
\hat D_t^{-k}S_t
\bigl(\hat Z_{t,T-t}^{\pi^b,-k}-\hat D_t^{-k}S_t^\top\beta_t\bigr)
\,\middle|\,\mathcal D_{-k}
\right]
\right}_\infty
\\
&\qquad\le
C\sqrt{\frac{\log(dTK/\delta)}{|\mathcal D_k|}}
\end{aligned}
\label{eq:tl-concentration-ec}
\end{equation}
with probability at least $1-\delta$. Condition \eqref{eq:tl-concentration-ec} follows by applying
Hoeffding's inequality for sub-Gaussian sums conditionally on
$\mathcal D_{-k}$, followed by a union bound over the $d$
coordinates, $T$ stages, and $K$ folds, provided the centered
score coordinates have uniformly bounded conditional
sub-Gaussian norms. The pooled score in \eqref{eq:tl-score-ec} is the fold-size-weighted average of these foldwise scores. Combining \eqref{eq:tl-pop-score-ec}--\eqref{eq:tl-concentration-ec}, \eqref{eq:tl-score-ec} holds whenever
\begin{equation}
\lambda_{t,n}
\ge
2\max_{k\le K}
\left\{
C\sqrt{\frac{\log(dTK/\delta)}{|\mathcal D_k|}}
+
\rho_{t,T-t}^{\pi^b,-k}
\right\}.
\label{eq:tl-lambda-ec}
\end{equation}
\Cref{asn-producterrorrates} makes the nuisance-product term second order. Thus the selection theorem uses the same current-stage nuisance-product condition as the main orthogonal estimation theorem. The remaining design requirement is the explicit empirical restricted-curvature condition \eqref{eq:tl-re-ec}, exactly as in classical thresholded-LASSO analyses and in \citet{zhou2024rewardrelevancefiltered}.

\end{revision}

\section{Proofs for Policy Optimization}\label{sec-pol-opt}

\begin{revision}
\subsection{Local policy improvement}

\begin{proof}{Proof of \Cref{prop-behavior-centered-improvement}}
The finite-horizon performance-difference identity gives
\begin{equation}
\E[V_1^{\pi^w}(S_1)]-\E[V_1^{\pi^b}(S_1)]
=
w\sum_{t=1}^T\gamma^{t-1}
\E_{d_t^{\pi^w}}
\left[\Delta_t^{\mathrm{gr}}(S_t)
\blipadv_t^{\pi^b}(S_t)\right].
\label{eqn-behavior-centered-pd}
\end{equation}
A coupling that uses the same initial state, transitions, and actions until
the two policies first disagree gives
\begin{equation}\textstyle 
\operatorname{TV}(d_t^{\pi^w},d_t^{\pi^b})
\leq
\min\{1,w\sum_{j<t}
\norm{\Delta_j^{\mathrm{gr}}}_\infty\}
\leq
w\sum_{j<t}\norm{\Delta_j^{\mathrm{gr}}}_\infty.
\label{eqn-behavior-centered-occupancy}
\end{equation}
For any bounded $f$, the difference between its expectations under two
distributions is at most $2\norm{f}_\infty$ times their total-variation
distance. Apply this inequality to
$f=\Delta_t^{\mathrm{gr}}\blipadv_t^{\pi^b}$ in
\eqref{eqn-behavior-centered-pd}, use
\eqref{eqn-behavior-centered-occupancy}, subtract the same expression under
$d_t^{\pi^b}$, and sum over $t$. More precisely, the absolute difference
between the value change and its first-order term is at most
\[
2\boundadv w^2
\sum_{t=1}^T\gamma^{t-1}
\norm{\Delta_t^{\mathrm{gr}}}_\infty
\sum_{j<t}\norm{\Delta_j^{\mathrm{gr}}}_\infty,
\]
which is at most
$2\boundadv w^2\sum_{t=1}^T\gamma^{t-1}(t-1)$ because
$\norm{\Delta_t^{\mathrm{gr}}}_\infty\leq1$. This proves
\eqref{eqn-behavior-centered-improvement}. \qed
\end{proof}

Thus the first-order term uses only the observed state distributions
$d_t^{\pi^b}$, while state-distribution shift enters through $O(w^2)$.
\end{revision}

\subsection{Global policy optimization}

\begin{proof}{Proof of \Cref{lemma-adverror-to-value}}
For binary actions, an action mistake implies
\[
Q_t^*(S_t,\pi_t^*(S_t))-Q_t^*(S_t,\hat\pi_t(S_t))
=\abs{\blipadv_t(S_t)}\mathbb I[\hat\pi_t(S_t)\ne\pi_t^*(S_t)]
\leq\abs{\hat\blipadv_t(S_t)-\blipadv_t(S_t)}.
\]

Suppose the event $\mathcal E$ supplies the integrated $L_2$ bound. For
$\epsilon>0$, keep
the mistake-region set
$\mathcal S_\epsilon=
\{s:\max_a Q_t^*(s,a)-Q_t^*(s,\hat\pi_t(s))\le\epsilon\}$.
The pointwise mistake bound and the disjoint partition give
\begin{align}
&\norm{
Q_t^*(S_t,\pi_t^*(S_t))-Q_t^*(S_t,\hat\pi_t(S_t))
}_2^2\notag\\
&\quad=
\norm{
\{Q_t^*(S_t,\pi_t^*(S_t))-Q_t^*(S_t,\hat\pi_t(S_t))\}
\mathbb I[S_t\in\mathcal S_\epsilon]
}_2^2\notag+
\norm{
\{Q_t^*(S_t,\pi_t^*(S_t))-Q_t^*(S_t,\hat\pi_t(S_t))\}
\mathbb I[S_t\notin\mathcal S_\epsilon]
}_2^2\notag\\
&\quad
\le
\norm{\hat\blipadv_t-\blipadv_t}_2^2.
\label{eqn-mistakeregion-ltwo}
\end{align}
For expected regret, the same partition yields
\begin{align*}
&\E[
Q_t^*(S_t,\pi_t^*(S_t))-Q_t^*(S_t,\hat\pi_t(S_t))
]\\
&\quad\leq
\epsilon
P\!\left(
0<\max_a Q_t^*(S_t,a)-Q_t^*(S_t,\hat\pi_t(S_t))
\le\epsilon
\right)
+
\E[
\abs{\hat\blipadv_t(S_t)-\blipadv_t(S_t)}
\mathbb I[\abs{\hat\blipadv_t(S_t)-\blipadv_t(S_t)}
>\epsilon]
]\\
&\quad\lesssim
\epsilon^{1+\alpha}
+\epsilon^{-1}
\norm{\hat\blipadv_t-\blipadv_t}_2^2.
\end{align*}
Choosing
$\epsilon=\norm{\hat\blipadv_t-\blipadv_t}_2^{2/(2+\alpha)}$
gives
\begin{align*}
&\E[
Q_t^*(S_t,\pi_t^*(S_t))-Q_t^*(S_t,\hat\pi_t(S_t))
] \lesssim
\norm{\hat\blipadv_t-\blipadv_t}_2^
{\frac{2+2\alpha}{2+\alpha}},\\
&\norm{
Q_t^*(S_t,\pi_t^*(S_t))-Q_t^*(S_t,\hat\pi_t(S_t))
}_2\leq
\norm{\hat\blipadv_t-\blipadv_t}_2.
\end{align*}
Substituting the eventwise bound
$\norm{\hat\blipadv_t-\blipadv_t}_2\leq\epsilon_t$
shows that both conclusions hold on the same event. \qed
\end{proof}

\subsubsection{Uniform one-step estimation over greedy continuation policies}
\label{sec:ec-uniform-greedy}

We now go into more detail about the assumptions required to lift our policy evaluation assumptions to hold uniformly over greedy policy optimization suffices, as required in \Cref{sec-policy-opt}. First, we define uniform versions of the policy classes, function classes, and fixed-policy error rates used for policy evaluation.

For each stage $j$, define
\[
\Pi_{j,n}^{\mathrm{gr}}
:=
\left\{
\pi_g:
\pi_g(1\mid s)=\mathbb I[g(s)>0],
\quad g\in\cG_{j,n}
\right\}.
\]
For $t<T$, define the continuation-suffix class
\[
\Pi_{t+1:T,n}^{\mathrm{gr}}
:=
\Pi_{t+1,n}^{\mathrm{gr}}\times\cdots\times
\Pi_{T,n}^{\mathrm{gr}},
\qquad
\Pi_{T+1:T,n}^{\mathrm{gr}}:=\{\emptyset\}.
\]
The main text suppresses the sample-size index $n$. For each suffix
$\pi_{t+1:T}\in\Pi_{t+1:T,n}^{\mathrm{gr}}$, let
$\hat\blipadv_{t,0}^{\pi_{t+1:T}}$ denote the ordinary cross-fitted
one-step estimator thereof. For fold $k$, let
\[
\ell_{t,0}^{\pi_{t+1:T},-k}(O;g)
:=
\left[
\hat Y_{t,0}^{\pi_{t+1:T},-k}
-\hat m_{t,0}^{\pi_{t+1:T},-k}(S_t)
-\{A_t-\hat\pi_t^{b,-k}(1\mid S_t)\}g(S_t)
\right]^2.
\]
Define the pairwise centered conditional loss class
\[
\mathcal H_{t,k}^{\mathrm{unif}}
:=
\left\{
\ell_{t,0}^{\pi_{t+1:T},-k}(\mathord\cdot;g)
-\ell_{t,0}^{\pi_{t+1:T},-k}
(\mathord\cdot;g'):
\pi_{t+1:T}\in\Pi_{t+1:T,n}^{\mathrm{gr}},
\quad g,g'\in\cG_{t,n}
\right\}.
\]
Fix $\delta\in(0,1)$. Let $\bar r_{t,n}$ be a
deterministic upper bound on the largest foldwise conditional critical radius
of the corresponding star hulls after normalization by their common bounded
envelope. Define the greedy-suffix-uniform errors
\[
\bar\varepsilon_{\mathrm{app},t,n}
:=
\sup_{\pi_{t+1:T}\in\Pi_{t+1:T,n}^{\mathrm{gr}}}
\inf_{g\in\cG_{t,n}}
\norm{g-\blipadv_t^{\pi_{t+1:T}}}_2,
\]
\[
\bar\rho_{t,n}
:=
\sup_{\pi_{t+1:T}\in\Pi_{t+1:T,n}^{\mathrm{gr}}}
\max_{k\leq K}\rho_{t,0}^{\pi_{t+1:T},-k},
\qquad
\bar b_{t,n}
:=
\sup_{\pi_{t+1:T}\in\Pi_{t+1:T,n}^{\mathrm{gr}}}
\max_{k\leq K}\norm{b_{t,0}^{\pi_{t+1:T},-k}}_2,
\]
where $\rho_{t,0}^{\pi_{t+1:T},-k}$ is the current-stage remainder in
\cref{eq:diff-rho-L}, and $\bar b_{t,n}$ is the uniform envelope of the
one-step target shift $b_{t,0}^{\pi_{t+1:T},-k}$ in
\cref{eq:diff-target-shift}.

\begin{remark}[Fixed-complexity threshold policies]
\label{rem:pdim-global}
Let $v_j:=\operatorname{Pdim}(\cG_{j,n})$. Thresholding at zero
gives $\operatorname{VCdim}(\Pi_{j,n}^{\mathrm{gr}})\leq v_j$. On a sample of $n$
trajectories, the logarithmic growth bound for the suffix class adds across
stages. Thus its policy-index contribution is of order
$\{\frac{1}{n}({\sum_{j=t+1}^T v_j\log(en/v_j)+\log(TK/\delta)})\}^{1/2}$
up to constants and the usual entropy logarithms, which is
$\widetilde O(n^{-1/2})$ for finite $T$ and pseudo-dimension.
For example, an affine contrast class on a fixed
$m$-dimensional feature set has $v_j\leq m+1$.

If the policy-indexed nuisance classes are also uniformly bounded with
fixed-complexity covering numbers, and the loss is Lipschitz in their outputs,
then the policy, contrast, and nuisance contributions combine to give
$\max_t\bar r_{t,n}=\widetilde O(n^{-1/2})$. This condition does not control
the projected target-shift error $\bar b_{t,n}$, which requires the separate
rate condition in \Cref{cor:uniform-endpoint-rootn} for the one-step estimator.
\end{remark}

  The following uniform oracle inequality verifies \Cref{asn-uniform-global} for
  $L_t=0$.

\begin{proposition}[Uniform one-step oracle inequality]
\label{prop:uniform-one-step}
Suppose \Cref{asn-bounded values,asn-overlap} hold. Suppose also that the
foldwise conditional classes
$\mathcal H_{t,k}^{\mathrm{unif}}$ have critical radii bounded by
$\bar r_{t,n}$, with
$\bar r_{t,n}^2\gtrsim\log(TK/\delta)/\underline n$. Suppose there is a
deterministic sequence $\varepsilon_n\downarrow0$ such that
$\max_{1\leq t\leq T}\{\bar r_{t,n}+\bar\varepsilon_{\mathrm{app},t,n}
+\bar\rho_{t,n}+\bar b_{t,n}\}=O_p(\varepsilon_n)$.
Then, with probability at least $1-\delta$, simultaneously for every stage $t$,
\[
\sup_{\pi_{t+1:T}\in\Pi_{t+1:T,n}^{\mathrm{gr}}}
\norm{
\hat\blipadv_{t,0}^{\pi_{t+1:T}}
-\blipadv_t^{\pi_{t+1:T}}
}_2
\lesssim
\bar r_{t,n}+\bar\varepsilon_{\mathrm{app},t,n}+\bar\rho_{t,n}+\bar b_{t,n}.
\]
Consequently,
\[
\max_{1\leq t\leq T}
\sup_{\pi_{t+1:T}\in\Pi_{t+1:T,n}^{\mathrm{gr}}}
\norm{
\hat\blipadv_{t,0}^{\pi_{t+1:T}}
-\blipadv_t^{\pi_{t+1:T}}
}_2
=O_p(\varepsilon_n),
\]
and \Cref{asn-uniform-global} holds for the one-step schedule $L_t=0$ and the
same deterministic sequence $\varepsilon_n$.
\end{proposition}

\subsubsection{Proofs}

\begin{proof}{Proof of \Cref{prop:uniform-one-step}}
Fix $t$ and $k$, and condition on $\cD_{-k}$. The nuisance fits and the
policy-indexed class $\mathcal H_{t,k}^{\mathrm{unif}}$ are then fixed.
Bounded squared-loss contraction and
\Cref{lemma-lemma14-foster2019}, together with the definition of
$\bar r_{t,n}$, give an event with conditional failure probability at most
$\delta/(TK)$ on which the localized-deviation bound used in the proof of
\Cref{thm-mse-rates-policy-evaluation-apx} holds simultaneously over all
suffixes and candidate pairs.

For any $\varepsilon_{\mathrm{app}}>0$ and each suffix, choose
$g_{t,\pi_{t+1:T}}^{\mathrm{cmp},\varepsilon_{\mathrm{app}}}\in\cG_{t,n}$ whose approximation
error is within $\varepsilon_{\mathrm{app}}$ of the best error over $\cG_{t,n}$:
\[
\norm{g_{t,\pi_{t+1:T}}^{\mathrm{cmp},\varepsilon_{\mathrm{app}}}
-\blipadv_t^{\pi_{t+1:T}}}_2
\leq
\inf_{g\in\cG_{t,n}}\norm{g-\blipadv_t^{\pi_{t+1:T}}}_2+\varepsilon_{\mathrm{app}}
\leq\bar\varepsilon_{\mathrm{app},t,n}+\varepsilon_{\mathrm{app}}.
\]
Set
$\nu_{t,\pi}:=\hat\blipadv_{t,0}^{\pi_{t+1:T}}
-g_{t,\pi_{t+1:T}}^{\mathrm{cmp},\varepsilon_{\mathrm{app}}}$.
The fold target is
$\blipadv_t^{\pi_{t+1:T}}+b_{t,0}^{\pi_{t+1:T},-k}$.
Overlap gives curvature at least $\lambda$ for every suffix, while
\Cref{lem:diff-contemporaneous-orthogonality,lemma-secondorderderivatives}
bound the current-stage plug-in error by $\bar\rho_{t,n}$.
Substituting \cref{eq:diff-rho-L,eq:diff-target-shift} into the basic
inequality for \cref{eq:recursive-unrolled-loss-main} gives, simultaneously
over the suffixes,
\[
\lambda\norm{\nu_{t,\pi}}_2^2
\lesssim \bar r_{t,n}
\{\norm{\nu_{t,\pi}}_2+\bar r_{t,n}\}
+\{\bar\varepsilon_{\mathrm{app},t,n}+\varepsilon_{\mathrm{app}}+\bar\rho_{t,n}+\bar b_{t,n}\}
\norm{\nu_{t,\pi}}_2.
\]
Young's inequality, the triangle inequality, and $\varepsilon_{\mathrm{app}}\downarrow0$ give the
stated bound. A union bound over the $TK$ stage-fold events gives probability
at least $1-\delta$.
\qed
\end{proof}

\begin{proof}{Proof of \Cref{cor:uniform-endpoint-rootn}}
Uniform well specification gives $\bar\varepsilon_{\mathrm{app},t,n}=0$, and the uniform
current-stage nuisance-product condition gives
$\max_t\bar\rho_{t,n}=o_p(n^{-1/2})$. By \Cref{rem:pdim-global}, fixed
pseudodimension and the stated nuisance-class conditions give
$\max_t\bar r_{t,n}=\widetilde O(n^{-1/2})$.

For $L_t=0$, the assumed bound on $\bar b_{t,n}$ and
\Cref{prop:uniform-one-step} give the result. For $L_t=T-t$, repeat the
conditional uniform empirical-process argument in that proposition with the
fully unrolled outcome. Its target shift has no terminal continuation-value
term. It contains only future contrast errors multiplied by policy differences.
Applying the overlap bound stage by stage gives the strictly upper-triangular
recursion used in the proof of \Cref{cor:full-unrolling-rate}, now with
suprema over the greedy-suffix classes. Its coefficients are bounded uniformly because
$\norm{\pi_j-\pi_j^b}_\infty\leq1$. Since $T$ is fixed, backward substitution
preserves the $\widetilde O_p(n^{-1/2})$ stagewise rate. This proves the
full-unrolling claim. \qed
\end{proof}

\begin{proof}{Proof of \Cref{thm-policy-optimization}}
\leavevmode
\paragraph{Step 1: Uniform endpoint estimation at the realized suffix.}
By construction,
$\hat\pi_{t+1:T}\in\Pi_{t+1:T,n}^{\mathrm{gr}}$ for every stage $t$.
\Cref{asn-uniform-global} implies
that, for each fixed $\delta>0$, there is a constant $C(\delta)$ such that, with
probability at least $1-\delta$, the uniform bound holds simultaneously over
the finite collection of stages. Thus the eventwise input bound $\epsilon_t$
of \Cref{lemma-adverror-to-value} can be set to $\varepsilon_n$ after
absorbing $C(\delta)$ into the constants below. On this common event,
\[
\norm{\hat\blipadv_{t,L_t}^{\hat\pi_{t+1:T}}
-\blipadv_t^{\hat\pi_{t+1:T}}}_2
\lesssim
\epsilon_t
\qquad\text{for every }t.
\]

\paragraph{Step 2: Establishing policy consistency.}
We first record the $L_2$ effect of replacing the optimal future policy by
the estimated future policy. For either fixed action $a\in\{0,1\}$,
\[
\norm{\E[V_{t+1}^{\pi^*}-V_{t+1}^{\hat\pi_{\hat\blipadv}}\mid S_t,A_t=a]}_2^2
\le
\E[\E[(V_{t+1}^{\pi^*}-V_{t+1}^{\hat\pi_{\hat\blipadv}})^2
\mid S_t,A_t=a]]
\le
C_\infty\norm{V_{t+1}^{\pi^*}-V_{t+1}^{\hat\pi_{\hat\blipadv}}}_2^2.
\]
Therefore
\begin{equation}
\norm{\blipadv_t^{\hat\pi_{t+1:T}}
-\blipadv_t^{\pi^*_{t+1:T}}}_2
=
\gamma\norm{\cC_{t,t+1}[V_{t+1}^{\hat\pi_{\hat\blipadv}}-V_{t+1}^{\pi^*}]}_2
\le
2\gamma\sqrt{C_\infty}
\norm{V_{t+1}^{\pi^*}-V_{t+1}^{\hat\pi_{\hat\blipadv}}}_2.
\label{eqn-apx-policyopt-inductionstep-decomposition}
\end{equation}
Combining this display with Step 1 gives the first ordinary $L_2$ recursion:
\[
\norm{\hat\blipadv_{t,L_t}^{\hat\pi_{t+1:T}}
-\blipadv_t^{\pi^*_{t+1:T}}}_2
\leq\norm{\hat\blipadv_{t,L_t}^{\hat\pi_{t+1:T}}
-\blipadv_t^{\hat\pi_{t+1:T}}}_2
+\norm{\blipadv_t^{\hat\pi_{t+1:T}}-\blipadv_t^{\pi^*_{t+1:T}}}_2
\lesssim\epsilon_t+2\gamma\sqrt{C_\infty}
\norm{V_{t+1}^{\pi^*}-V_{t+1}^{\hat\pi_{\hat\blipadv}}}_2.
\]

For binary actions, apply the eventwise $L_2$ bound in
\Cref{lemma-adverror-to-value} on this same event. It gives
$\norm{Q_t^*(S_t,\pi_t^*)-Q_t^*(S_t,\hat\pi_t)}_2
\le
\norm{\hat\blipadv_{t,L_t}^{\hat\pi_{t+1:T}}
-\blipadv_t^{\pi^*_{t+1:T}}}_2$.
The Bellman recursion and the fixed-action bound above give the second
ordinary $L_2$ recursion:
\begin{align*}
\norm{V_t^{\pi^*}-V_t^{\hat\pi_{\hat\blipadv}}}_2
&\le
\norm{Q_t^*(S_t,\pi_t^*)-Q_t^*(S_t,\hat\pi_t)}_2
+\gamma
\norm{\E[V_{t+1}^{\pi^*}-V_{t+1}^{\hat\pi_{\hat\blipadv}}
\mid S_t,A_t=\hat\pi_t(S_t)]}_2\\
&\le
\norm{\hat\blipadv_{t,L_t}^{\hat\pi_{t+1:T}}
-\blipadv_t^{\pi^*_{t+1:T}}}_2
+\gamma\sqrt{C_\infty}
\norm{V_{t+1}^{\pi^*}-V_{t+1}^{\hat\pi_{\hat\blipadv}}}_2.
\end{align*}

Since $V_{T+1}^{\pi^*}=V_{T+1}^{\hat\pi_{\hat\blipadv}}=0$, the two
recursions give the base case at $T$. If the value-gap norm at $t+1$ is
$O(\max_{k'\geq t+1}\epsilon_{k'})$, then
\[
\norm{\hat\blipadv_{t,L_t}^{\hat\pi_{t+1:T}}
-\blipadv_t^{\pi^*_{t+1:T}}}_2
=O\!\left(\max_{k'\geq t}\epsilon_{k'}\right),
\qquad
\norm{V_t^{\pi^*}-V_t^{\hat\pi_{\hat\blipadv}}}_2
=O\!\left(\max_{k'\geq t}\epsilon_{k'}\right).
\]
Thus, for every $k$,
\begin{equation}
\norm{
\hat\blipadv_{k,L_k}^{\hat\pi_{k+1:T}}
-\blipadv_k^{\pi^*_{k+1:T}}
}_2
=O\!\left(\max_{k'\geq k}\epsilon_{k'}\right).
\label{eqn-apx-policyopt-inductionhyp1}
\end{equation}
The inductive use of
\cref{eqn-apx-policyopt-inductionstep-decomposition} requires
\cref{eqn-apx-policyopt-inductionhyp1}.

Finally, apply the eventwise expected-regret bound in
\Cref{lemma-adverror-to-value} once on the same common event. The lemma does
not require concentrability. The finite-horizon performance-difference
identity then uses \Cref{asn-concentrability} to change the state distribution
and gives
\begin{align}
\abs{\E[V_k^{\pi^*}-V_k^{\hat\pi_{\hat\blipadv}}]}
&\le
C_\infty
\sum_{j=k}^T\gamma^{j-k}
\E[
Q_j^*(S_j,\pi_j^*)-Q_j^*(S_j,\hat\pi_j)
]
\lesssim
C_\infty
\sum_{j=k}^T\gamma^{j-k}
\norm{
\hat\blipadv_{j,L_j}^{\hat\pi_{j+1:T}}
-\blipadv_j^{\pi^*_{j+1:T}}
}_2^{\frac{2+2\alpha}{2+\alpha}}
\notag\\
&\lesssim
C_\infty
\sum_{j=k}^T
\left(\max_{k'\geq j}\epsilon_{k'}\right)^
{\frac{2+2\alpha}{2+\alpha}}
\lesssim
C_\infty
\left(\max_{k'\geq k}\epsilon_{k'}\right)^
{\frac{2+2\alpha}{2+\alpha}}.
\label{eqn-apx-policyopt-value-rate}
\end{align}
Substituting $\epsilon_t=\varepsilon_n$ makes
$\max_{k'\geq k}\epsilon_{k'}=\varepsilon_n$. This proves the theorem. \qed
\end{proof}

\section{Experimental details}\label{sec-experimental-details}

All experiments use synthetic data generated by  the accompanying Python code. The replication package includes the simulation, estimation, evaluation, and figure-generation scripts, including software-environment specification. Computations were performed either on an Apple M1 MacBook Pro with eight CPU cores and 16 GB of memory or on a CPU cluster using up to 64 cores and 8 GB of memory per core. 

\subsection{Data generating process details on ``Adapting to Structure'' experiment}

For the reward-filtered DGP, $\vert \mathcal{S} \vert = 100$ though the first $15$ 
dimensions are the reward-relevant sparse component, where $\rho$ is the indicator vector of the sparse support, and $\mathcal{A} = \{ 0, 1\}$. The reward and states evolve according to $r_t(s,a) = \beta^\top \phi_t(s,a) + a*\sum_{k=1}^5 s_k /2 + \epsilon_r, \;\; s_{t+1}(s,a) = M_a s + \epsilon_s,$ satisfying the graphical restrictions of \Cref{fig:rewardrelevant}. Therefore the transition matrices are $M_a = \begin{bmatrix} M_{a}^{\rho \to \rho } & 0 \\
M_{a}^{\rho \to {\rho_c} } & M_{a}^{{\rho_c} \to {\rho_c} } 
\end{bmatrix}.$ {We generate the coefficient matrices $M_0,M_1$ with independent normal random variables $\sim N(0.2, 1)$.} The nonzero mean ensures the beta-min condition. We normalize $M_{a}^{\rho \to \rho }$ to have spectral radius 1, then introduce mild instability in the exogenous component by dividing $M_{a}^{{\rho_c} \to {\rho_c} }$ by $0.8$x the largest eigenvalue. Therefore, recovering the sparse component is stable but including distracting dimensions destabilizes. The noise terms are normally distributed with standard deviations $\sigma_s = 0.3, \sigma_r = 0.5.$ Features $\phi(s,a) = \langle s, s  a, 1 \rangle$ are the interacted state-action space. 
The behavior policy is a mixture of logistic, with coefficients $\sim N(0,0.3)$, and 20\% probability of uniform random sampling. The evaluation policy is logistic, with coefficients $\sim \text{Unif}[-0.5,0.5].$ (We fix the random seed).

In ``Misaligned endo-exo,'' { we follow the same data-generating process as the ``Reward-Filtered DGP'' described earlier, but} we change the blockwise conditional independences to follow the exogenous--endogenous model of \citet{dietterich2018discovering} (see \Cref{fig:exoendo}). {We additionally added dense rewards to the reward vector, adding $\beta^{\top}_{dense} \phi_t(s, a)$ where the entries of $\beta_{dense}$ are 1 w.p. 0.9.} Here, reward sparsity of $R(s,a), a \in \{0,1\}$ alone does not recover the sparse component. Reward-filtered thresholded LASSO is simply misspecified and does very poorly (off the graph limits). Likewise, in small samples, vanilla thresholded LASSO FQE ($\textrm{FQE-TL}$, dark-blue) includes too many extra dimensions. But for small-data regimes, imposing thresholded LASSO \textit{on the difference of $Q$ functions} remains stable. 

The final DGP introduces ``nonlinear main effects'': again we generate a $50\%$ dense vector $\beta_{dense}$ and we add $s^\top \beta_{dense} + 3\sin(\pi s_{49}s_{48} ) +  0.5(s_{49}-0.5)^2 +0.5(s_{48} - 0.5)^2
$. (These nonlinear main effects are  disjoint from the sparse difference-of-Q terms). For small $n$, FQE wrongly includes extraneous dimensions that destabilize estimation, and our methods estimating $\tau$ with reward-thresholded-LASSO outperform naive FQE with thresholded-LASSO for small data sizes. 
\begin{revision}
\subsection{Linear-Gaussian sparse policy-learning experiment}
\label{app:linear-gaussian-policy-details}

This experiment isolates policy-learning gains from estimating a sparse
difference-of-$Q$ function when the action-invariant prognostic component of
the $Q$-function is dense. There are eight decision periods, binary actions,
discount factor $\gamma=0.95$, and a 150-dimensional state
$S_t=(X_t,Z_t)$, where $X_t\in\mathbb R^{120}$ contains prognostic
coordinates and $Z_t\in\mathbb R^{30}$ contains candidate effect modifiers.
The initial state is standard normal. For $t=1,\ldots,8$, the transition and
reward equations are
\begin{align}
 X_{t+1}&=0.60X_t+\epsilon^X_t, \qquad
 Z_{t+1}=0.65Z_t-0.40A_te_1+\epsilon^Z_t,
 \label{eq:linear-policy-transition}\\
 R_t&=\mu_t+w_X^\top X_t+w_Z^\top Z_t
 +A_t\{\theta_0+\theta_Z^\top Z_t\}+\epsilon^R_t,
 \label{eq:linear-policy-reward}
\end{align}
where $\mu_t$ is evenly spaced from $0.20$ to $-0.20$,
$\epsilon^X_t$ and $\epsilon^Z_t$ have independent centered Gaussian
coordinates with standard deviation $0.55$, and
$\epsilon^R_t\sim N(0,1)$. We draw $w_X$ and $w_Z$ once from independent
standard-normal vectors and rescale them to have Euclidean norms $3.0$ and
$1.5$, respectively. The action-effect parameters are
$\theta_0=-1.6$ and
$\theta_Z=8(1,0.9,-1.1,0,\ldots,0)^\top$. Thus, all 150 state coordinates
can enter the action-invariant reward, but only the first three coordinates
of $Z_t$ modify the action contrast. The behavior policy independently
assigns each action with probability $1/2$. Feasible estimator starts from the same feature dictionary $\Phi(S_t,A_t)=(1,S_t,A_t,A_tS_t)$.
For each total training size
$n\in\{100,200,400,800,1600,3200,6400\}$, we independently generate one
selection sample of $n/2$ trajectories and one refit sample of $n/2$
trajectories. DiffQ uses $K=2$ folds for sample splitting. 
The future full-$Q$ and state-conditional mean nuisances use ridge regression;
their ridge penalties are selected from 16 logarithmically spaced values
between $10^{-3}$ and $10^4$ using training data only.

For the refit on the selected support, we standardize the data. At each stage, a ridge
pilot supplies residuals for 2,000 Gaussian multiplier-bootstrap draws. We
take the $1-0.05/8$ quantile of the maximum absolute score and multiply it by
$0.75$; the resulting value is used as both the LASSO penalty and the
post-LASSO coefficient threshold. The horizon-union support is the union of
the eight stage-specific supports. Holding this support fixed, we refit every
stage by least squares on the independent refit sample, always retaining the
action-intercept column. The matched stage-specific screen instead retains only the corresponding stage's support, and we discuss it as an ablation for our union rule. 

 The full-$Q$ comparator applies ridge fitted-$Q$
evaluation under to all selection and refit
trajectories, using the same dictionary and ridge grid, and then takes actions greedy with respect to $Q$. 
For each $n$, we use 32 independent training replications. All policies are evaluated on 2,000 fresh trajectories using the
same initial states and transition and reward noise. The figure reports mean
discounted returns and two-sided 95\% Student-$t$ intervals across
replications.

\subsection{Nonlinear derivative-thresholded KRR experiment}
\label{app:nonlinear-krr-details}

This experiment compares the $L=0$ and $L=T-t$ endpoints for estimating the
fixed-policy contrast $\blipadv_0^{\pi^e}$. $T=5$, $\gamma=1$, and $S_t\in\mathbb R^{20}$. Initial
states have independent $\operatorname{Unif}[-1,1]$ coordinates. Let
$S_{t,B}=S_{t,1:5}$, $S_{t,I}=S_{t,6:7}$, and
$S_{t,N}=S_{t,8:20}$. The transition is
\begin{align}S_{t+1,B}&=B_tS_{t,B}+A_tC_tS_{t,B}+\epsilon_{t,B},\nonumber\\
 S_{t+1,I}&=0.65S_{t,I}+\epsilon_{t,I},\qquad
 S_{t+1,N}=N_tS_{t,N}+\epsilon_{t,N},
 \label{eq:nonlinear-krr-transition}
\end{align}
with independent centered Gaussian transition noise of standard deviation
$0.3$. The entries of the unscaled $B_t$ and $N_t$ matrices are drawn from
$N(0.2,1)$ and each matrix is rescaled to have spectral radius $0.65$. The
entries of the unscaled $C_t$ matrices are drawn from $N(0.1,0.5^2)$ and each
matrix is rescaled to have operator norm $0.12$. The reward is
\begin{align}
 R_t={}&b(S_t)+A_t\Delta(S_t)+\epsilon^R_t,\qquad
 \epsilon^R_t\sim N(0,0.3^2),\nonumber\\
 b(s)={}&\beta_{\rm dense}^\top s
 +3\sin(\pi s_{19}s_{20})
 +0.5(s_{19}-0.5)^2+0.5(s_{20}-0.5)^2,\nonumber\\
 \Delta(s)={}&1.5\sum_{j=1}^5s_j+3\sin(\pi s_6s_7),
 \label{eq:nonlinear-krr-reward}
\end{align}
where the entries of $\beta_{\rm dense}$ are drawn once from
$\operatorname{Bernoulli}(1/2)$. The behavior and evaluation policies are
\begin{align*}
 \pi_t^b(1\mid s)&=0.1+0.8\operatorname{expit}(s_{1:5}^\top u_t),\\
 \pi_t^e(1\mid s)&=0.1+0.8\operatorname{expit}(-0.5s_{1:5}^\top u_t),
\end{align*}
where each $u_t$ is drawn once coordinatewise from
$\operatorname{Unif}[-0.3,0.3]$. All coefficient draws are fixed across
replications. Under this construction, the coordinate support of
$\blipadv_0^{\pi^e}$ is
$K_{\Delta,0}=\{1,\ldots,7\}$: the first five coordinates enter the
action-responsive transition and the next two enter the nonlinear immediate
effect. The remaining coordinates contribute only action-invariant main
effects and autonomous dynamics.

For each $n\in\{400,800,1600,3200,6400\}$ and each endpoint, one sample of
$n$ trajectories is used for screening and an independent sample of $n$
trajectories is used for refitting, and we use $K=5$ folds for cross-fitting. For $L=0$, the continuation value in the
one-step pseudo-outcome is estimated by action-specific RBF-kernel fitted-$Q$
evaluation. For $L=T-t$, nuisance functions also use RBF kernel ridge regression, with
three-fold training-only tuning inside each outer fold.

The screening regression fits the residualized outcome on the residualized
action using a Gaussian RBF kernel. Coordinates are standardized using the
training sample, and the bandwidth is the median positive pairwise distance.
For every ridge value $\lambda=c/\sqrt n$, where $c\in\{0.0006,0.002,0.006,0.02,0.06,0.2\}$,
we calculate the empirical integrated squared derivative
\[
 \widehat D_j(\lambda)
 =\frac{1}{n}\sum_{i=1}^n
 \left\{
 \frac{\partial\widehat\blipadv_{0,\lambda}(S_{i0})}
 {\partial s_j}
 \right\}^2.
\]
At a given ridge value, coordinate $j$ is retained when
$\widehat D_j(\lambda)>0.30\max_k\widehat D_k(\lambda)$. The final support is
the union of these six ridge-specific sets. 

On the independent refit sample, three-fold held-out residualized loss selects
one ridge value with the union support held fixed. We use that same ridge parameter for
three RBF-kernel refits compared in the figure: the feasible union support, all 20 state coordinates,
and the infeasible oracle support $K_{\Delta,0}$. MSE is evaluated on 2,000
independent initial states. The reference value of
$\blipadv_0^{\pi^e}$ at each state is computed from 512 paired
common-random-number continuation rollouts per initial action. The one-step
policy replaces only the time-zero action of $\pi^e$ by
$\widehat\pi_0(s)=\mathbb I\{\widehat\blipadv_0^{\pi^e}(s)\geq0\}$ and follows
$\pi^e$ thereafter. Its reported gain is
\[
 \E\!\left[
 \{\widehat\pi_0(S_0)-\pi_0^e(1\mid S_0)\}
 \blipadv_0^{\pi^e}(S_0)
 \right].
\]
We use 50 independent replications at each sample size and report two-sided
95\% Student-$t$ intervals across replications.

\subsection{Markovian interference experiments: ridesharing simulator details}\label{app:dgp}

This appendix defines the two simulation experiments used in the main text.
We use the event-level notation for the released ridesharing simulator,
following \citet{johari2025estimation}.

Each trajectory contains 500,000 chronologically ordered potential ride
requests. The request record contains an arrival time and Manhattan pickup and
dropoff nodes. Travel times come from the released Manhattan distance matrix.
The simulator starts with 300 vehicles. When a rider accepts, the simulator
dispatches the vehicle with the earliest pickup arrival time. The vehicle then
becomes unavailable until its trip ends. Rejection leaves the fleet state
unchanged.

Let \(i\) index requests and let \(b(i)\) denote the 1,200-second block that
contains request \(i\). A block action \(Z_b\in\{0,1\}\) applies to every
request in block \(b\). Under the behavior policy,
\(Z_b \stackrel{\mathrm{ind}}{\sim} \operatorname{Bernoulli}(1/2)\).

The control and treatment price rates are
\(\alpha_0=0.01\) and \(\alpha_1=0.02\). For trip duration \(d_i\), the quoted
price is \(P_i(z)=\alpha_zd_i\). Let \(\eta_i\) denote pickup ETA. In the
released response model, rider acceptance \(C_i\) satisfies

\begin{equation}
 P(C_i=1\mid X_i,Z_{b(i)}=z)
 =\operatorname{logit}^{-1}\!\left(
   w_p P_i(z)+w_\eta\eta_i+w_0
 \right),
 \label{eq:released-choice}
\end{equation}

where \(w_p=-0.3\), \(w_\eta=-0.005\), and \(w_0=4\). An accepted request
produces reward \(P_i(z)\); a rejected request produces zero. These values and
the dispatch rule match the released simulator.

We aggregate requests within each nonempty switchback block. For trajectory
\(r\) and block \(b\), \(X_{rb}\) is the state recorded before assigning
\(Z_{rb}\), and \(Y_{rb}\) is mean reward per request in that block. A policy's
trajectory value is the request-weighted mean reward

\begin{equation}
  V_r(\pi)=
  \frac{\sum_b N_{rb}Y_{rb}}{\sum_b N_{rb}},
  \label{eq:trajectory-value}
\end{equation}

where \(N_{rb}\) is the number of requests in the block. Policy comparisons
average \eqref{eq:trajectory-value} across independent rollout trajectories.

\begin{table}[ht]
\centering
\caption{Configuration shared by both experiments.}
\label{tab:common-config}
\small
\begin{tabular}{@{}ll@{}}
\toprule
Component & Frozen value \\
\midrule
Input events & First 500,000 valid September 2024 Manhattan requests \\
Vehicles & 300 \\
Control and treatment price rates & 0.01 and 0.02 \\
Acceptance coefficients & \(w_p=-0.3\), \(w_\eta=-0.005\), \(w_0=4\) \\
Switchback block length & 1,200 seconds \\
Behavior treatment probability & 0.5, independently by block \\
Truncation lag & \(k=2\), hence three consecutive block rewards \\
Training sizes & \(n\in\{4,8,16,32,64,100\}\) trajectories \\
\bottomrule
\end{tabular}
\end{table}
\end{revision}
\subsection{Details on nonlinear mutual information extension}\label{apx-experiments-mi}
Our experiments showcase that support recovery is necessary. To illustrate how the loss function approach permits more complex parametrizations, we now consider neural-nets and introduce a heuristic nonlinear sparsity-encouraging regularizer based on mutual information regularization. We seek a simpler representation that retains information related to the loss, while discarding irrelevant information unrelated to the proxy loss for the difference-of-Q functions. We use a mutual information regularizer (MIR), $\hat{\mathcal{L}}_{MI} (\phi,\theta)$, to encourage decomposing the state $S$ into independent nonlinear representations $X_c^\phi, X_a^\theta$, parametrized respectively by $\phi,\theta.$ Mutual information quantifies the dependency between two variables and it equals zero if and only if they are (marginally) independent. We parametrize the difference-of-Q function as $\blipadv(X_c^\phi)$, depending only on the \textit{confounding} information $X_c^\phi$ which is relevant to the difference-of-Q loss function, while the \textit{auxiliary} information $X_a^\theta$ is independent of the loss function. We also add a reconstruction loss function $\hat{\mathcal{L}}_{rec}(\phi,\theta)$ which ensures that these two representations jointly recover the state. These additional loss functions are weighted by hyperparameters $\lambda_{m}, \lambda_{r}$.
\begin{align*}
     &\hat{\mathcal{L}}^{nl}(\tau, \eta;\phi,\theta)  = \hat{\mathcal{L}}(\tau_\phi, \eta) + \lambda_{m} \hat{\mathcal{L}}_{MI} (\phi,\theta)
 + \lambda_{r} \hat{\mathcal{L}}_{rec}(\phi,\theta),
 \\
 & 
 \text{ where }
  \hat{\mathcal{L}}_{MI} = |\hat{I}(X_a^{\phi};X_c^{\theta})|, \hat{\mathcal{L}}_{rec} = \E[(X_a^{\phi}+X_c^{\theta} - S)^2 ]
\end{align*}
Estimating mutual information is challenging. We use a recently developed mutual information neural-networks based estimator, abbreviated MINE \citep{belghazi2018mutual}. (See \Cref{sec-experimental-details} for more details). MINE defines a neural information measure $I_{\Lambda}(X_a,X_c)  =  {\sup_{\lambda\in\Lambda} \EE_{\PJ{X_a}{X_c}}[\FGen_\lambda] - \log(\EE_{\PI{X_a}{X_c}}[e^{\FGen_\lambda}])}.$ Usually MI requires access to probability densities, though only samples from the joint distribution in ML-based methods are available. MINE uses these samples.

\textbf{Cartpole with Distractors results.} We use a cartpole-with-distractors environment from \citet{hao2024forward}, which adds additional autoregressive noise to the state in CartPole \citep{brockman2016openai}. We treat this long finite-horizon environment with time-homogeneous transitions as a $\gamma=0.99$ discounted infinite-horizon problem and pool the data. \citet{hao2024forward} focuses on off-policy evaluation with abstractions, so the methods are not comparable.
We illustrate how our method improves upon naive FQE (learned with neural nets) learned on the full state space. The results are in \Cref{tab:nonlinear_results}. We evaluate the MSE relative to an oracle difference-of-Q function obtained by differencing $Q$ estimates from FQE from a large dataset, $n=2000,$ trained only on the original 4-d state space. We compare to our DiffQ estimation with neural nets, and a regularized version. 

\begin{table*}[t!]
\centering
\caption{Performance comparison across different sample sizes in nonlinear settings. Results show MSE ± SE.}
\resizebox{\textwidth}{!}{
\begin{tabular}{l|c|c|c|c|c}
\toprule
\multirow{2}{*}{Method} & \multicolumn{5}{c}{Number of Samples (n)} \\ \cline{2-6}
 & 100 & 200 & 400 & 600 & 800 \\ \hline
\hline
\multicolumn{6}{c}{\textbf{CartPole with Distractors}} \\ \hline
FQE & $2.367 \pm 2.157$ & $0.587 \pm 0.772$ & $1.157 \pm 2.219$ & $1.793 \pm 1.618$ & $4.123 \pm 3.901$ \\ \hline
DiffQ & $2.212 \pm 2.376$ & $0.415 \pm 0.463$ & $1.228 \pm 1.831$ & $1.929 \pm 2.126$ & $2.440 \pm 1.912$ \\ \hline
DiffQ+MI & $\mathbf{2.104 \pm 2.392}$ & $\mathbf{0.280 \pm 0.222}$ & $\mathbf{1.179 \pm 1.840}$ & $\mathbf{1.286 \pm 1.123}$ & $\mathbf{2.342 \pm 1.812}$ \\ \hline
\hline
\multicolumn{6}{c}{\textbf{Fourrooms Navigation ($\times 10^{-3}$)}} \\ \hline
FQE & $0.299 \pm 0.520$ & $0.194 \pm 0.167$ & $0.238 \pm 0.230$ & $0.175 \pm 0.096$ & $0.201 \pm 0.151$ \\ \hline
DiffQ & $0.281 \pm 0.456$ & $0.177 \pm 0.142$ & $0.224 \pm 0.261$ & $0.173 \pm 0.055$ & $0.187 \pm 0.158$ \\ \hline
DiffQ+MI & $\mathbf{0.272 \pm 0.461}$ & $\mathbf{0.173 \pm 0.171}$ & $\mathbf{0.219 \pm 0.239}$ & $\mathbf{0.170 \pm 0.094}$ & $\mathbf{0.180 \pm 0.174}$ \\ \bottomrule
\end{tabular}}
\label{tab:nonlinear_results}
\end{table*}

\textbf{Four Rooms Environment Results.} To further validate our approach in complex discrete environments, we evaluated our method on the D4RL/minigrid fourrooms-random-v0 environment, which features a discrete action space with 7 actions and high-dimensional state representations (148 dimensions from visual input). This environment requires complex spatial reasoning and navigation. The Four Rooms results in Table~\ref{tab:nonlinear_results} confirm that our method successfully \textit{scales} to environments with multiple discrete actions and high-dimensional state representations. Further including irrelevant distractors could yield further improvements. 

\begin{figure}[t!]
    \centering
\includegraphics[width=0.5\textwidth]{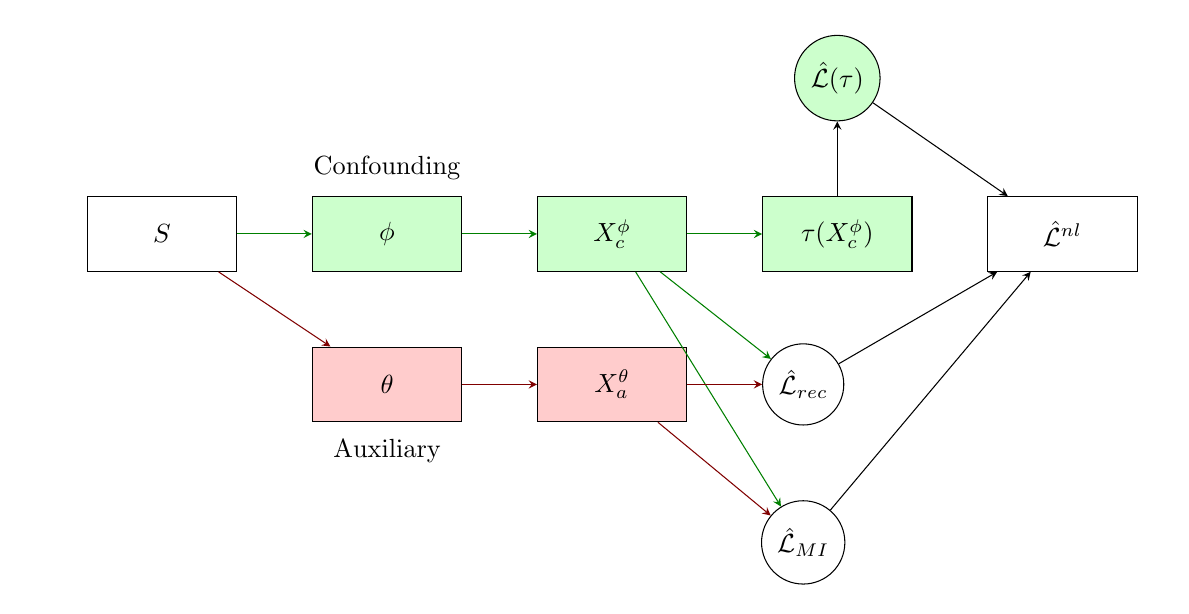}
    \caption{Heuristic neural network architecture diagram for nonlinear mutual information regularization.}
    \label{fig:architecture-nonlin}
\end{figure}

\end{document}